\documentclass[10pt,twocolumn,letterpaper]{article}

\usepackage{cvpr}              
\definecolor{cvprblue}{rgb}{0.21,0.49,0.74}
\usepackage[pagebackref,breaklinks,colorlinks,allcolors=cvprblue]{hyperref}

\def\paperID{xxx} 
\def\confName{CVPR}
\def\confYear{2026}

\usepackage{booktabs} 
\usepackage[table]{xcolor}
\definecolor{ForestGreen}{RGB}{34,139,34}
\usepackage{array}
\usepackage{caption}       
\usepackage{multirow}      
\usepackage{pifont}
\newcommand{\x}{×}
\usepackage{tikz}
\usepackage{subcaption}
\usepackage{overpic}
\newcommand{\xmark}{\ding{55}}  
\newcommand{\cmark}{\ding{51}}  

\usepackage{xspace}  
\definecolor{chatgptgreen}{RGB}{16,163,127}
\definecolor{OTpurple}{RGB}{134,123,203}
\definecolor{Cyan_Light}{RGB}{238,255,255}
\definecolor{Cyan_Dark}{RGB}{220,255,255}
\definecolor{Yellow_Light}{RGB}{255,252,235}
\definecolor{Yellow_Dark}{RGB}{255,240,180}
\newcolumntype{P}[1]{>{\centering\arraybackslash}m{#1}}
\usepackage{tabularx}
\usepackage{needspace}
\usetikzlibrary{arrows.meta}
\usepackage{amsmath}

\title{Seeing Through Fog: Towards Fog-Invariant Action Recognition}

\author{
Enqi Liu$^{1,2}$, 
Liyuan Pan$^{1,3}$\thanks{Corresponding author.}, 
Zhi Gao$^{1,2}$\footnotemark[1], 
Lingzhi Li$^{1}$, 
Qing Li$^{2}$\\
$^{1}$ Beijing Institute of Technology, Beijing, China\\
$^{2}$ Beijing Institute for General Artificial Intelligence, Beijing, China\\
$^{3}$ Yangtze Delta Region Academy of Beijing Institute of Technology, Jiaxing, China \\
{\tt\small \{enqi.liu, liyuan.pan, zhi.gao, lingzhi.li\}@bit.edu, dylan.liqing@gmail.com}
}

\begin{document}
\maketitle
\begin{abstract}
Foggy conditions are commonly encountered in real-world applications; however, existing action recognition approaches typically assume favorable weather and high-quality video inputs. On foggy days, unpredictable visibility degradation and reduced contrast obstruct the extraction of semantic cues, posing significant challenges for current action recognition methods. In this paper, we mitigate the issues faced in action recognition under foggy conditions by employing two strategies. First, we present FogAct, the first benchmark dataset for foggy action recognition, consisting of paired clean and foggy videos captured with a stereo camera system. The dataset spans 10 scenes and 55 action categories, comprising nearly 10,000 video clips.
Second, we propose FogNet, a two-stream CLIP model that discovers fog-invariant semantic information hidden behind the degraded videos. FogNet learns robust representations of foggy videos with guidance from clean videos, effectively capturing shared structural and motion cues between clean and foggy videos. Extensive experiments on FogAct and three other popular datasets demonstrate that our method achieves competitive performance compared with state-of-the-art (SOTA) approaches. Our FogAct and FogNet are given in \href{https://github.com/Liu-arch/Seeing-Through-Fog-Towards-Fog-Invariant-Action-Recognition}{our project page}.
\end{abstract}
\section{Introduction}
Action recognition plays a critical role in surveillance \cite{doshi2022multi, fangsurveillance}, autonomous driving \cite{hu2023planning, dong2025rgbevent}, and human-computer interaction \cite{hassan2021populating}, all requiring accurate action recognition under adverse weather conditions such as fog. Videos captured in foggy conditions, with reduced visibility, blurring, and low contrast, hinder robust feature extraction and complicate the detection of motion details invariant to fog, leading to reduced performance in existing action recognition approaches. This paper seeks to address these challenges and enable accurate foggy action recognition.

\begin{figure}[t]
    \centering
    \setlength{\fboxsep}{0pt}
    \begin{minipage}{0.32\linewidth}
        \centering
        \fcolorbox{gray}{white}{\includegraphics[width=\linewidth]{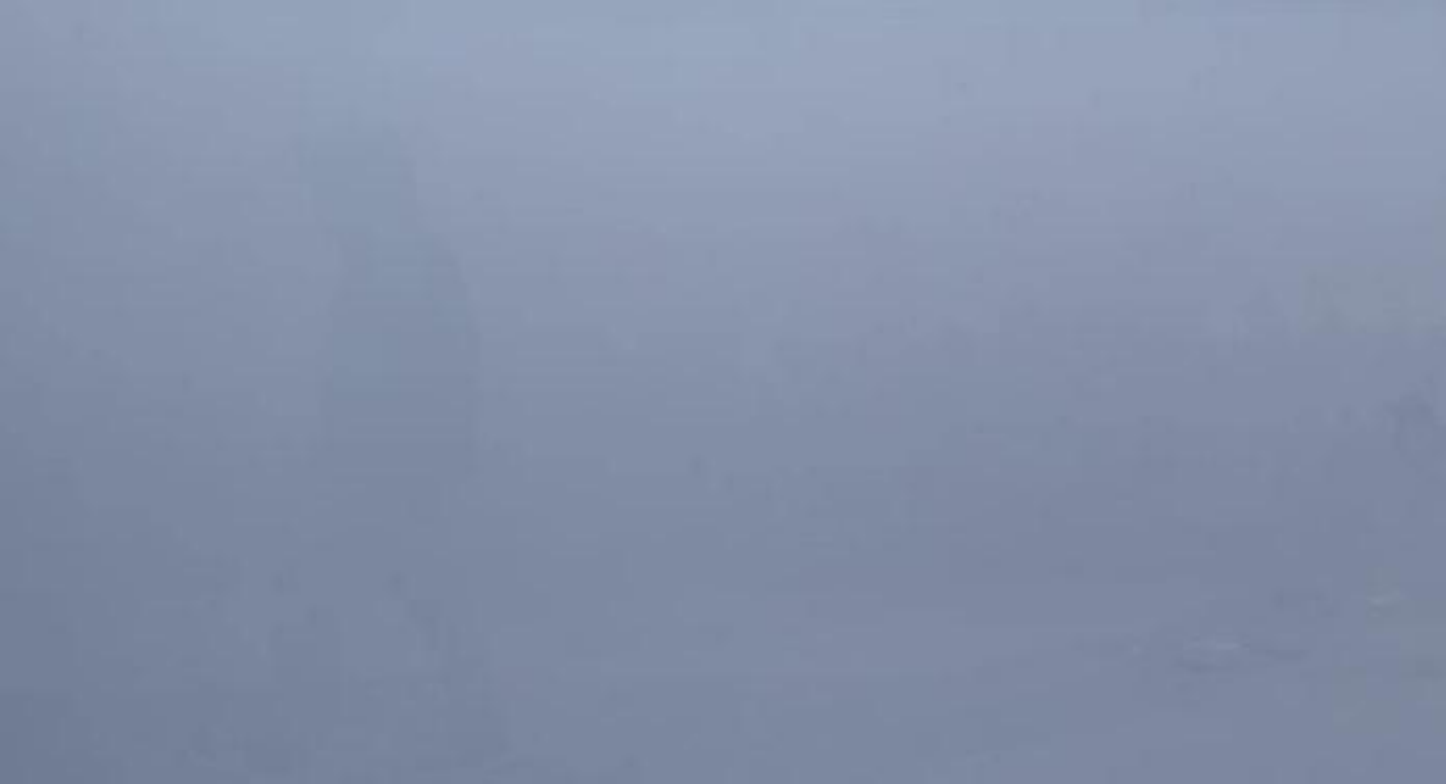}}
        \subcaption{Foggy Image}\label{fig:new1}
    \end{minipage}
    \begin{minipage}{0.32\linewidth}
        \centering
        \fcolorbox{gray}{white}{\includegraphics[width=\linewidth]{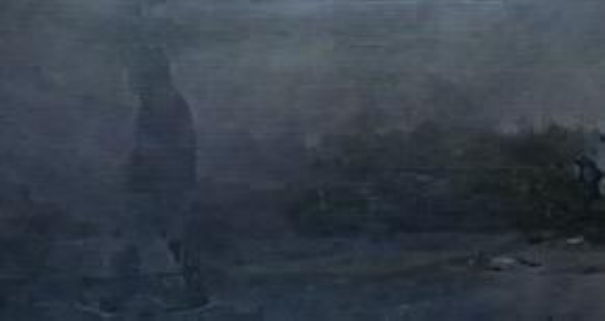}}
        \subcaption{Defogged Image}\label{fig:new2}
    \end{minipage}
    \begin{minipage}{0.32\linewidth}
        \centering
        \fcolorbox{gray}{white}{\includegraphics[width=\linewidth]{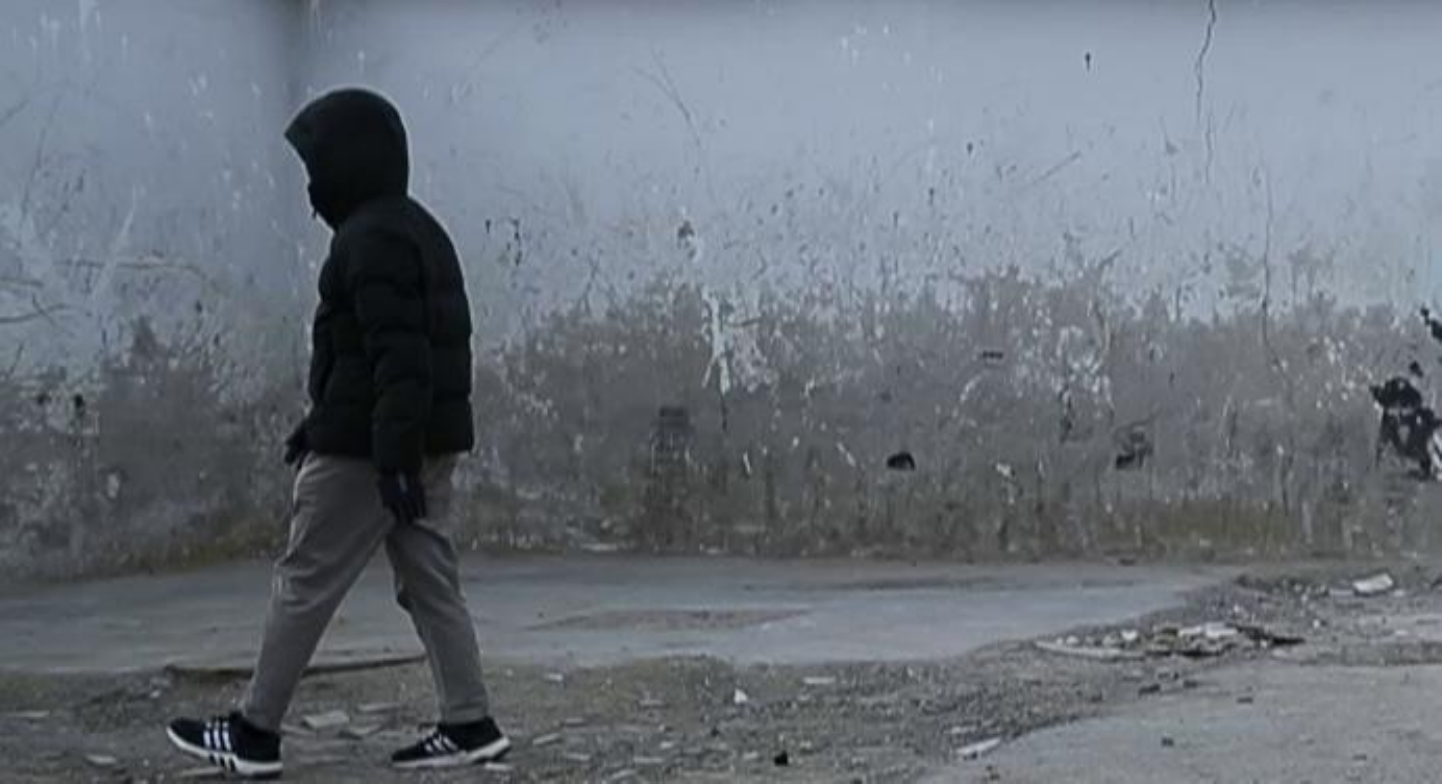}}
        \subcaption{Clean Image}\label{fig:new3}
    \end{minipage}

    \begin{minipage}{0.24\linewidth}
        \centering
        \includegraphics[width=\linewidth]{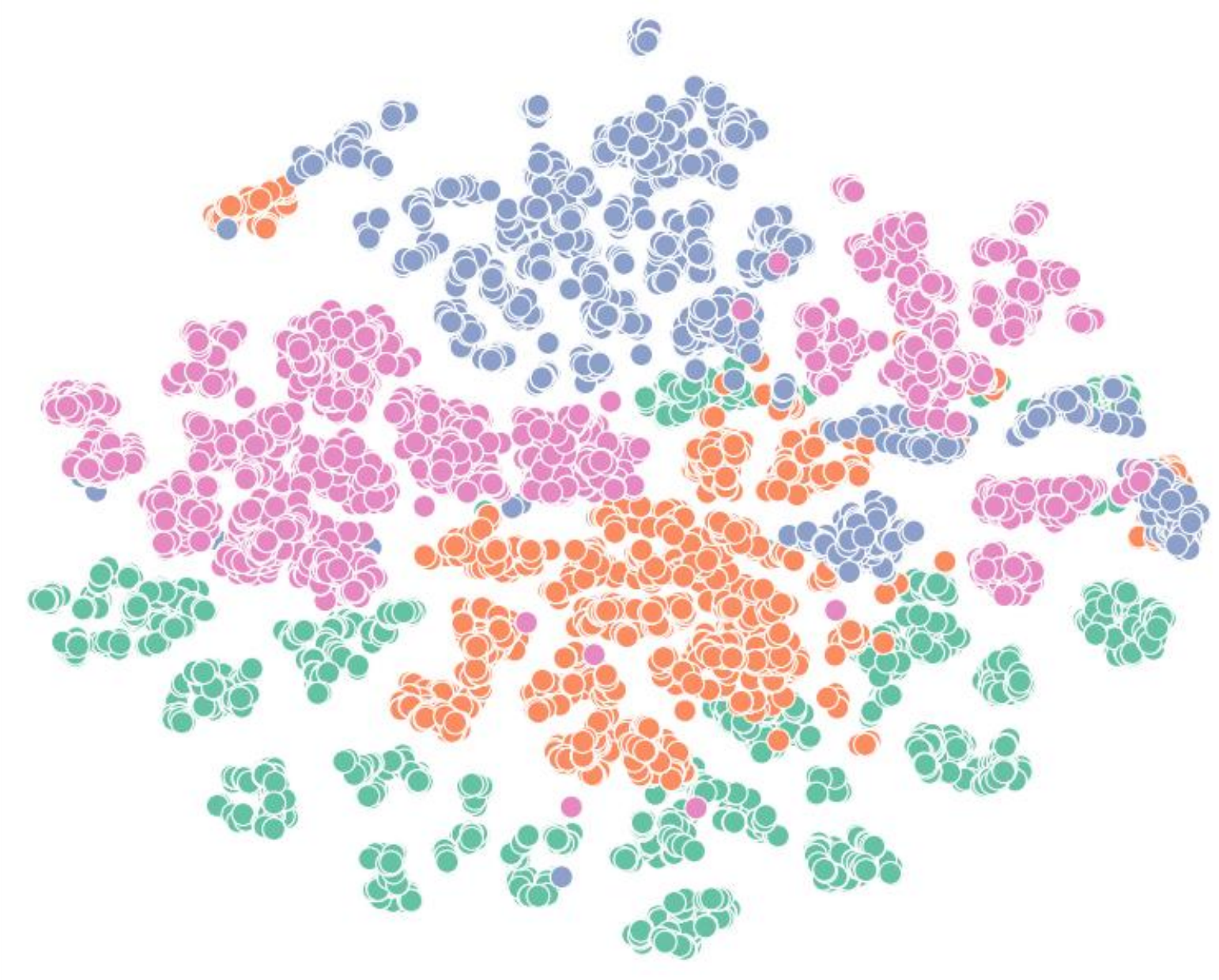}
        \subcaption{Foggy}\label{fig:fog}
    \end{minipage}
    \begin{minipage}{0.24\linewidth}
        \centering
        \includegraphics[width=\linewidth]{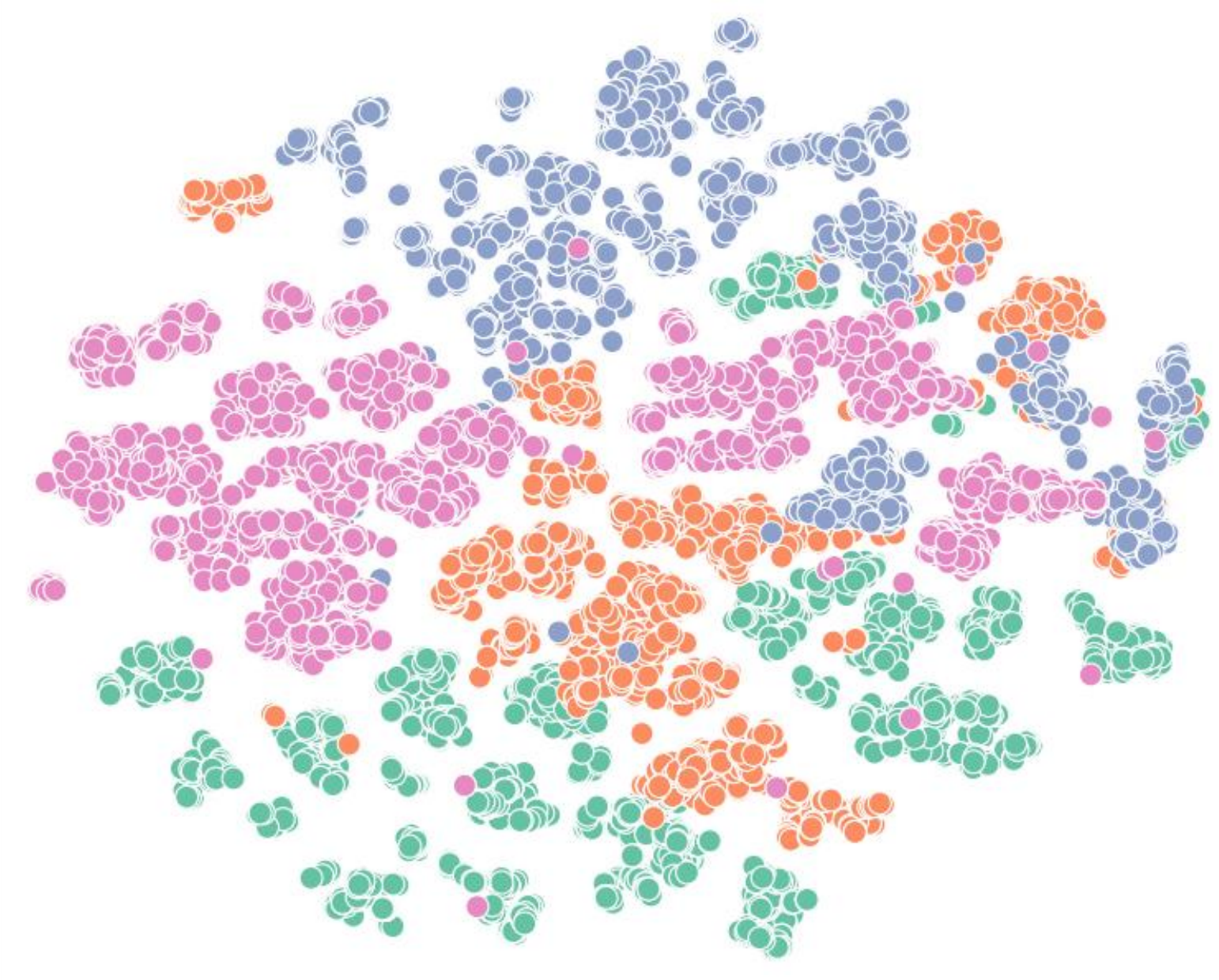}
        \subcaption{Defogged}\label{fig:dehaze}
    \end{minipage}
    \begin{minipage}{0.24\linewidth}
        \centering
        \includegraphics[width=\linewidth]{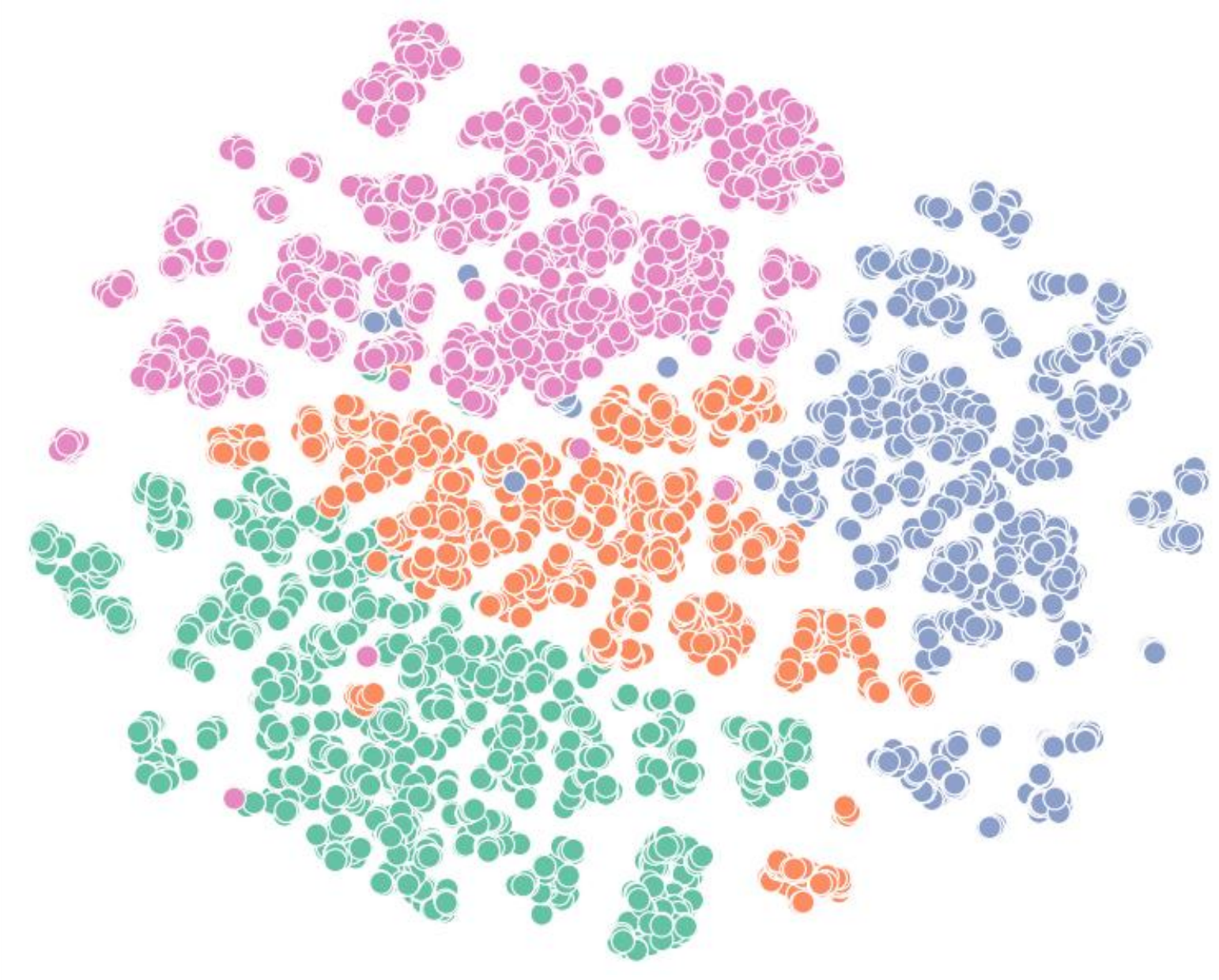}
        \subcaption{Clean}\label{fig:clean}
    \end{minipage}
    \begin{minipage}{0.24\linewidth}
        \centering
        \includegraphics[width=\linewidth]{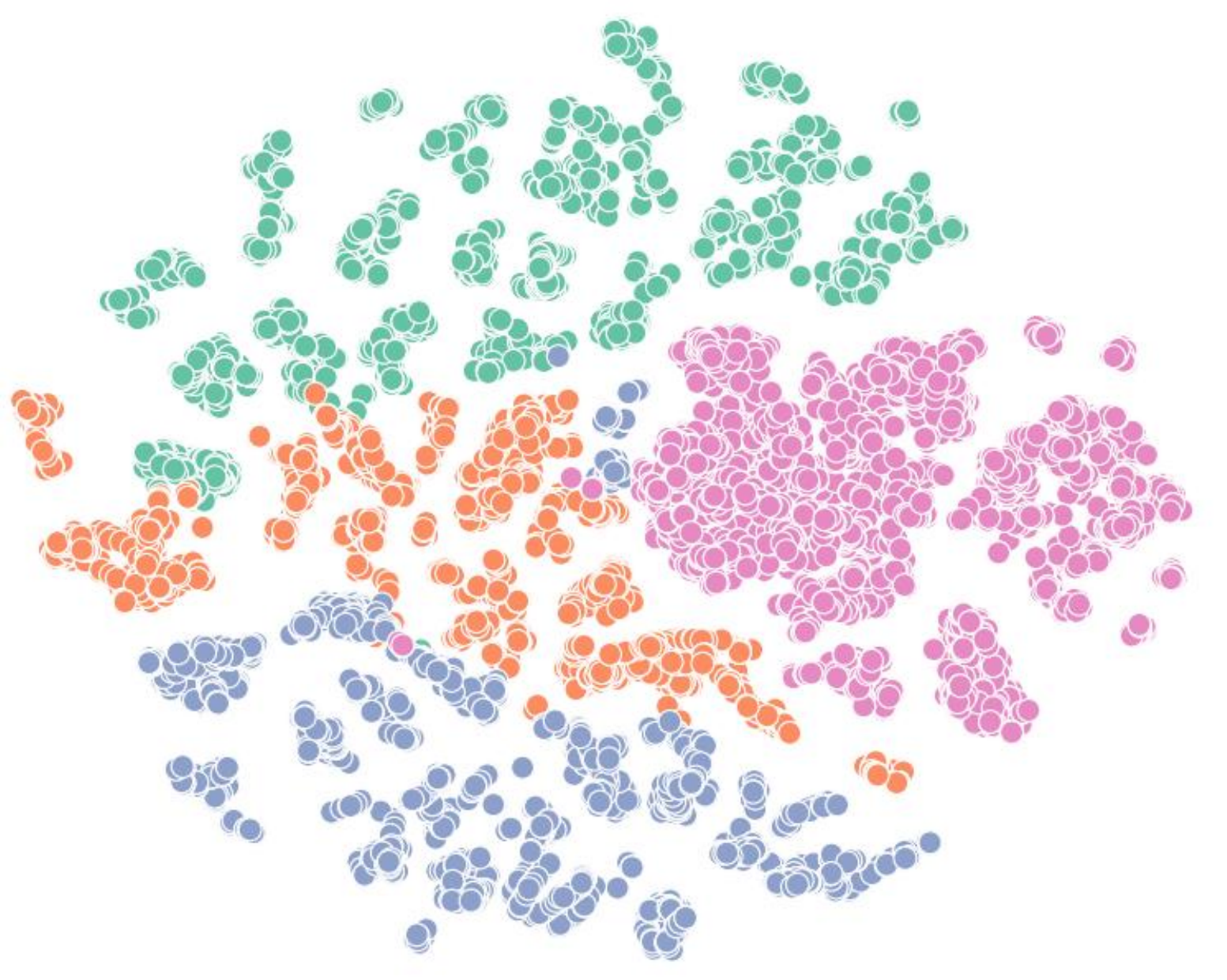}
        \subcaption{Ours}\label{fig:ours}
    \end{minipage}

    \caption{Comparison of foggy, defogged~\cite{chen2024prompt}, and clean images in FogAct (top row), and corresponding feature distributions (bottom row). The SOTA defogging result still shows residual fog and halo artifacts. Features are extracted via CLIP and visualized using t-SNE. Our learned embeddings are more aligned with clean images, while defogged features show larger intra-class variation and blurred class boundaries.}

    \label{fig:1}
\end{figure}

Existing methods for foggy action recognition \cite{tanneru2021action, chaudhary2018tsnet} follow a two-stage framework. In the first stage, defogging modules are applied to restore relatively clean videos, which are then fed into the second action classification stage. However, these frameworks perform poorly on real-world foggy data as the defogging modules yield suboptimal restorations, ultimately degrading action recognition performance. As shown in Fig.~\ref{fig:new2}, the defogged example exhibits residual fog and halo artifacts. These issues stem from defogging modules relying on synthetic training data, limiting their performance in handling diverse foggy patterns and environmental variations in real-world conditions. 
To the best of our knowledge, no real foggy dataset featuring dynamic scenes for action recognition currently exists.

\begin{table*}[t]
        \hspace*{-0.7cm}
		\centering
        \setlength{\fboxsep}{0pt}
        \captionsetup{type=figure}
			\begin{tabular}{cccccccccccccccc}
            & 
            \multicolumn{2}{c}{\makebox[2.2cm][c]{\colorbox{gray!20}{\parbox[c][0.6cm][c]{2.8cm}{\centering\small Light Examples}}}
            }&
            \multicolumn{2}{c}{\makebox[2.2cm][c]{\colorbox{gray!20}{\parbox[c][0.6cm][c]{2.8cm}{\centering\small Dense Examples}}}
            }& & & & & & &
            \\
            \vspace{0.2cm}
            \hspace{-0.0cm}\raisebox{0.38cm}{\parbox[c]{1.2cm}{\centering\small\textbf{\shortstack{Dribble\\basketball}}}} &
            \fcolorbox{gray}{white}{\includegraphics[width=1.4cm,height=1.03cm]{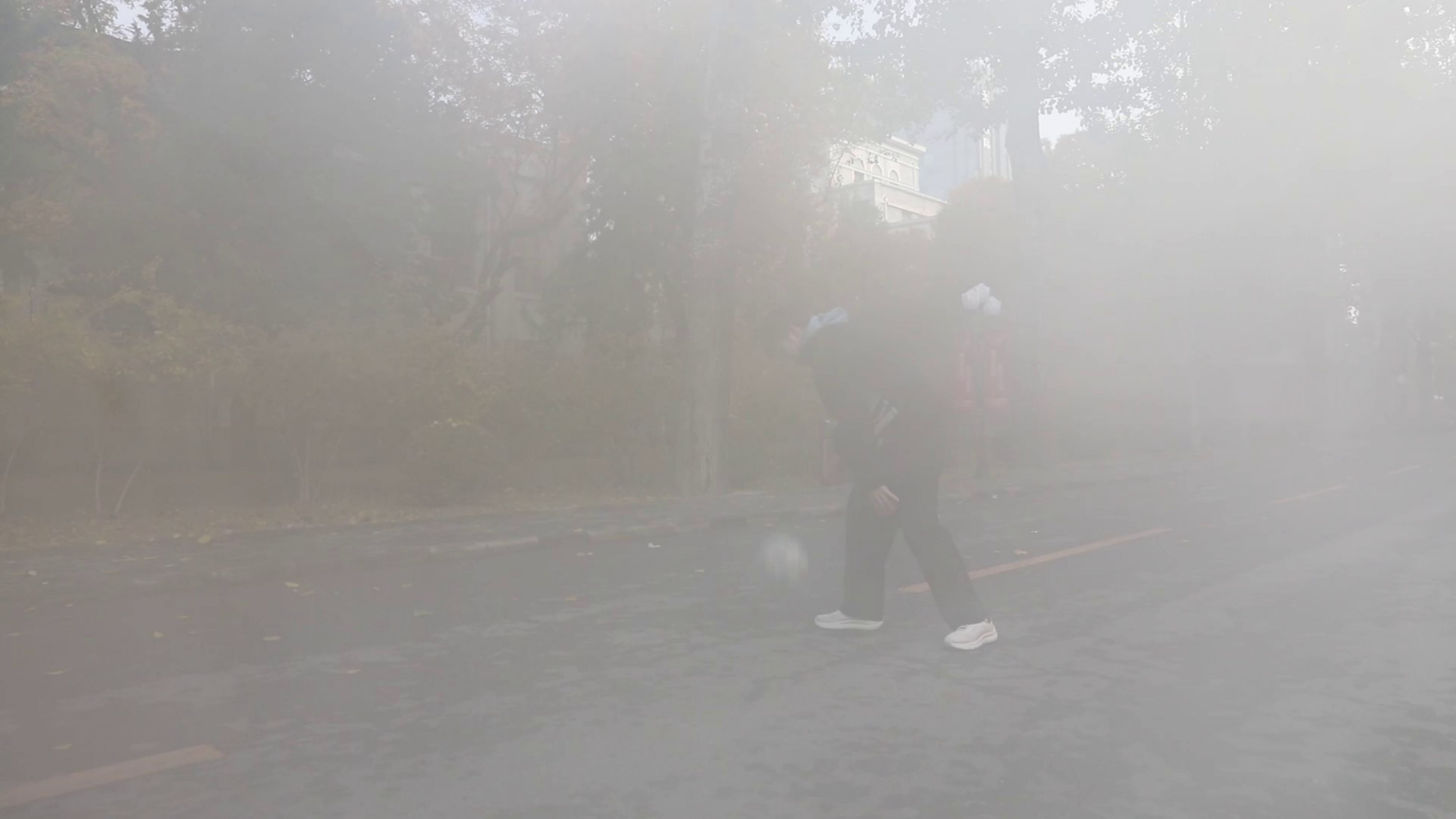}} &
            \hspace{-0.42cm}\fcolorbox{gray}{white}{\includegraphics[width=1.4cm,height=1.03cm]{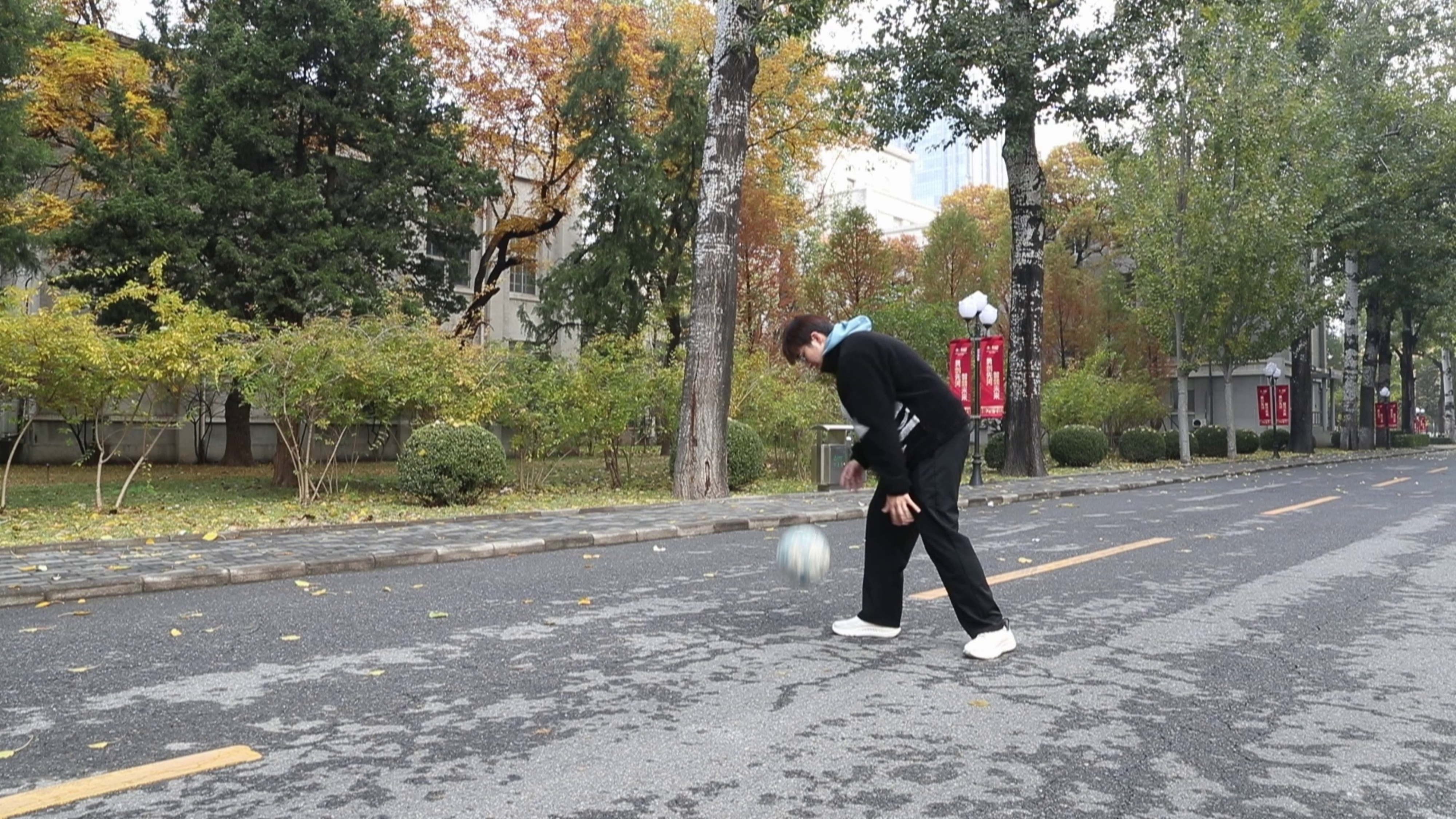}} & 
            {\setlength\fboxrule{1.5pt}\fcolorbox{red}{white}{\includegraphics[width=1.4cm,height=1.03cm]{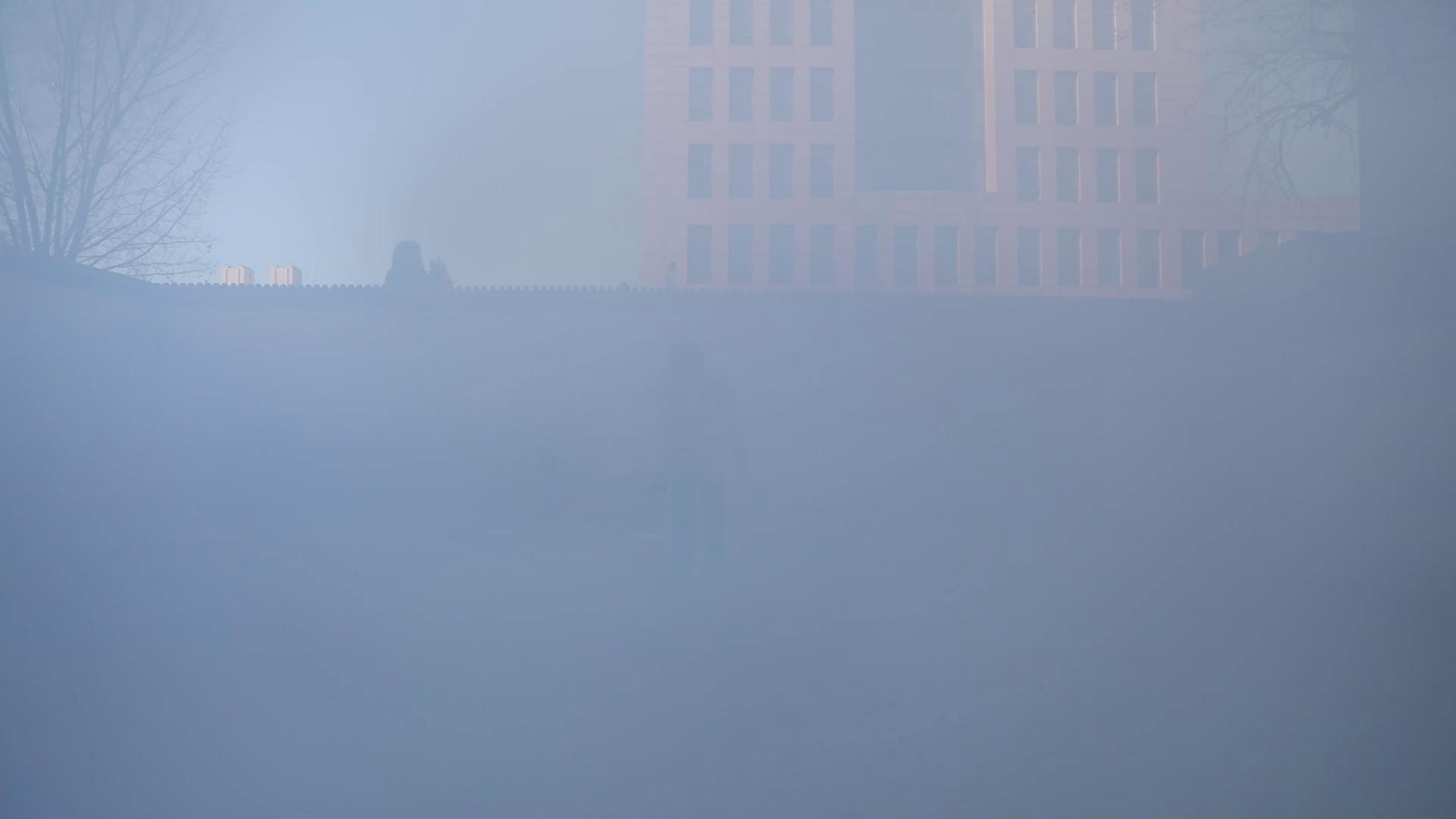}}} &  
            \hspace{-0.5cm}{\setlength\fboxrule{1.5pt}\fcolorbox{red}{white}{\includegraphics[width=1.4cm,height=1.03cm]{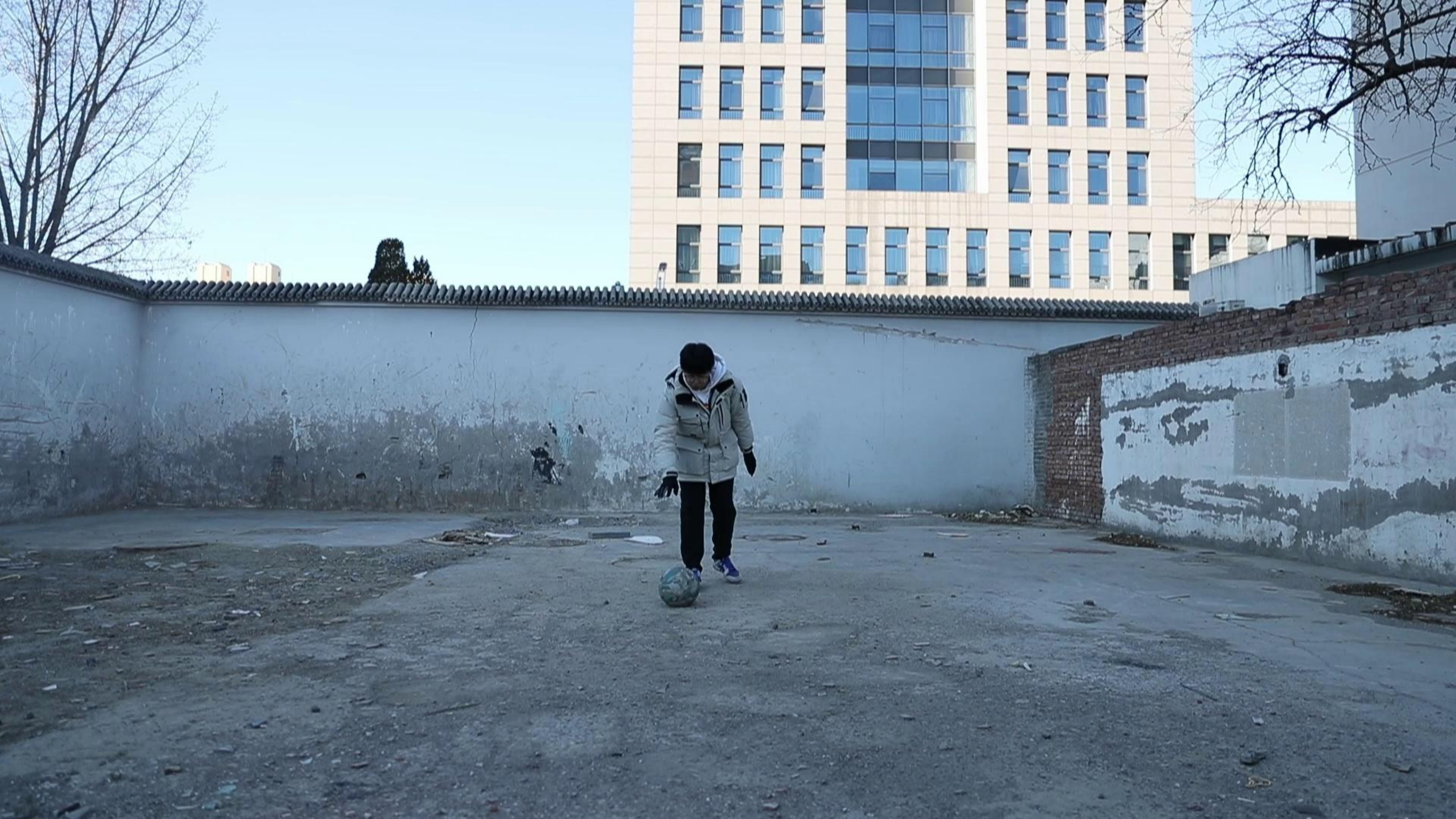}}} &
            \hspace{-0.0cm}\raisebox{0.38cm}{\parbox[c]{0.8cm}{\centering\small\textbf{Front}}} &
		      \hspace{-0.4cm}\fcolorbox{gray}{white}{\includegraphics[width=1.4cm,height=1.03cm]{samples/pictures/sup1/fog/frame_00018.pdf}} &  
            \hspace{-0.35cm}\fcolorbox{gray}{white}{\includegraphics[width=1.4cm,height=1.03cm]{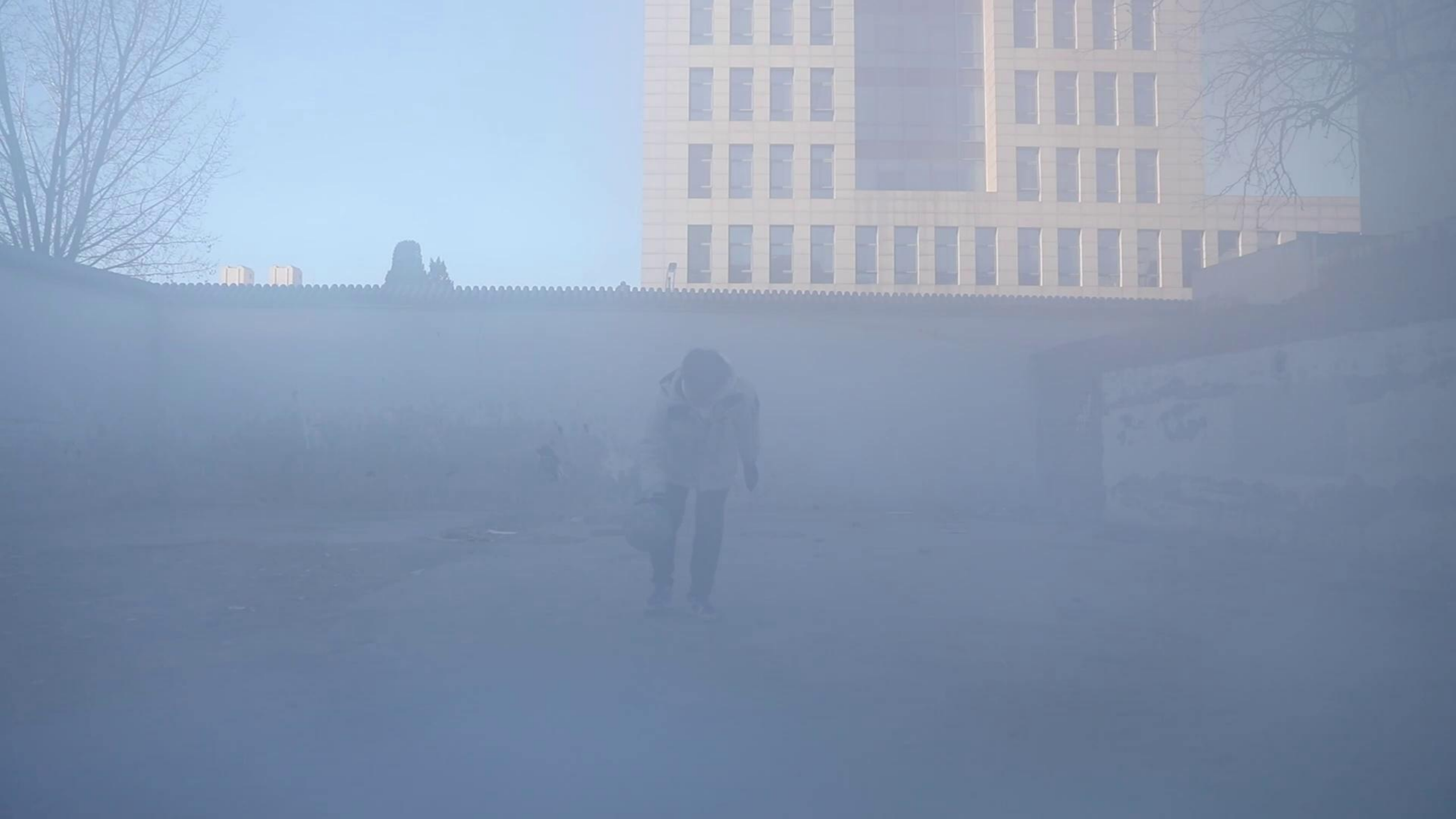}} &
            \hspace{-0.35cm}\fcolorbox{gray}{white}{\includegraphics[width=1.4cm,height=1.03cm]{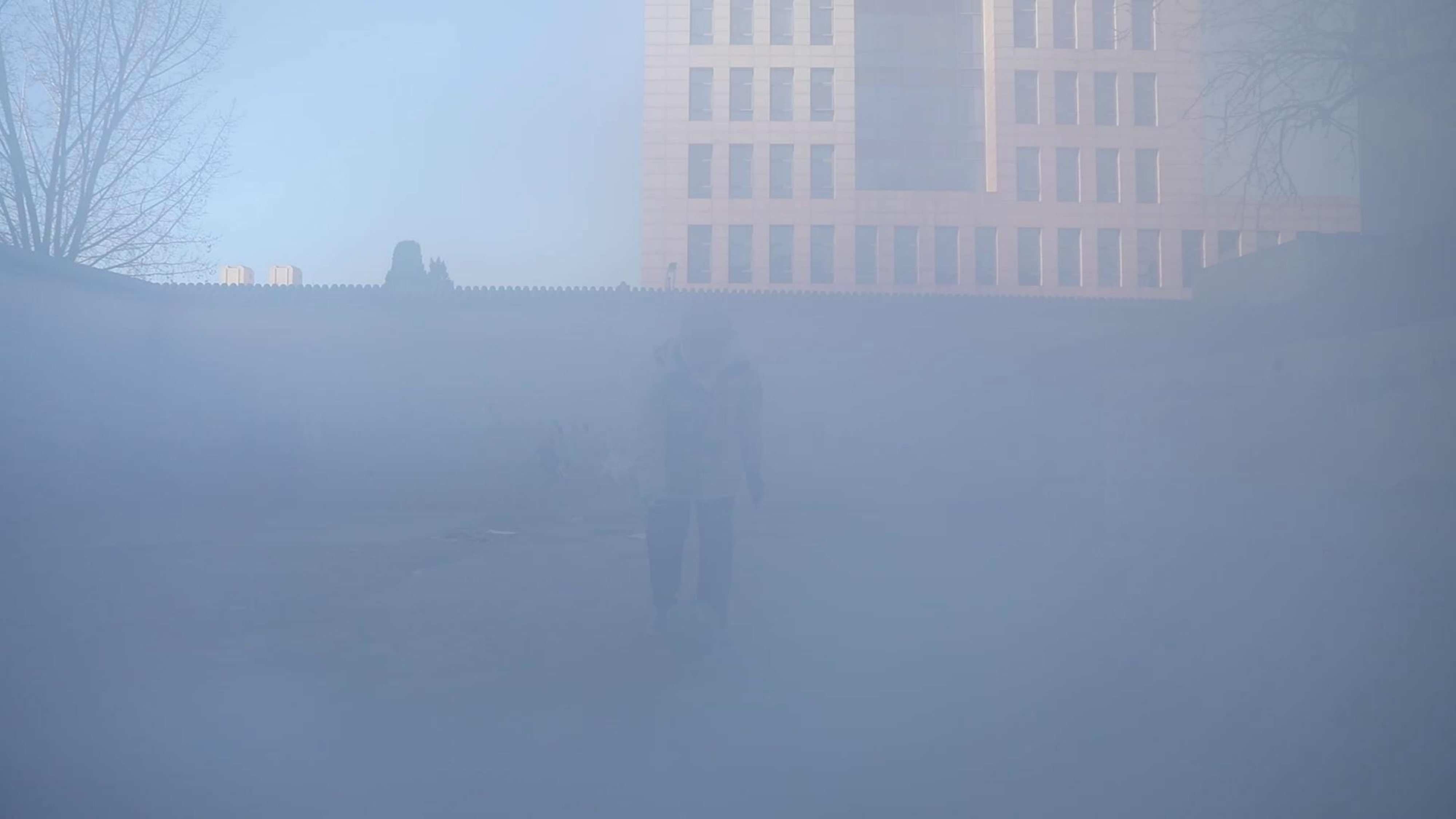}} & 
            \hspace{-0.35cm}\fcolorbox{gray}{white}{\includegraphics[width=1.4cm,height=1.03cm]{samples/pictures/sup1/clean/ball1/frame_00018.pdf}} 
            & \hspace{-0.35cm}\fcolorbox{gray}{white}{\includegraphics[width=1.4cm,height=1.03cm]{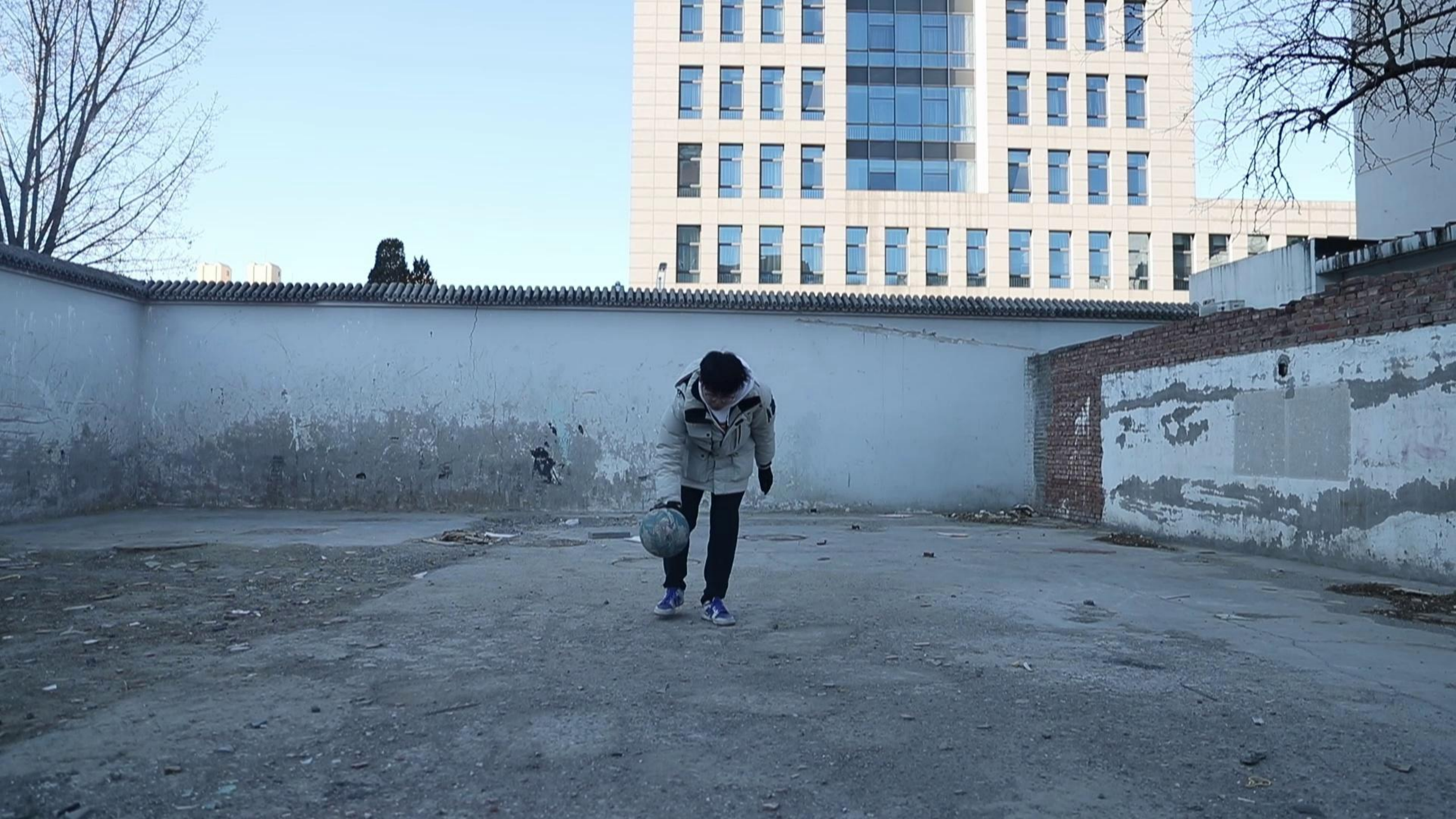}} & \hspace{-0.35cm}\fcolorbox{gray}{white}{\includegraphics[width=1.4cm,height=1.03cm]{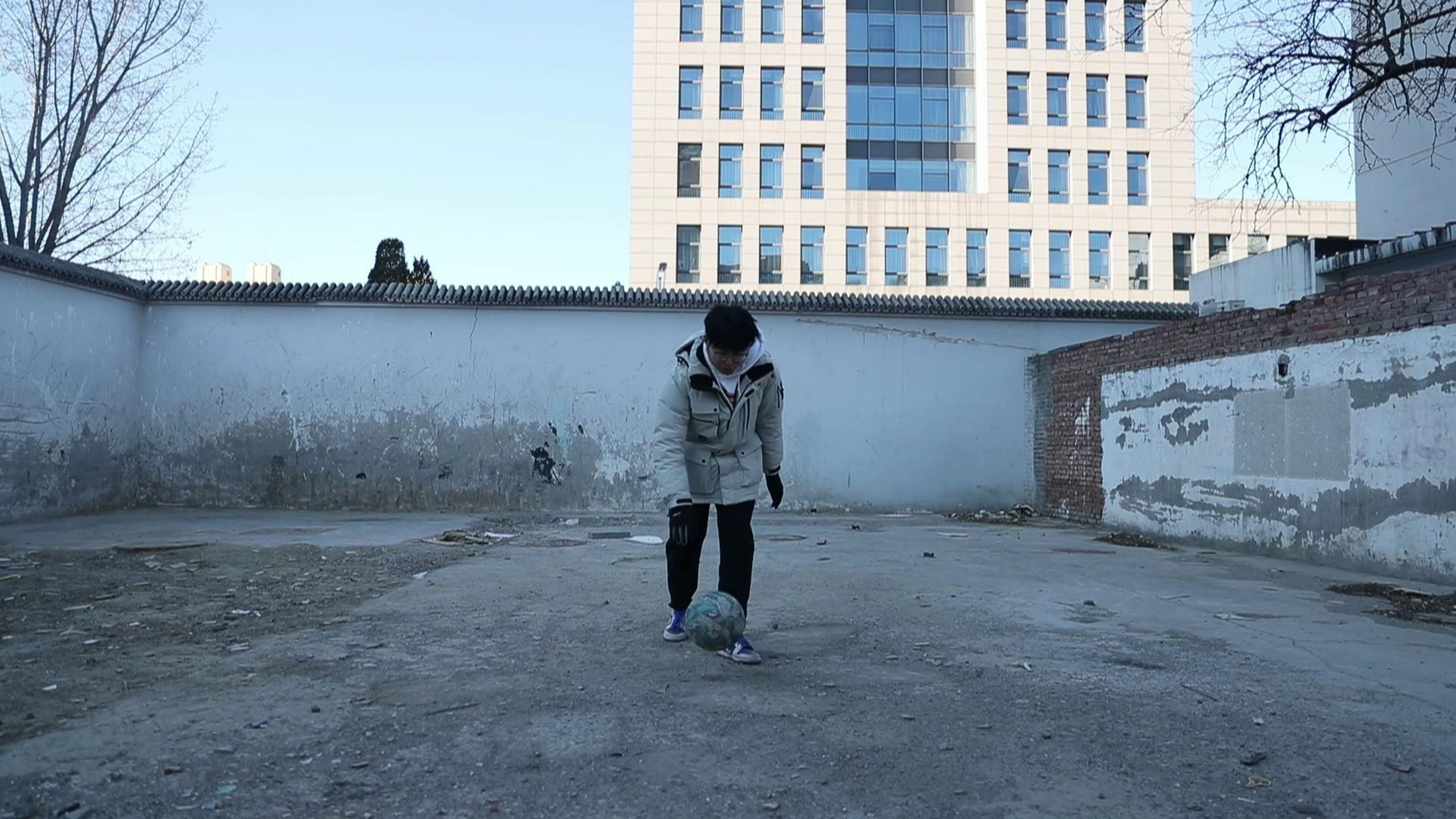}}
            \\
            \vspace{0.2cm}
            \hspace{-0.0cm}\raisebox{0.38cm}{\parbox[c]{1.2cm}{\centering\small\textbf{\shortstack{Listen\\to music}}}} &

        \fcolorbox{gray}{white}{\includegraphics[width=1.4cm,height=1.03cm]{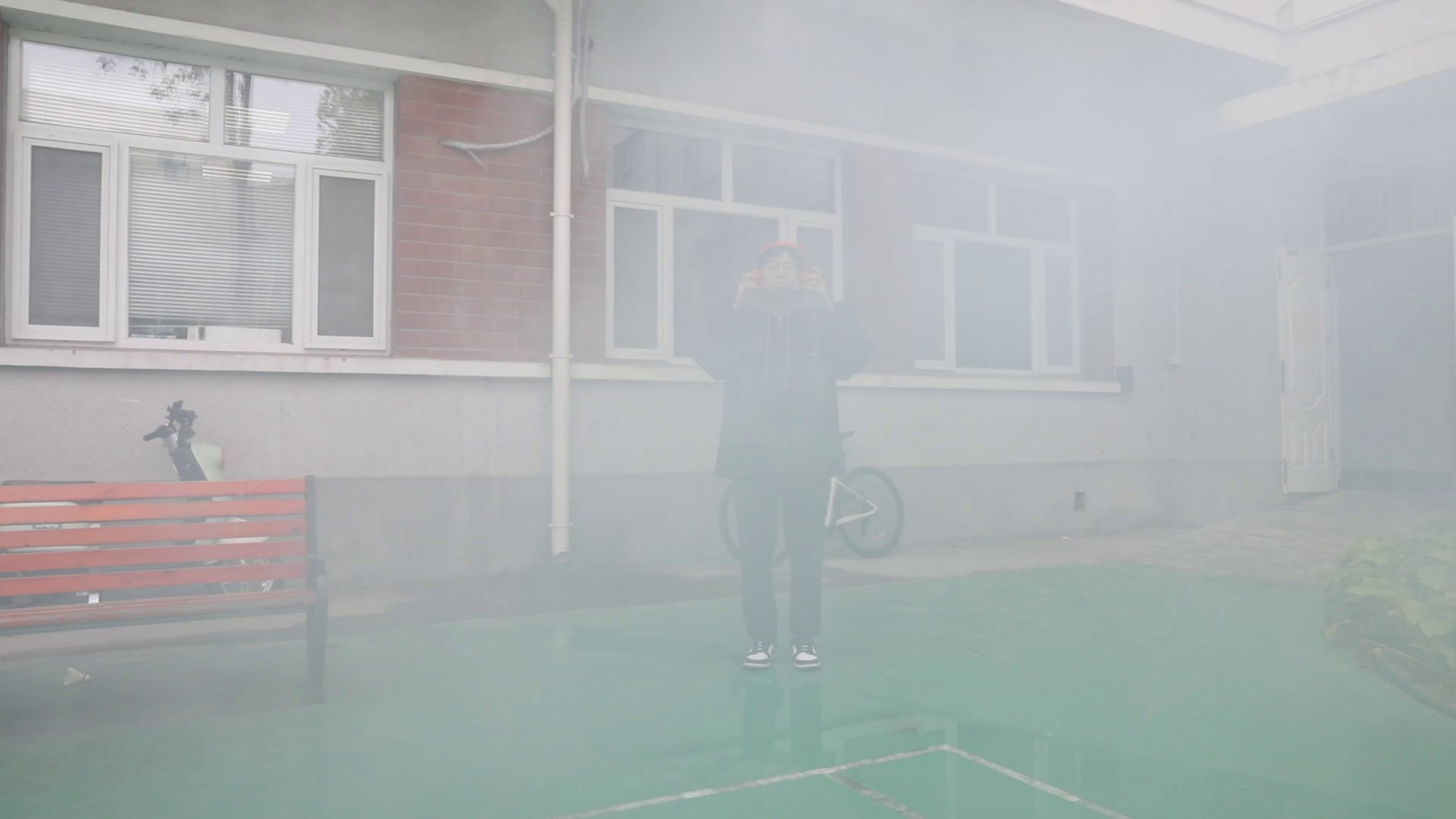}} &
        \hspace{-0.42cm}\fcolorbox{gray}{white}{\includegraphics[width=1.4cm,height=1.03cm]{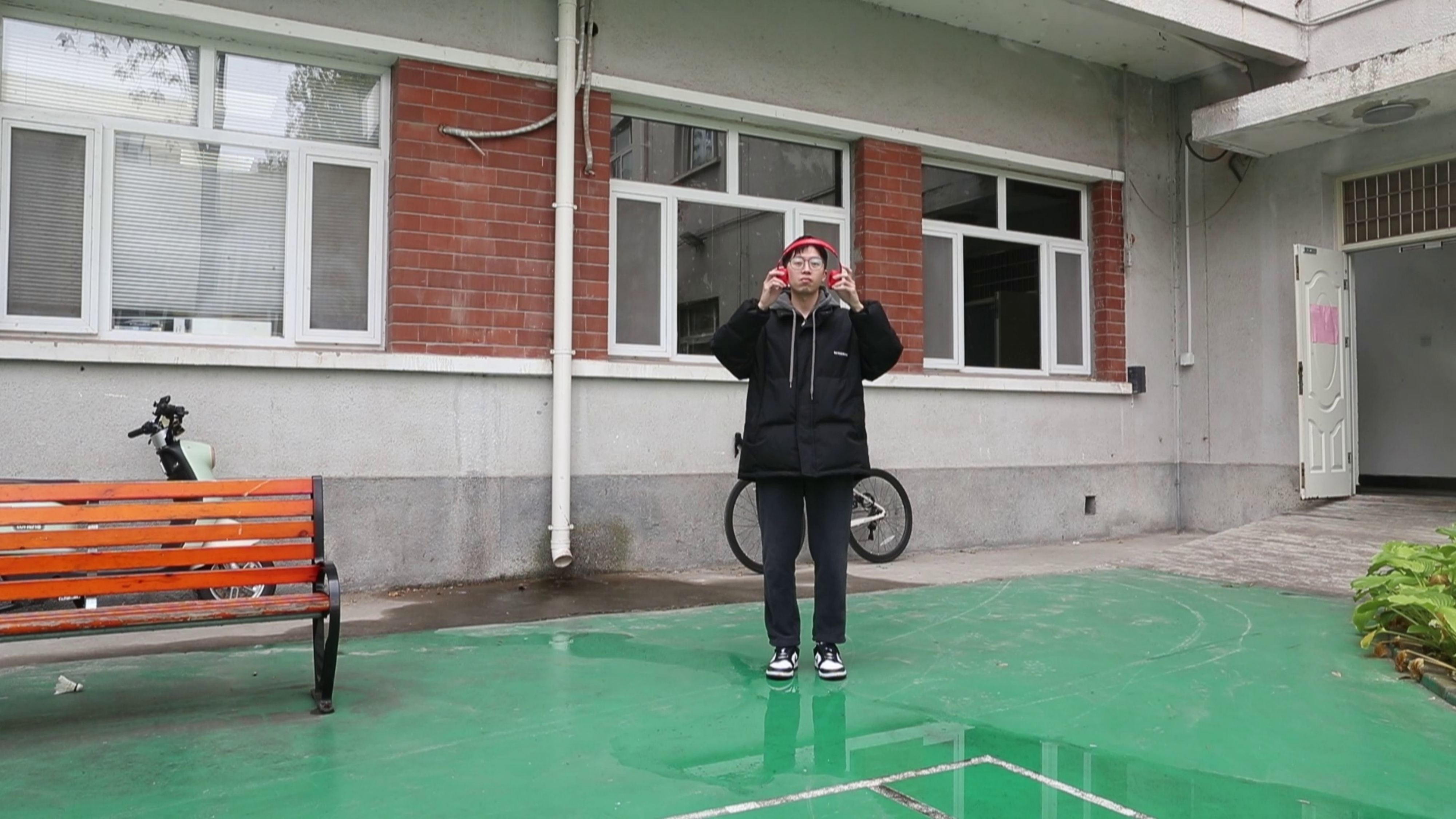}} &
        
       \fcolorbox{gray}{white}{\includegraphics[width=1.4cm,height=1.03cm]{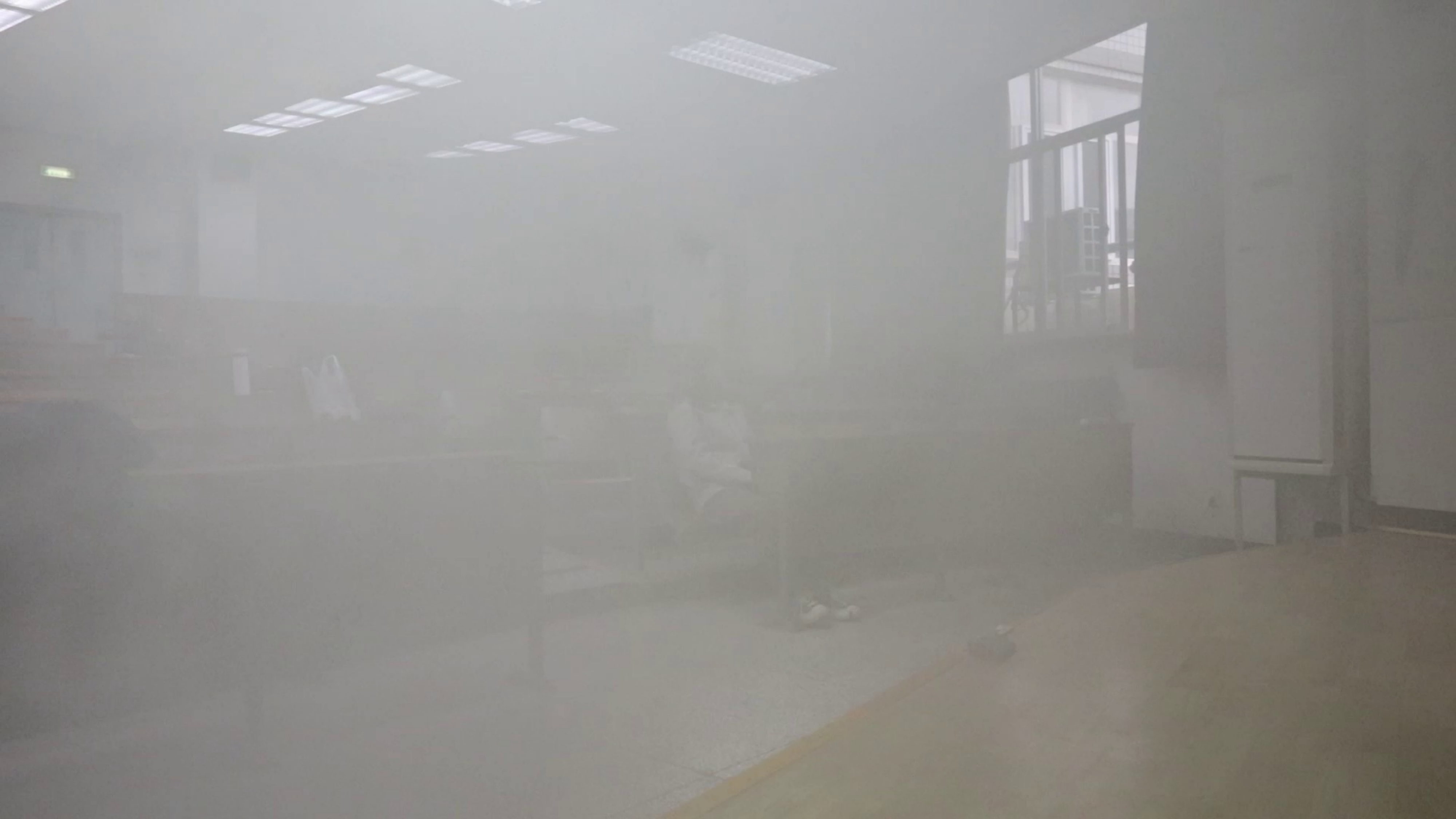}} &
        \hspace{-0.5cm}\fcolorbox{gray}{white}{\includegraphics[width=1.4cm,height=1.03cm]{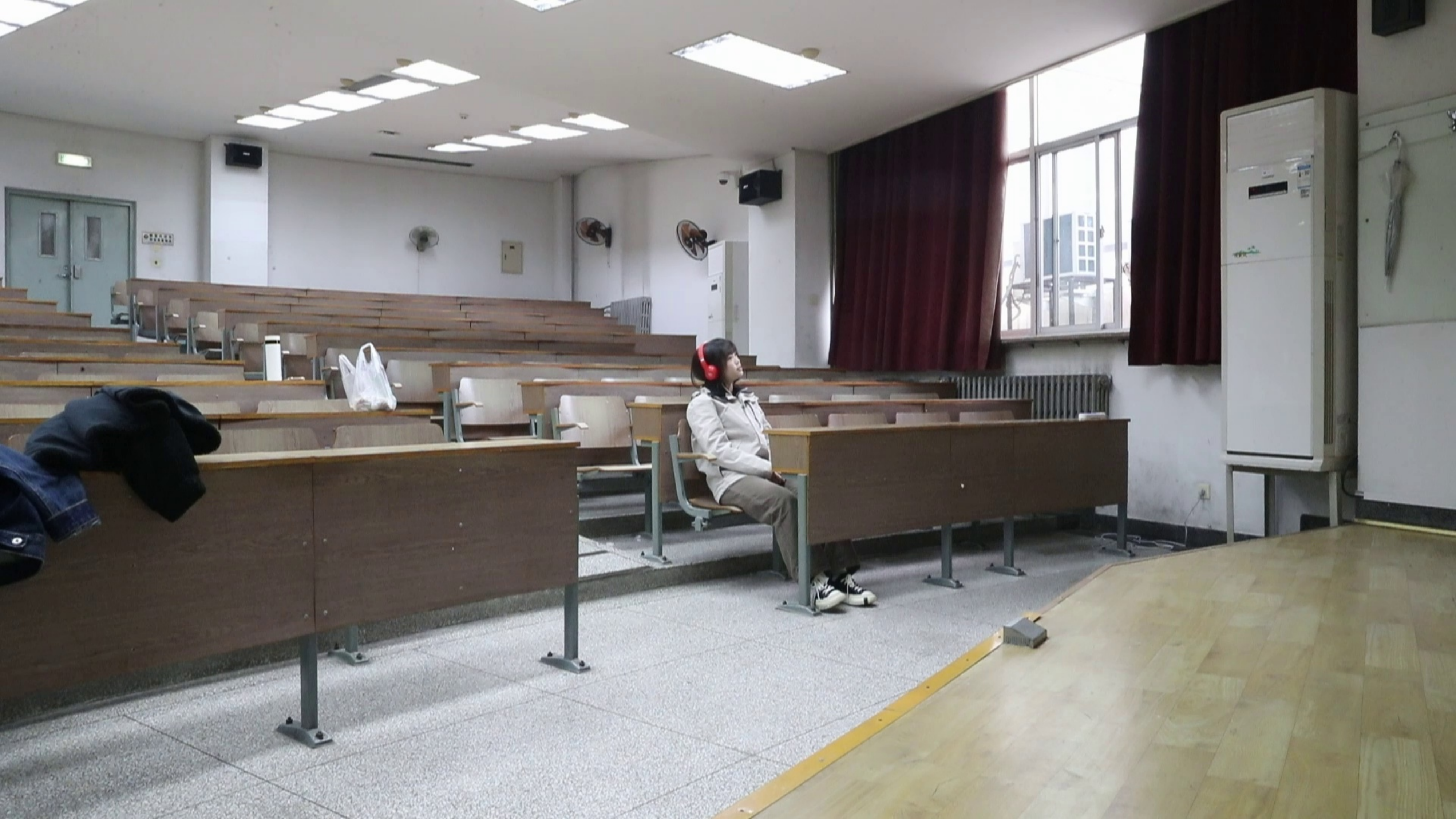}}&
            \hspace{-0.0cm}\raisebox{0.42cm}{\parbox[c]{0.8cm}{\centering\small\textbf{Back}}} &
			\hspace{-0.4cm}\fcolorbox{gray}{white}{\includegraphics[width=1.4cm,height=1.03cm]{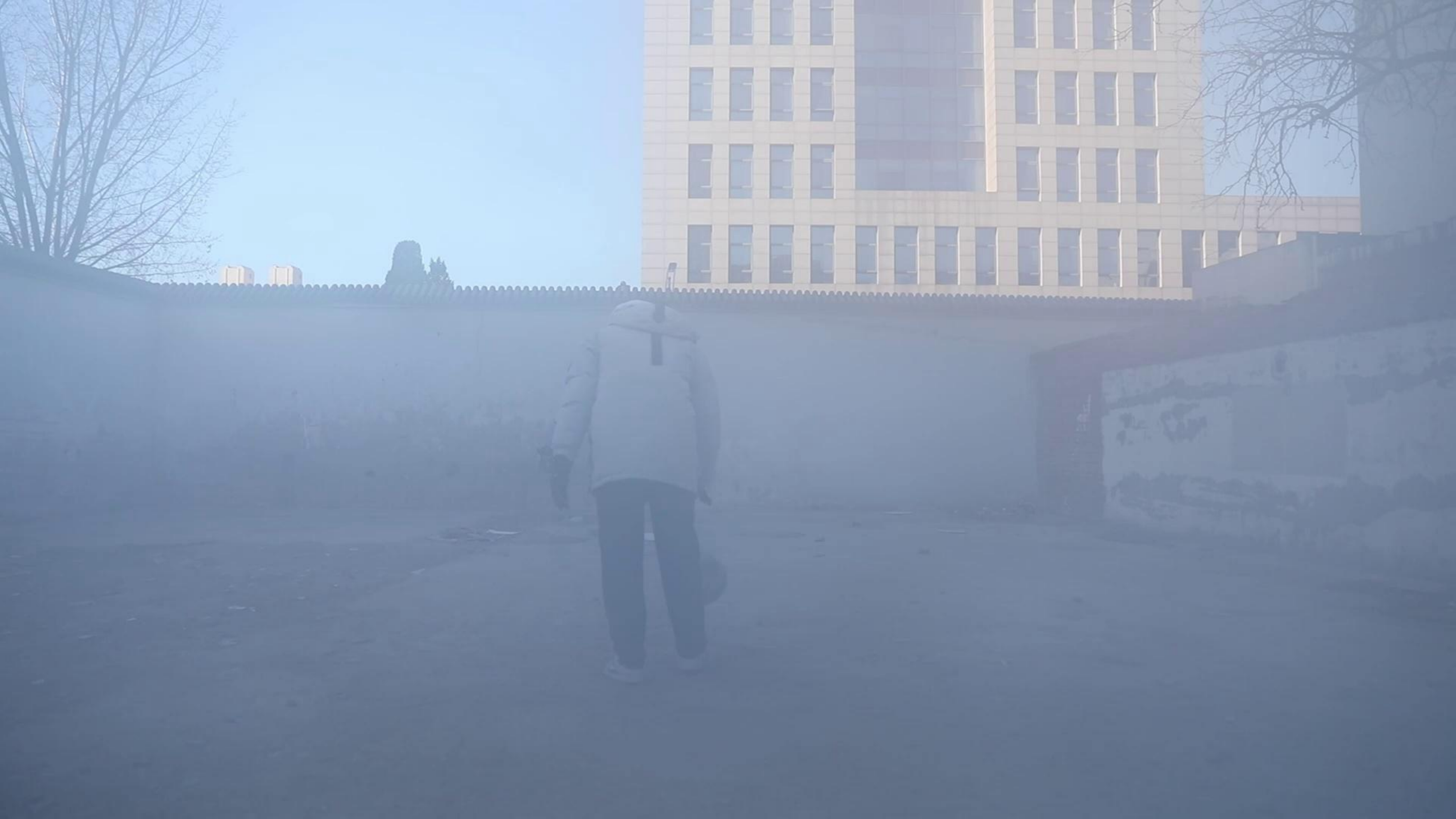}} & 
            \hspace{-0.35cm}\fcolorbox{gray}{white}{\includegraphics[width=1.4cm,height=1.03cm]{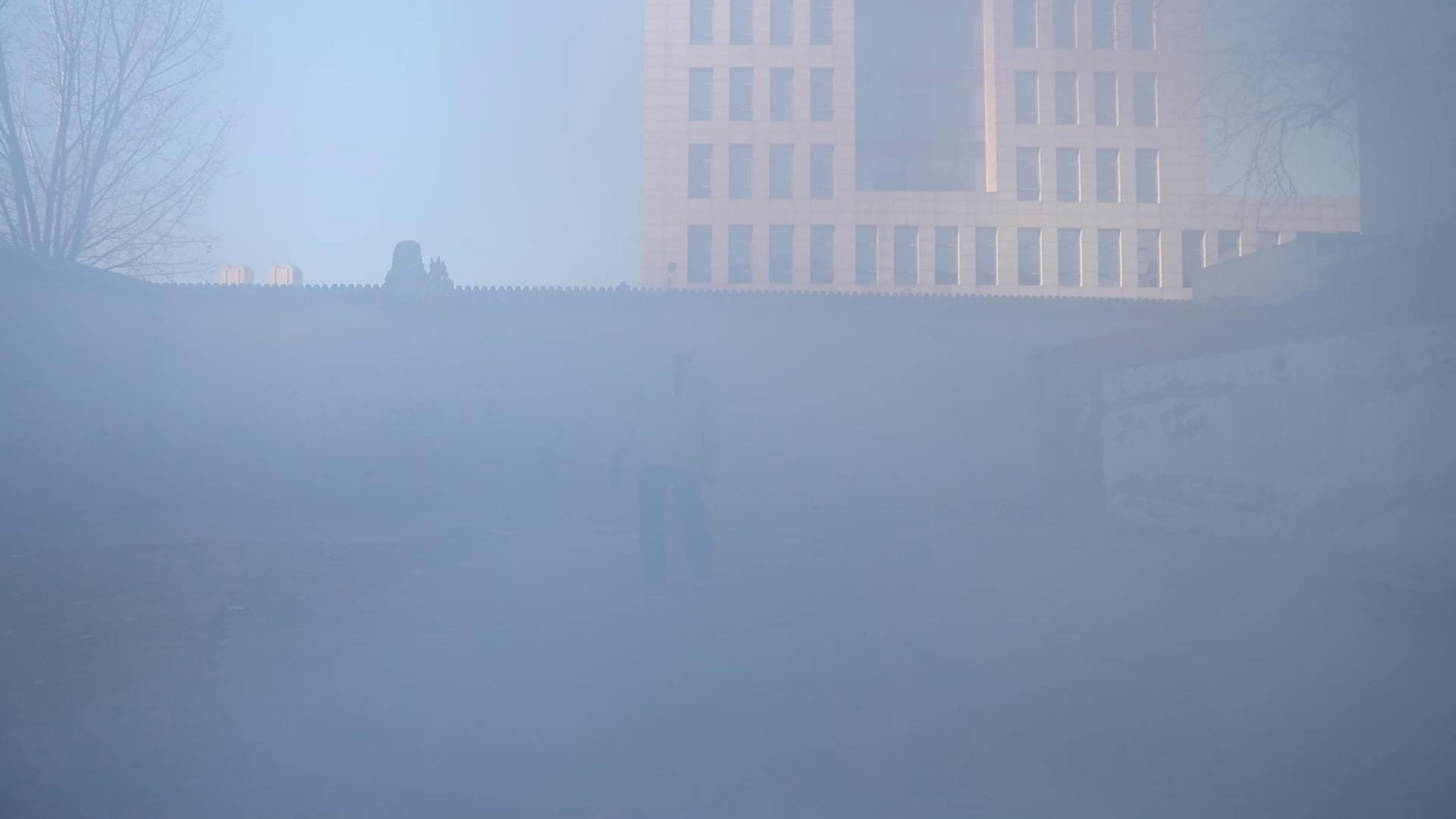}} &
            \hspace{-0.35cm}\fcolorbox{gray}{white}{\includegraphics[width=1.4cm,height=1.03cm]{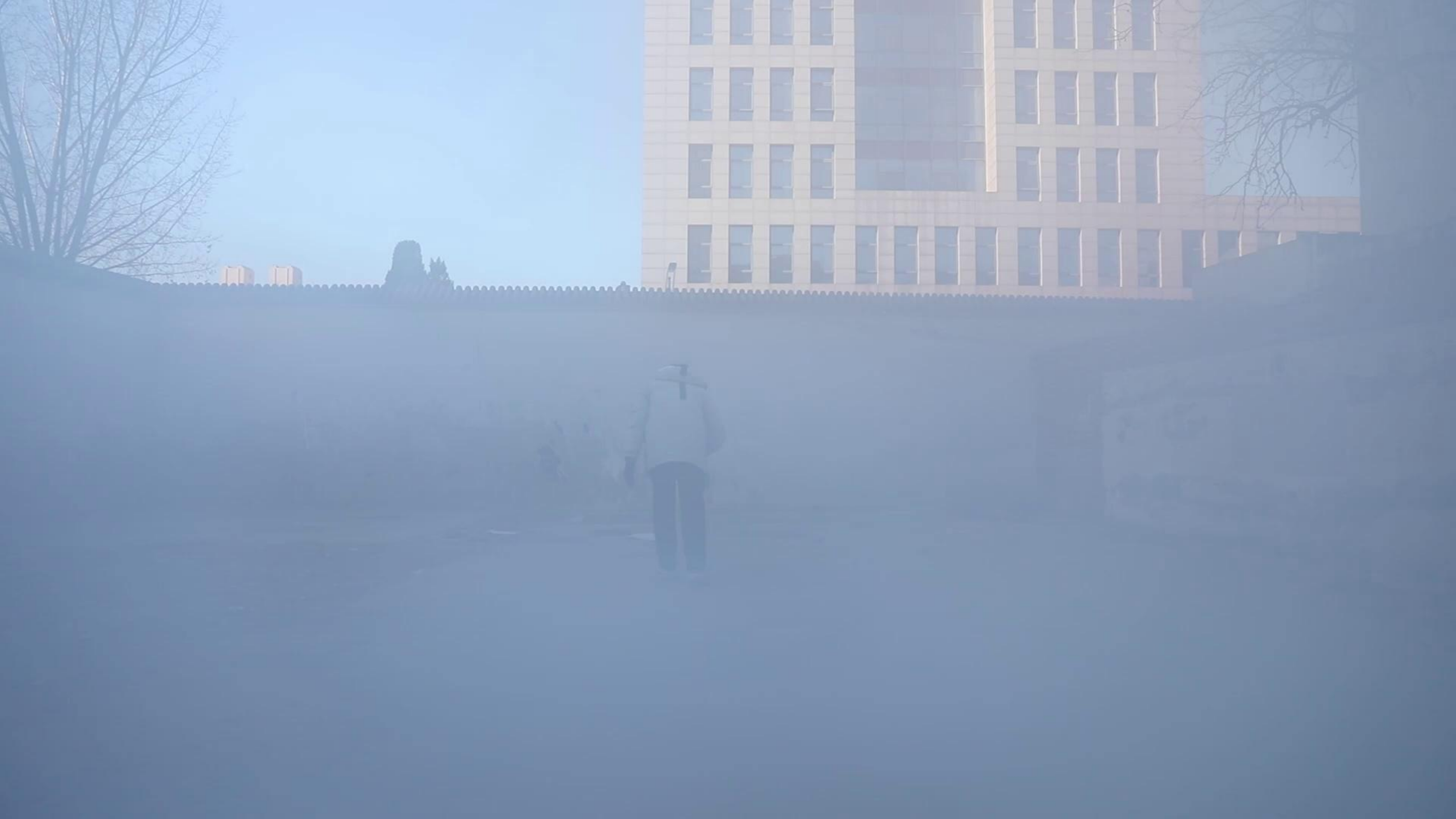}} & 
            \hspace{-0.35cm}\fcolorbox{gray}{white}{\includegraphics[width=1.4cm,height=1.03cm]
            {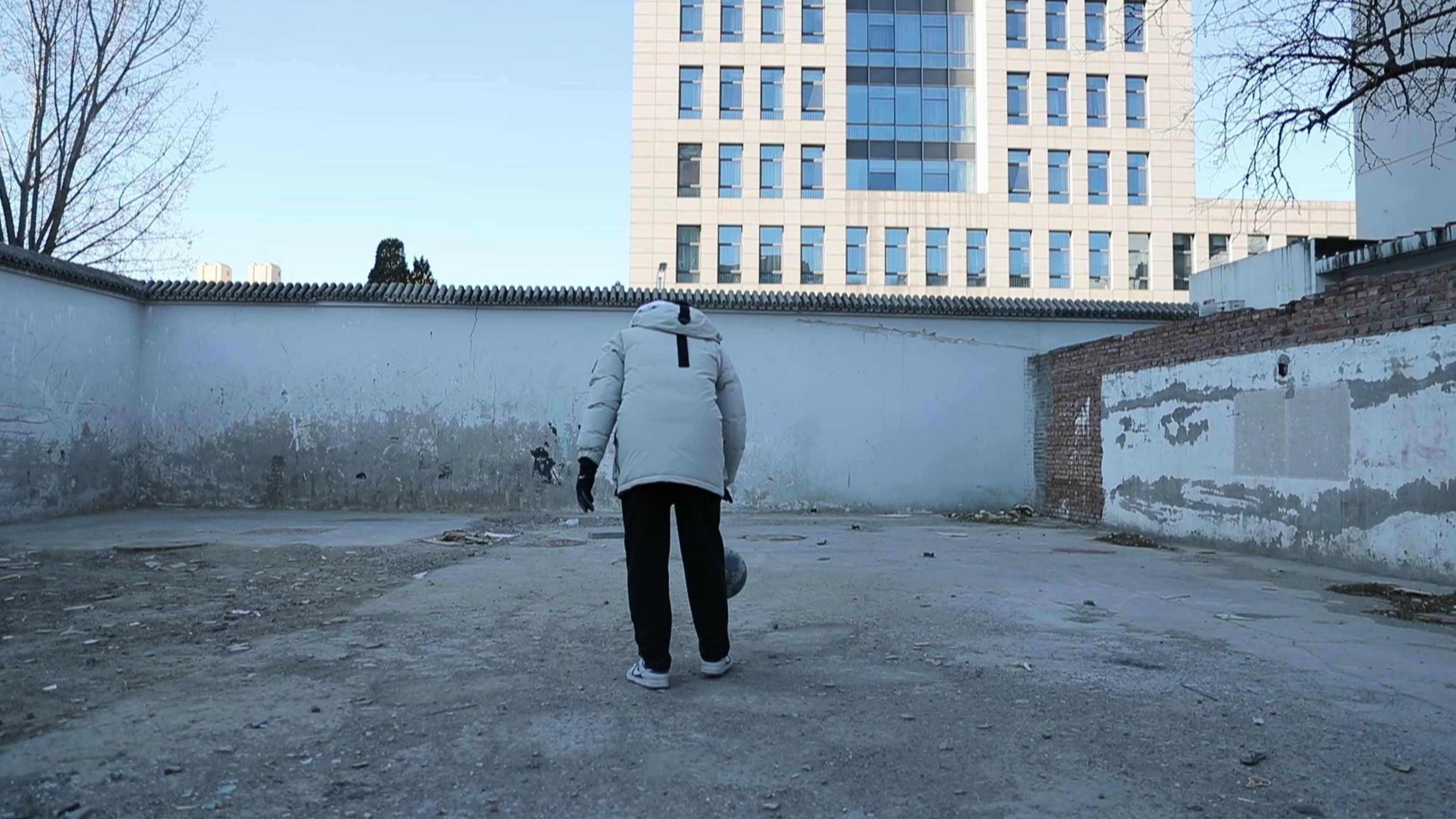}} 
            & \hspace{-0.35cm}\fcolorbox{gray}{white}{\includegraphics[width=1.4cm,height=1.03cm]{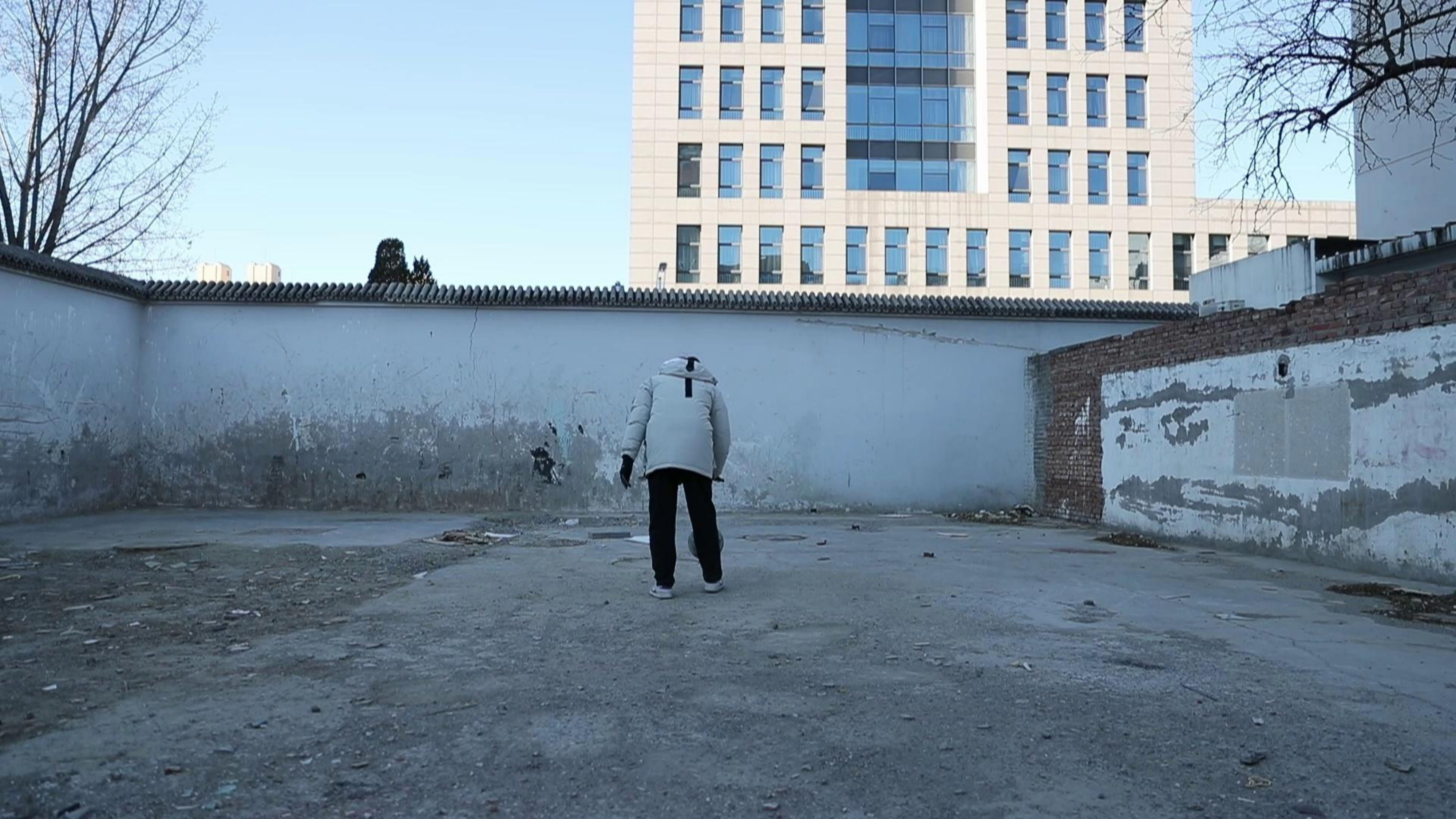}} & \hspace{-0.35cm}\fcolorbox{gray}{white}{\includegraphics[width=1.4cm,height=1.03cm]{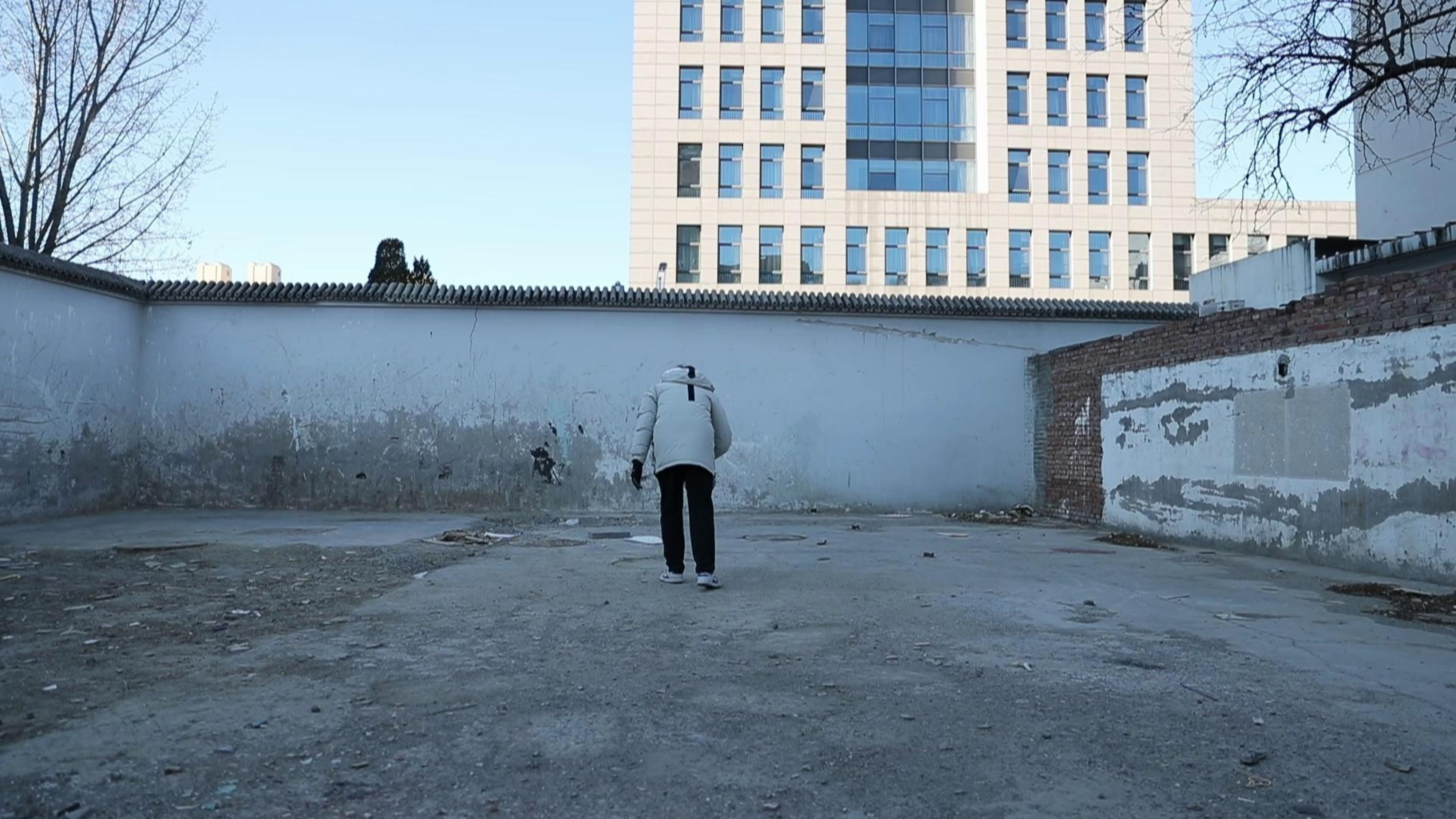}} 

            \\
             \vspace{0.2cm}
             \hspace{-0.0cm}\raisebox{0.38cm}{\parbox[c]{1.2cm}{\centering\small\textbf{Wave}}} &
            \fcolorbox{gray}{white}{\includegraphics[width=1.4cm,height=1.03cm]{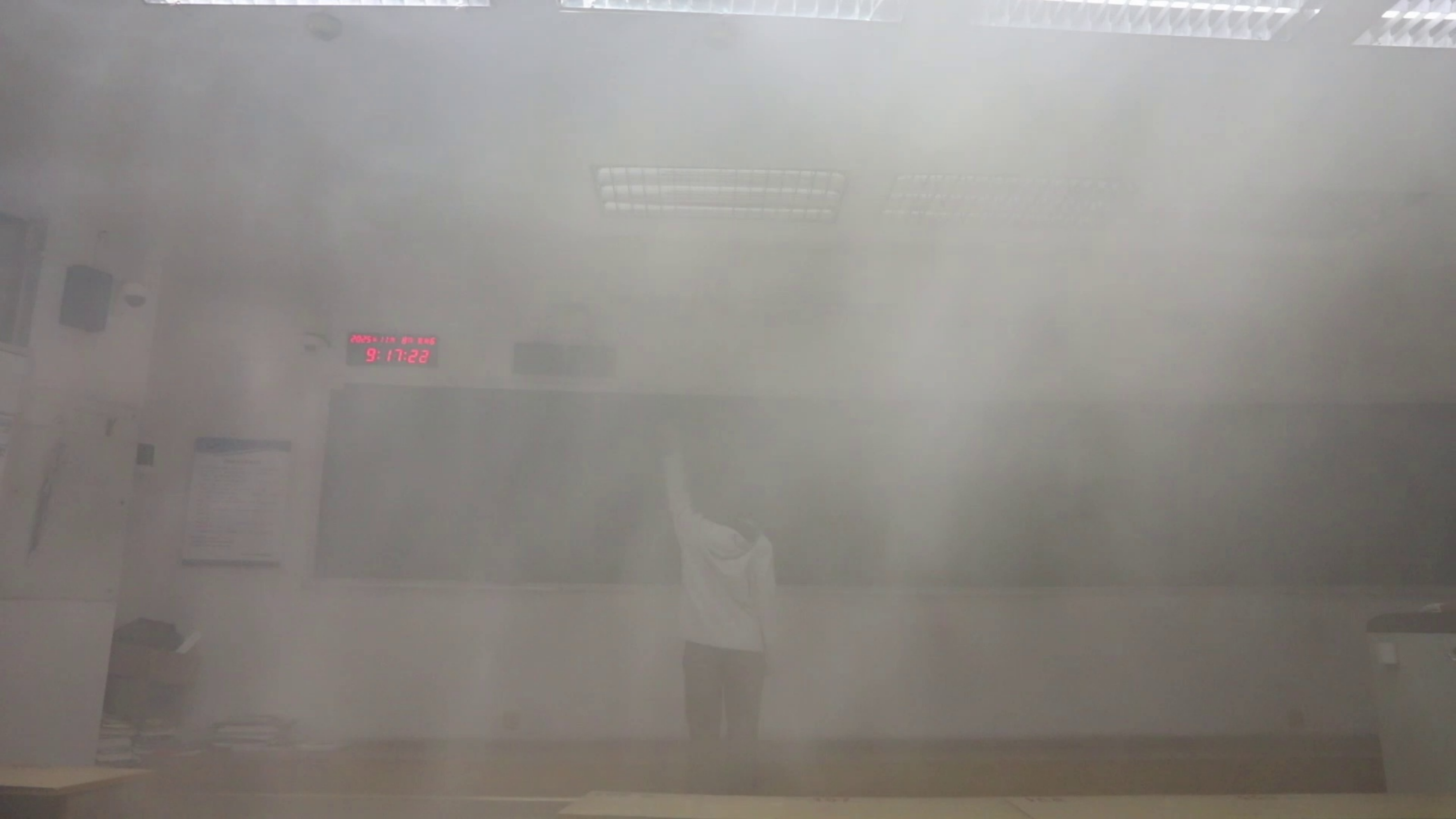}} &
            \hspace{-0.42cm}\fcolorbox{gray}{white}{\includegraphics[width=1.4cm,height=1.03cm]{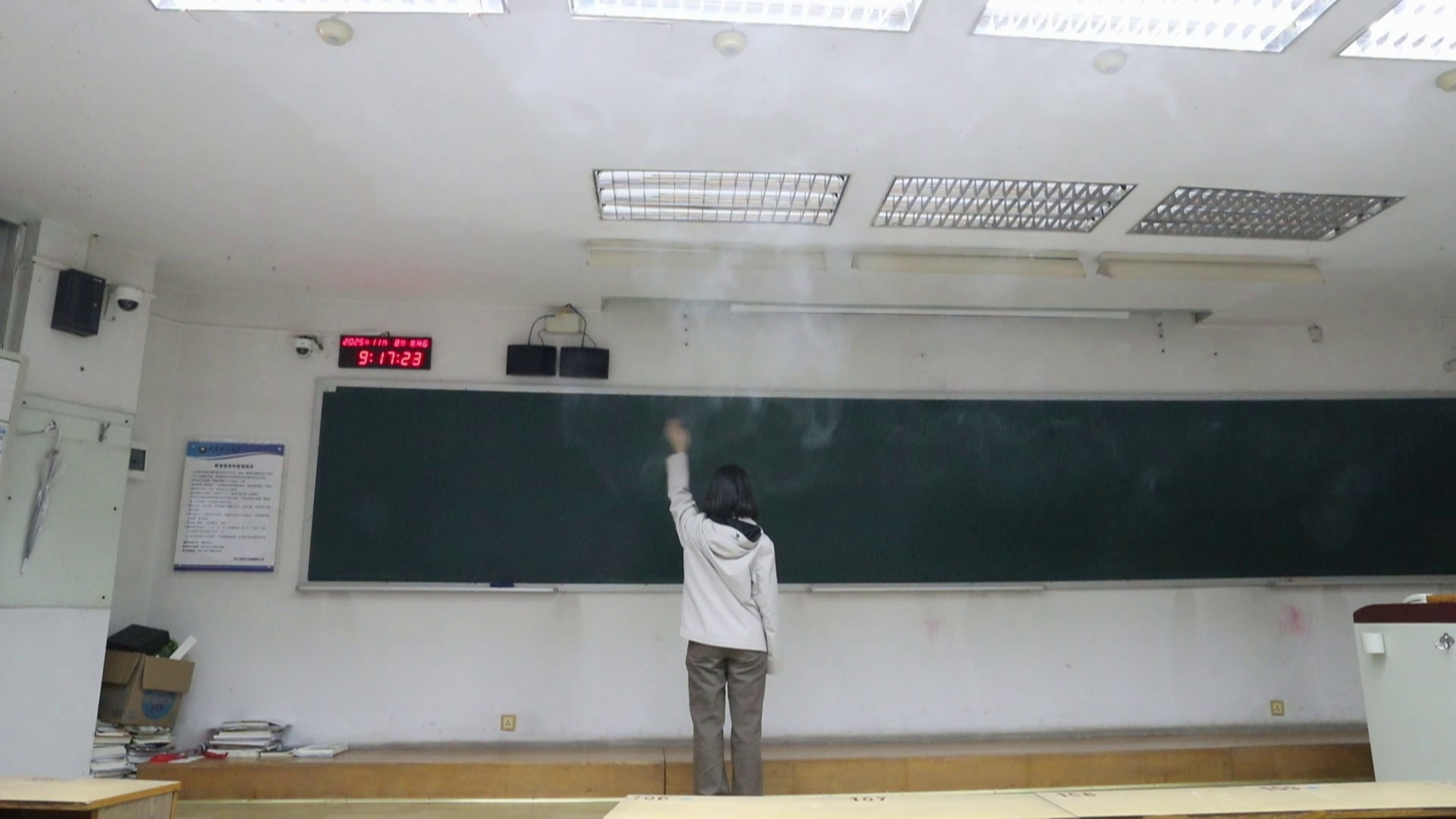}}&  
            \fcolorbox{gray}{white}{\includegraphics[width=1.4cm,height=1.03cm]{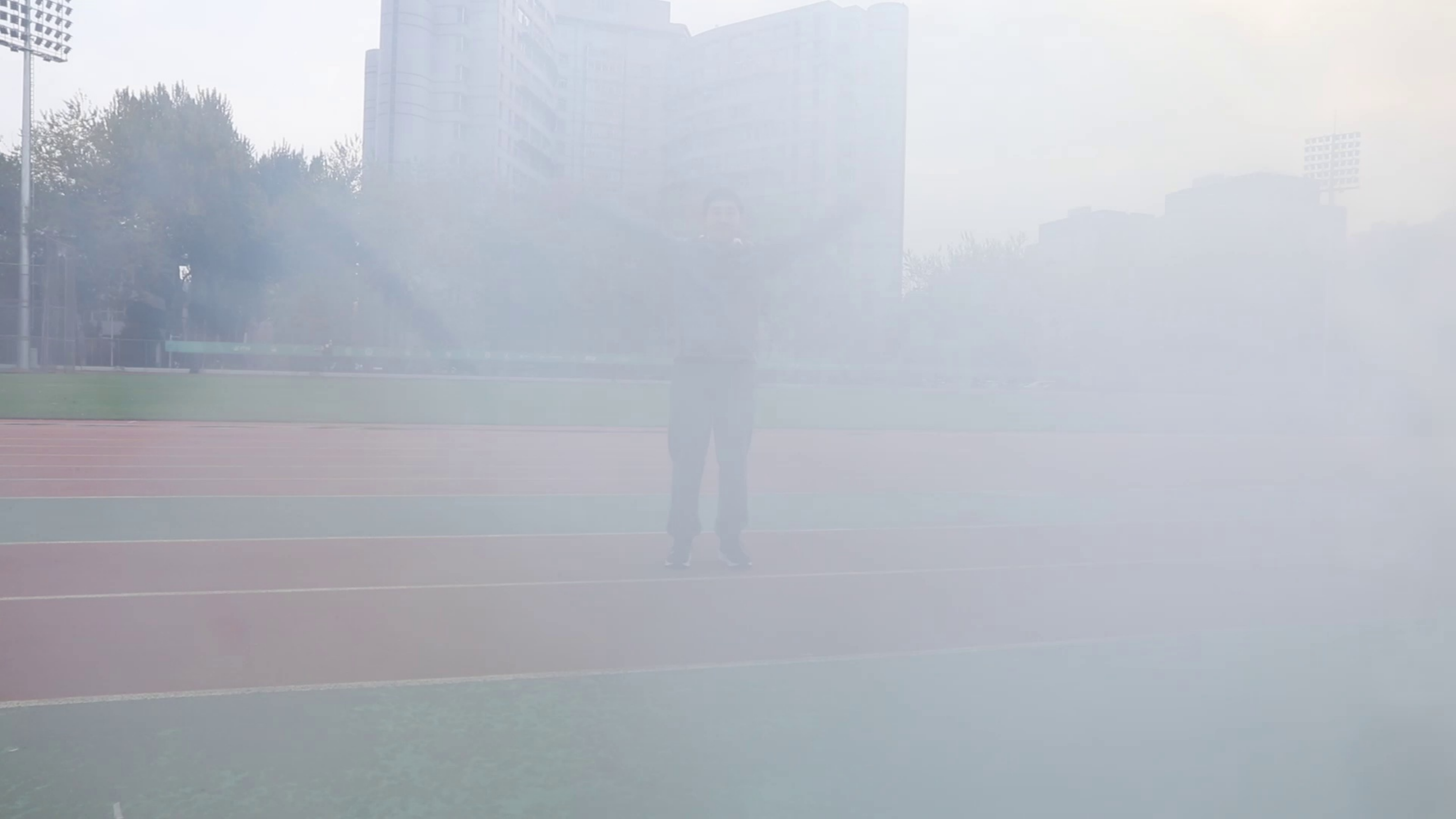}} &  \hspace{-0.5cm}\fcolorbox{gray}{white}{\includegraphics[width=1.4cm,height=1.03cm]{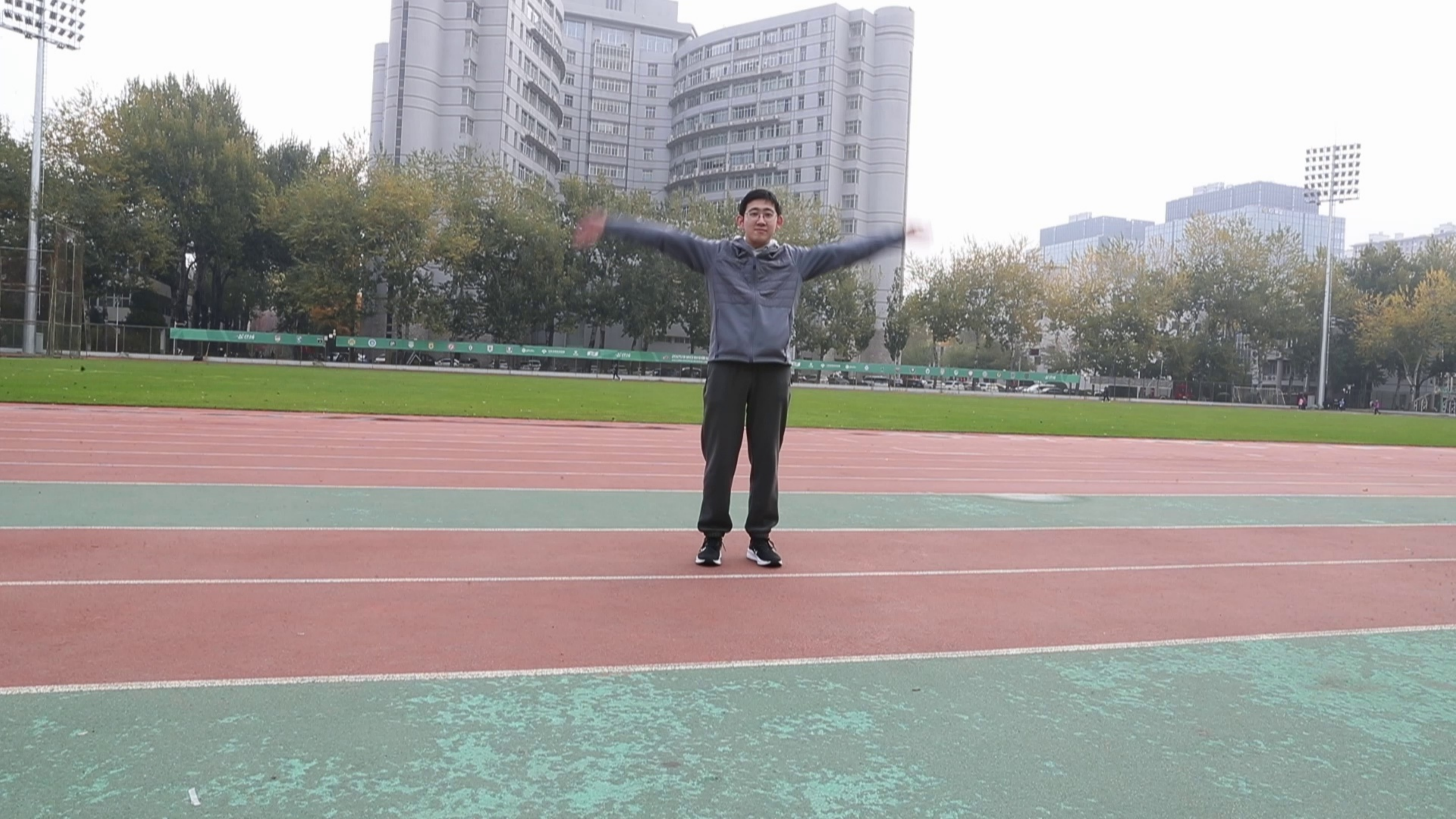}}&
            \hspace{-0.0cm}\raisebox{0.42cm}{\parbox[c]{0.8cm}{\centering\small\textbf{Left}}} &
			\hspace{-0.4cm}\fcolorbox{gray}{white}{\includegraphics[width=1.4cm,height=1.03cm]{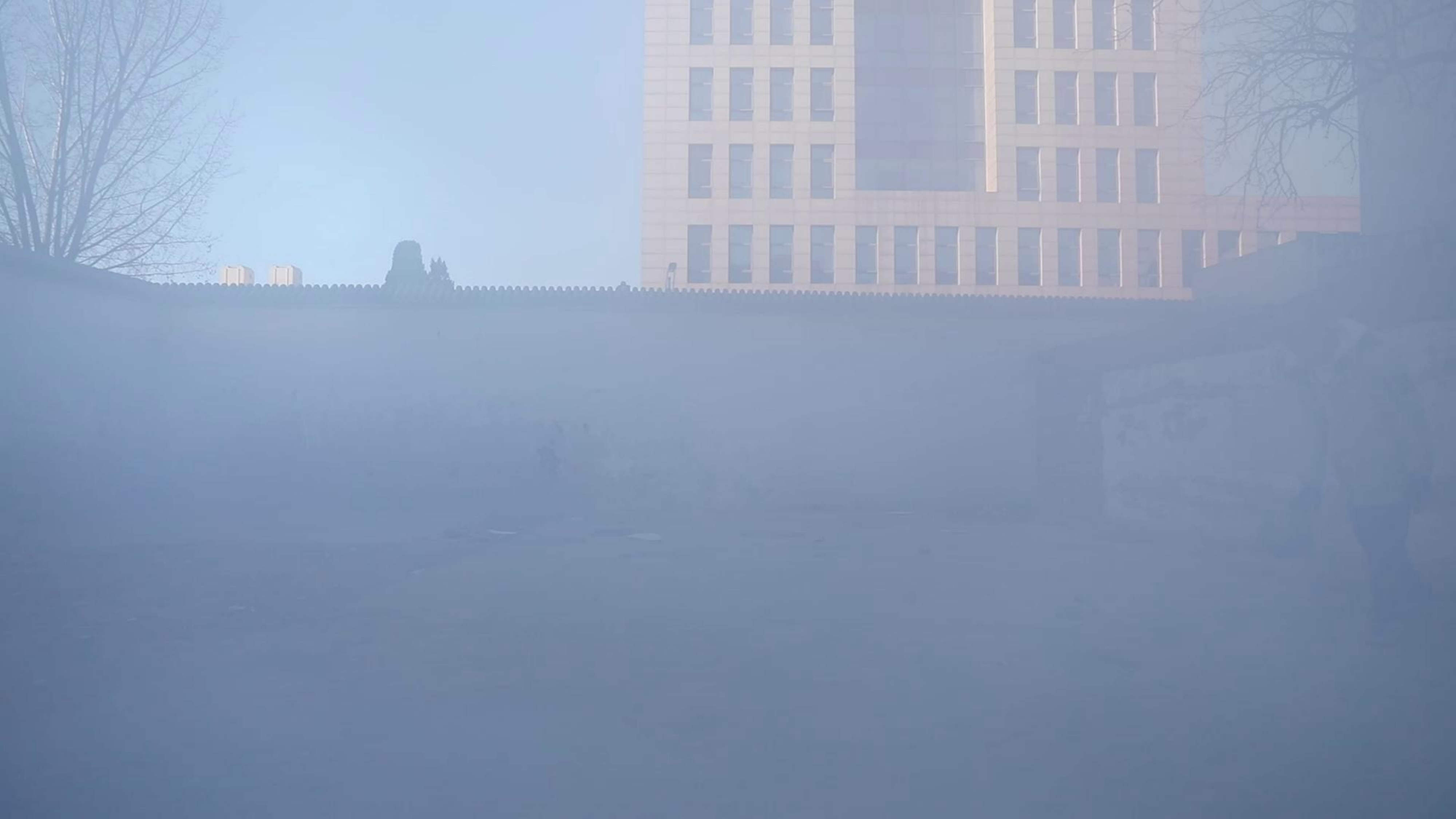}} & 
            \hspace{-0.35cm}\fcolorbox{gray}{white}{\includegraphics[width=1.4cm,height=1.03cm]{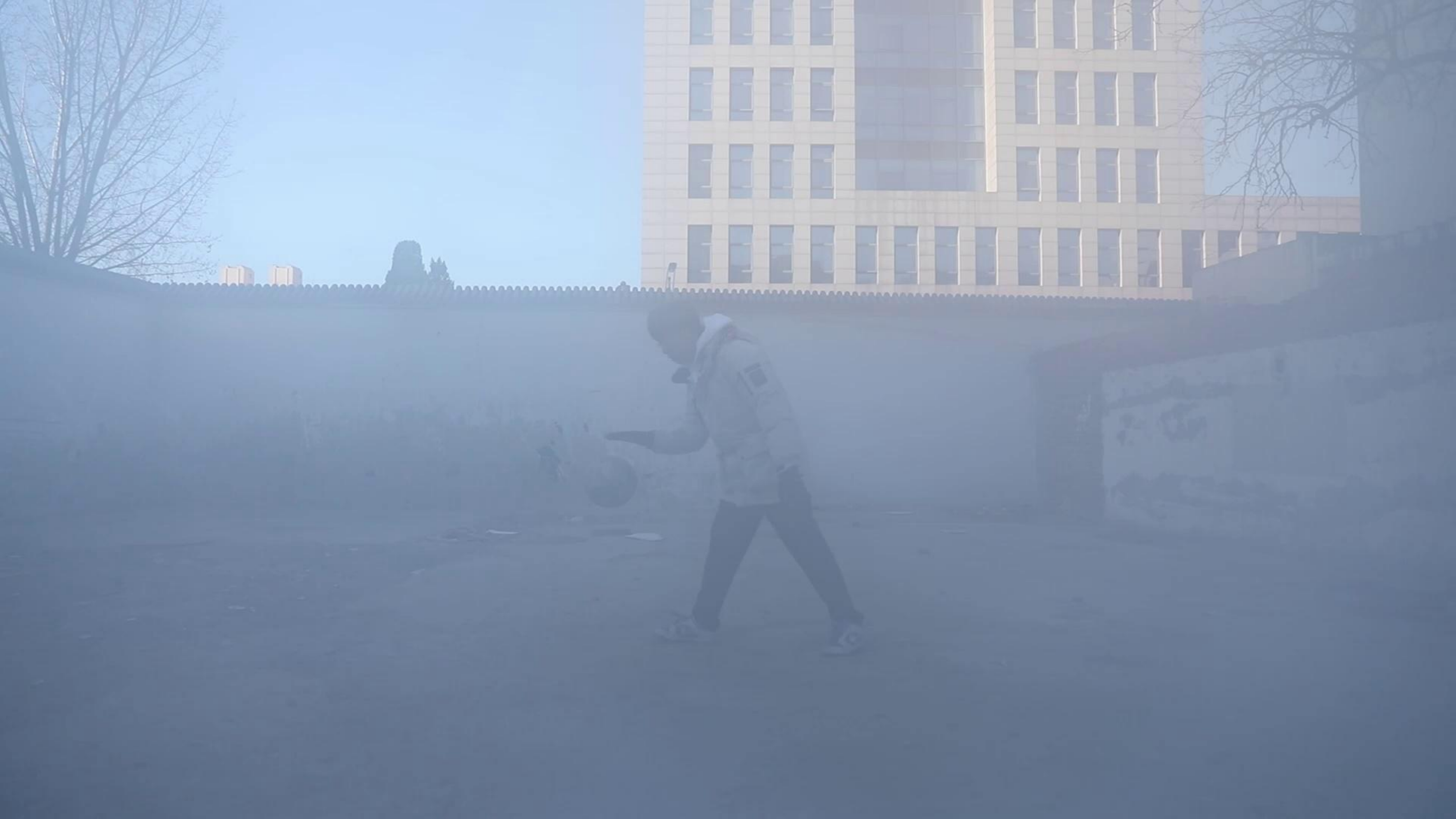}} & \hspace{-0.35cm}\fcolorbox{gray}{white}{\includegraphics[width=1.4cm,height=1.03cm]{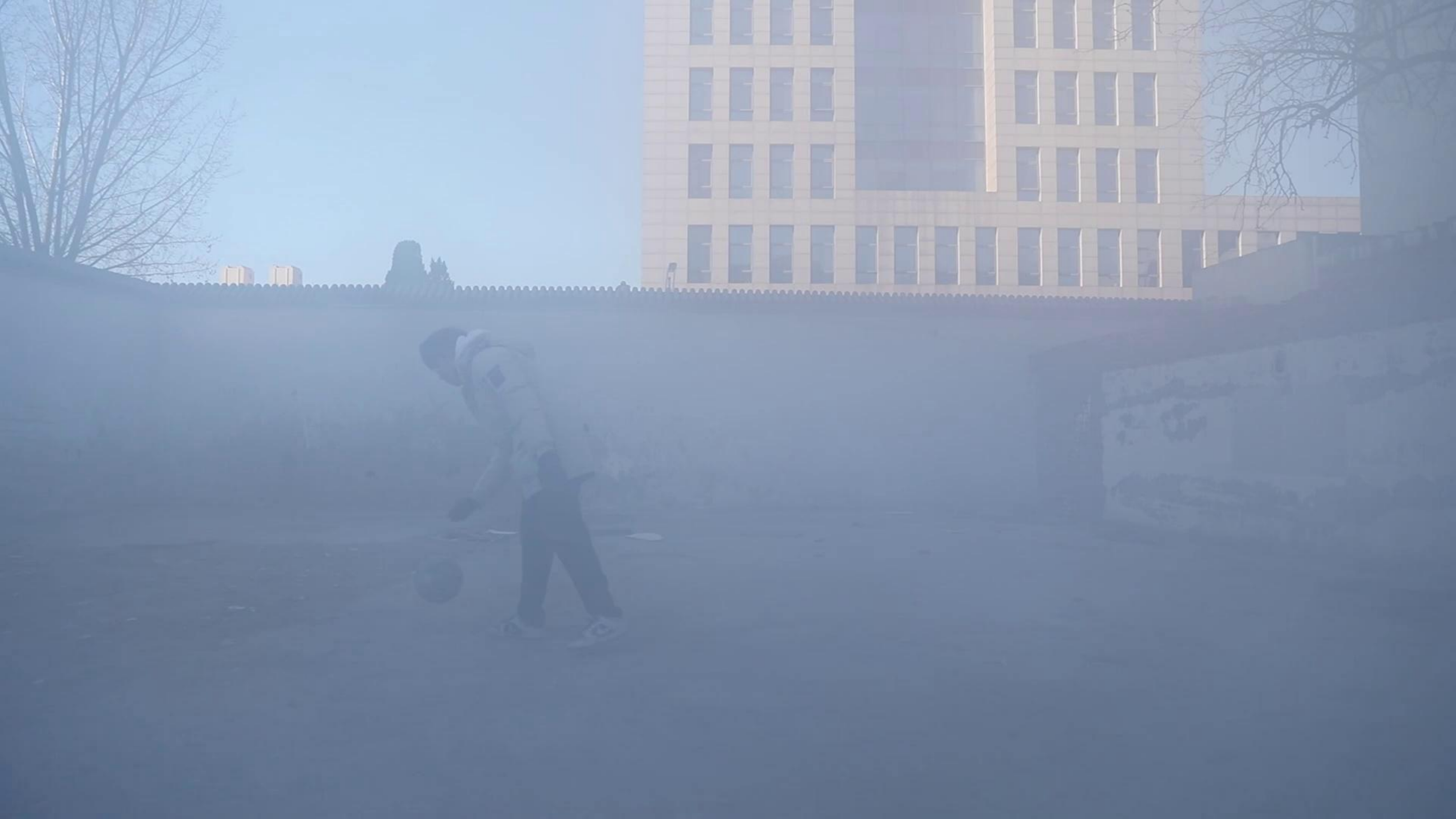}} & 
            \hspace{-0.35cm}\fcolorbox{gray}{white}{\includegraphics[width=1.4cm,height=1.03cm]{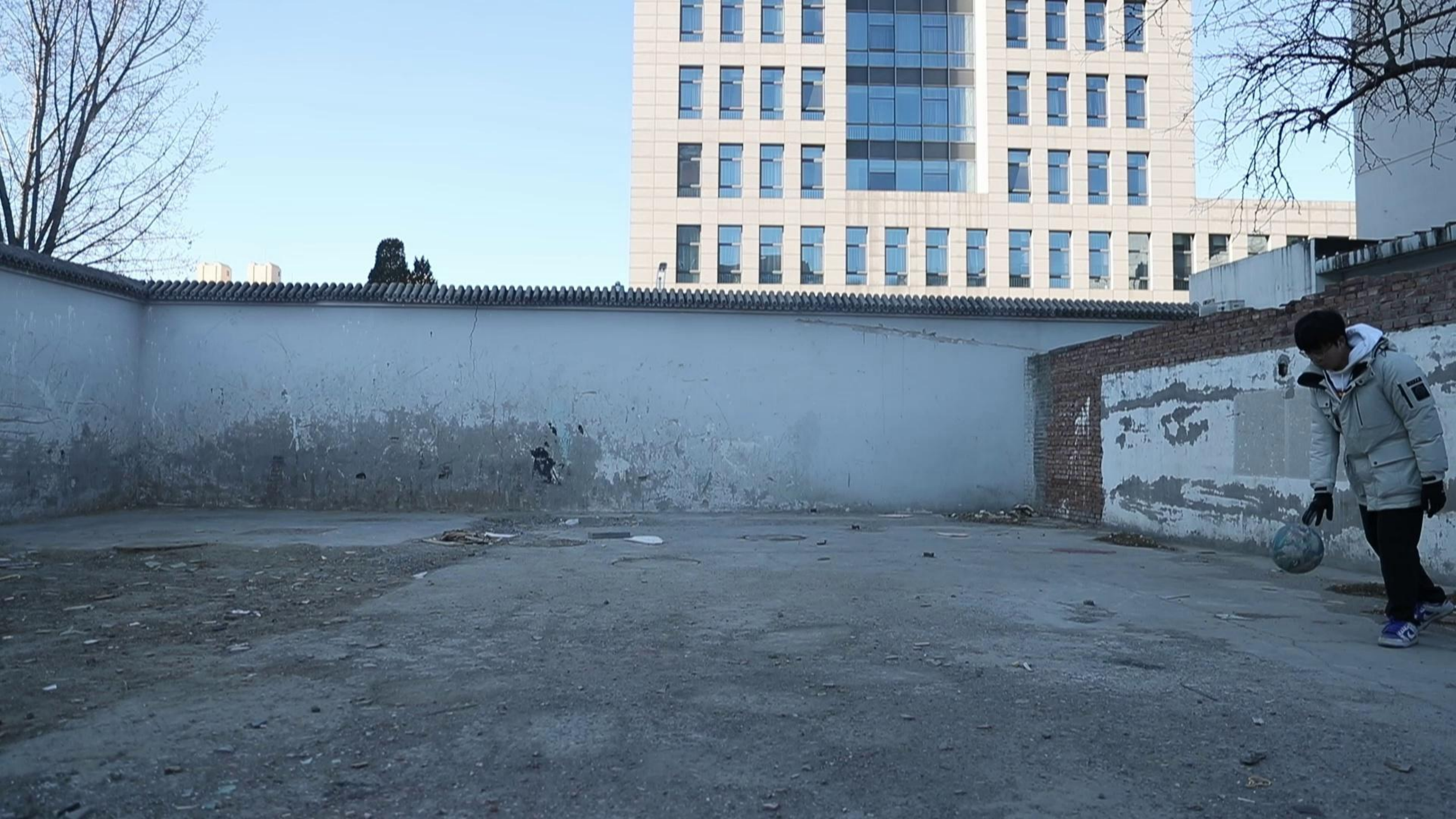}} 
            & \hspace{-0.35cm}\fcolorbox{gray}{white}{\includegraphics[width=1.4cm,height=1.03cm]{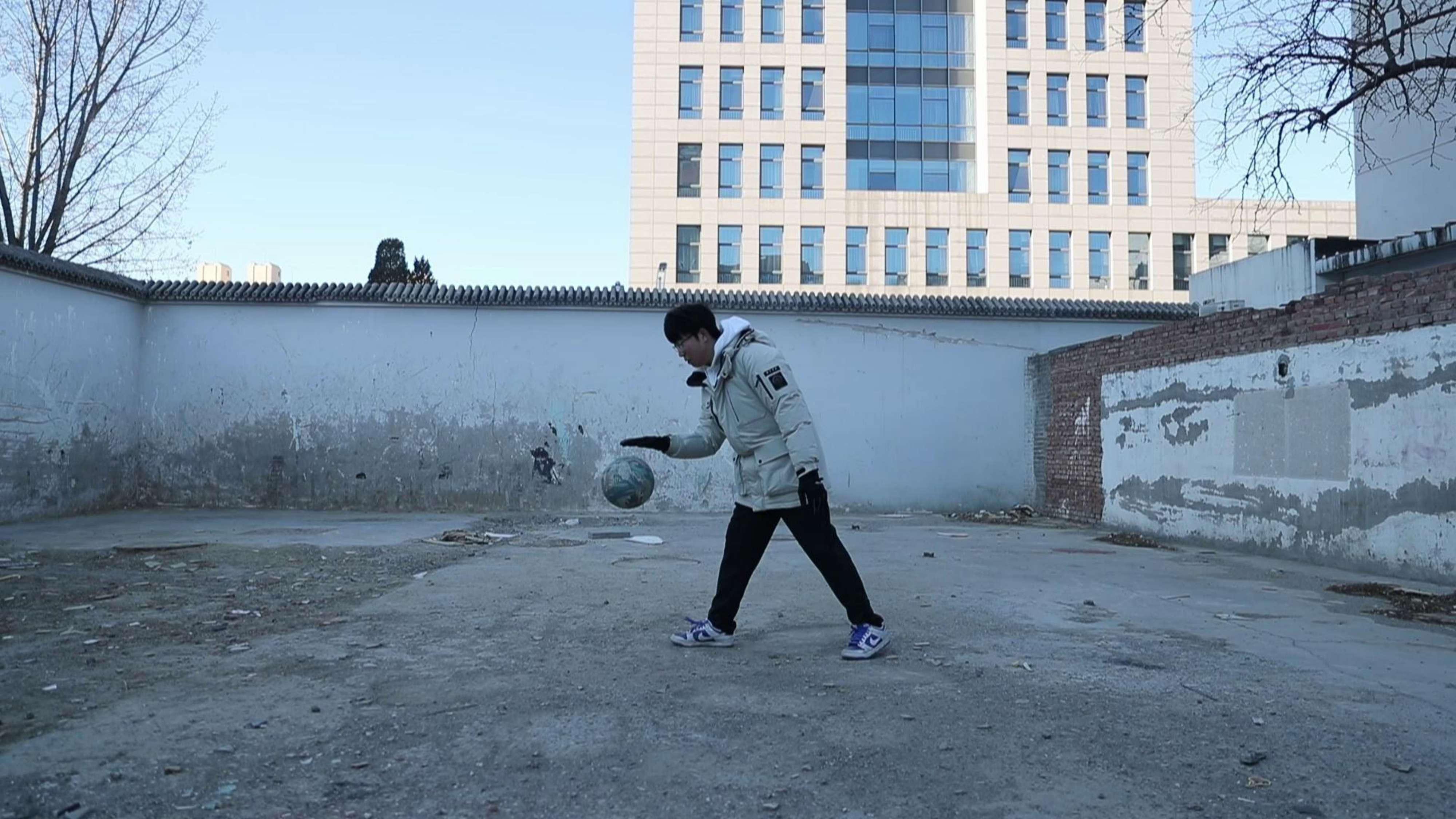}} & \hspace{-0.35cm}\fcolorbox{gray}{white}{\includegraphics[width=1.4cm,height=1.03cm]{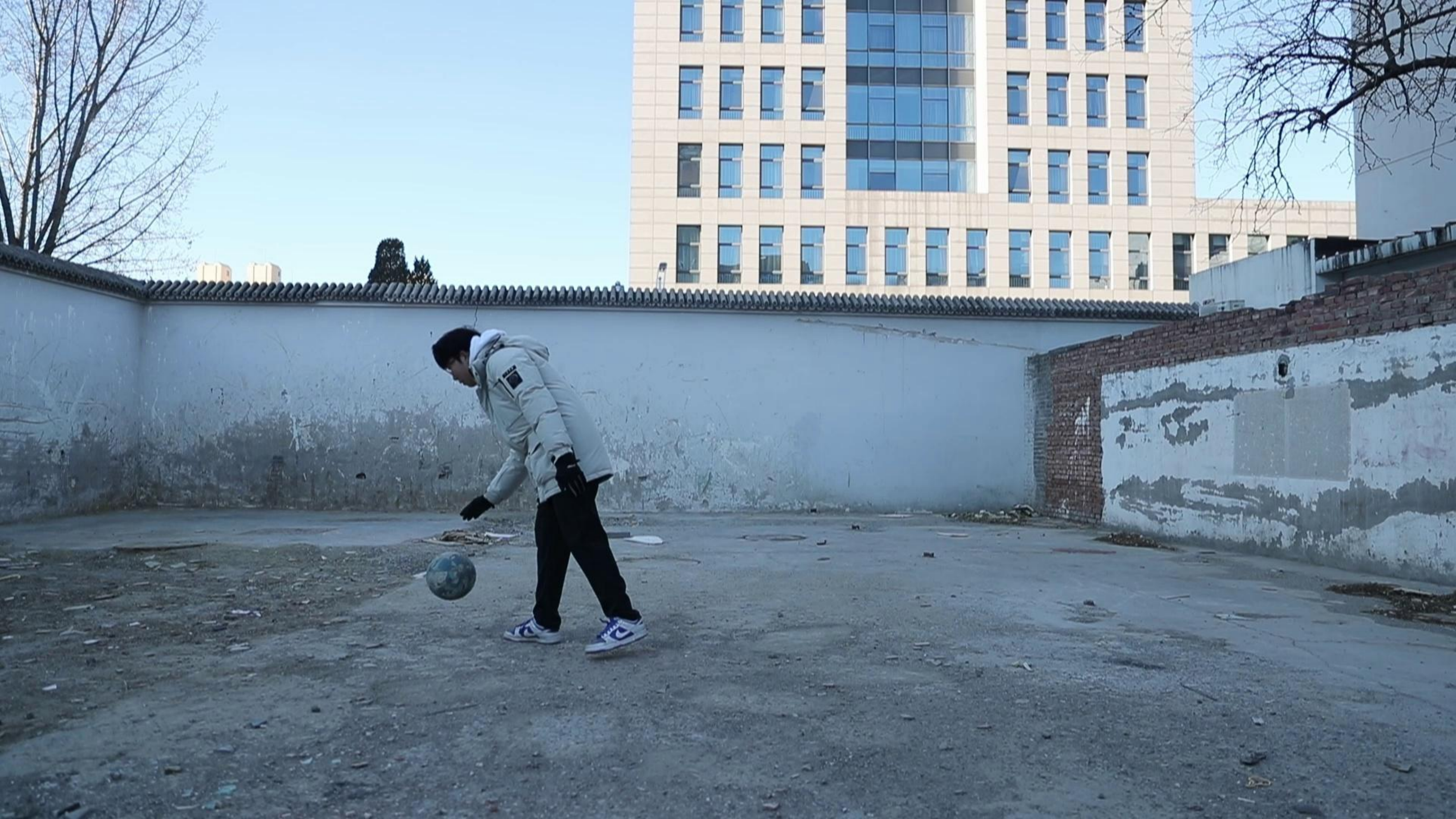}} & 
            \\
             \vspace{0.2cm}
             \hspace{-0.0cm}\raisebox{0.38cm}{\parbox[c]{1.2cm}{\centering\small\textbf{\shortstack{Mop\\floor}}}} &
            \fcolorbox{gray}{white}{\includegraphics[width=1.4cm,height=1.03cm]{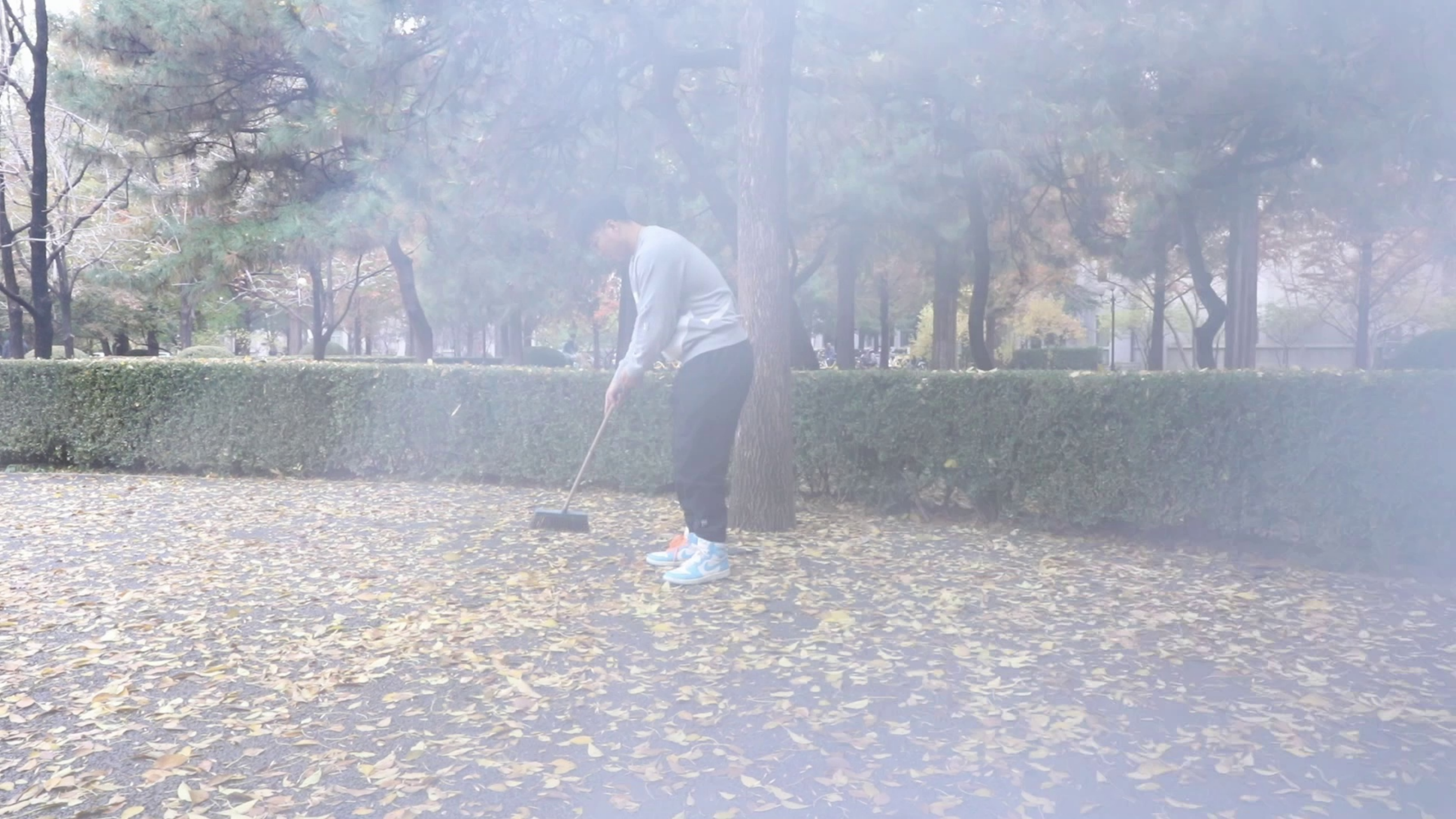}} &
            \hspace{-0.42cm}\fcolorbox{gray}{white}{\includegraphics[width=1.4cm,height=1.03cm]{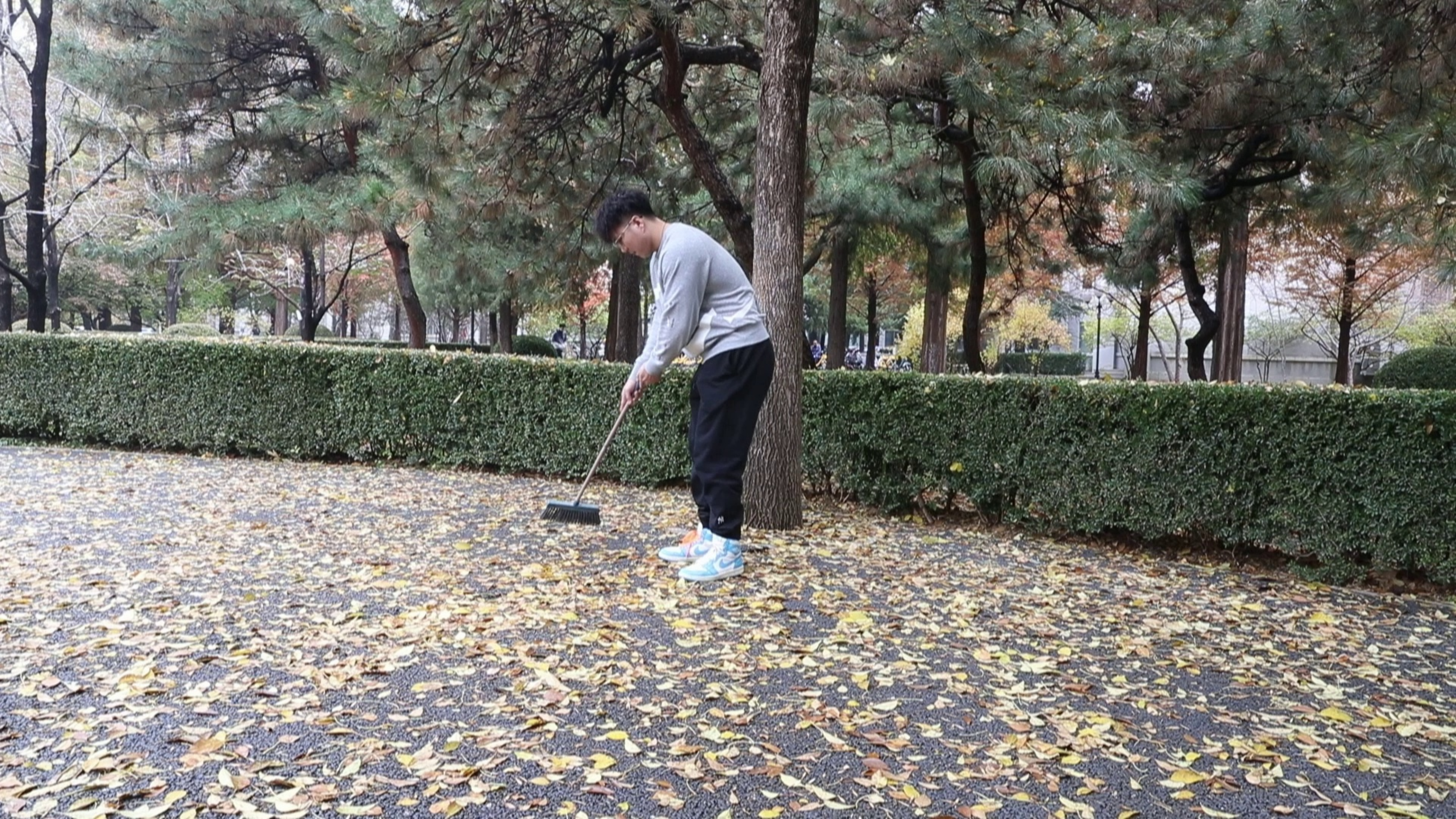}}&  

            \fcolorbox{gray}{white}{\includegraphics[width=1.4cm,height=1.03cm]{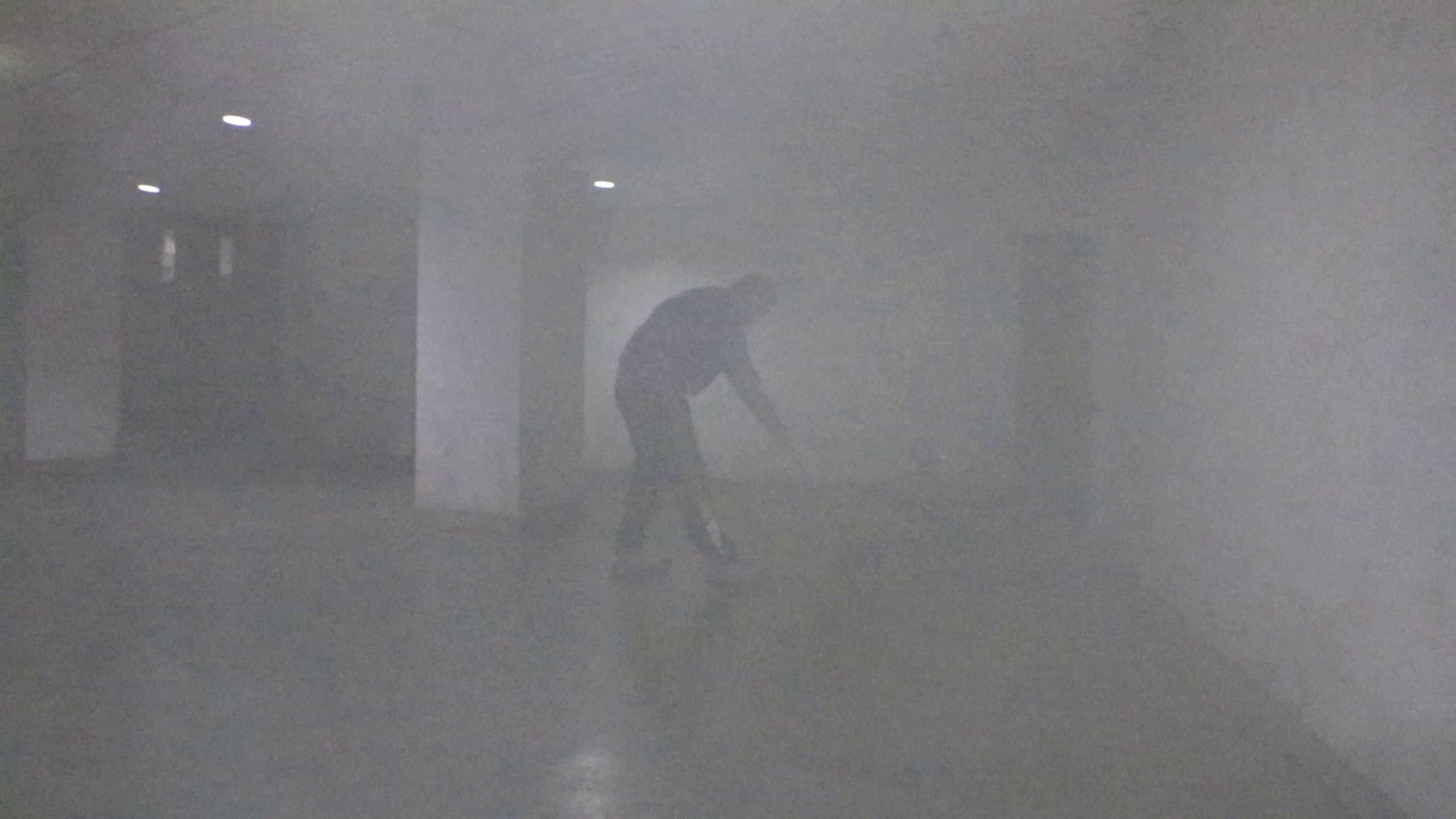}} &  
            \hspace{-0.5cm}\fcolorbox{gray}{white}{\includegraphics[width=1.4cm,height=1.03cm]{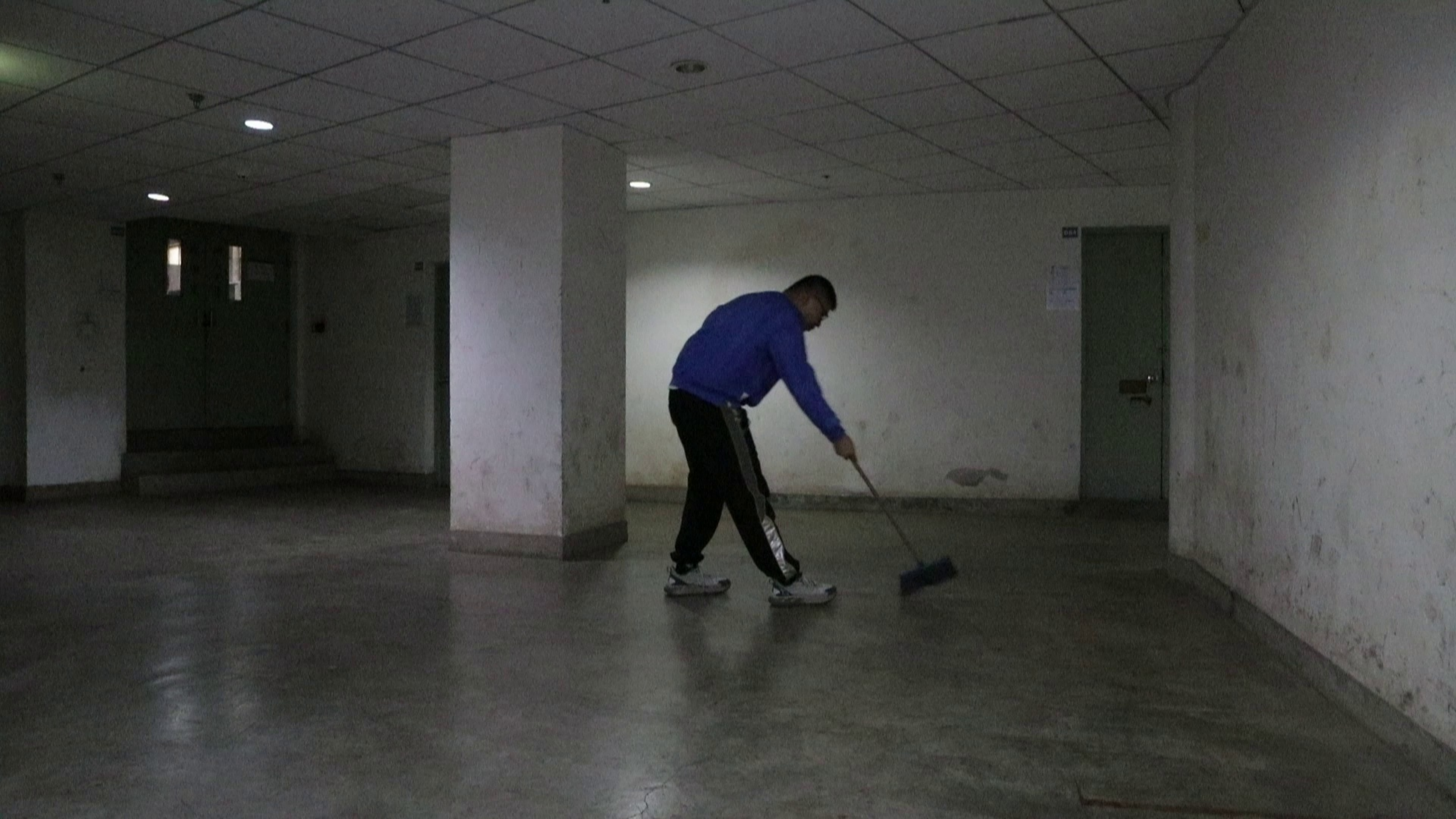}}&
            
            \hspace{-0.0cm}\raisebox{0.42cm}{\parbox[c]{0.8cm}{\centering\small\textbf{Right}}} &
			\hspace{-0.4cm}\fcolorbox{gray}{white}{\includegraphics[width=1.4cm,height=1.03cm]{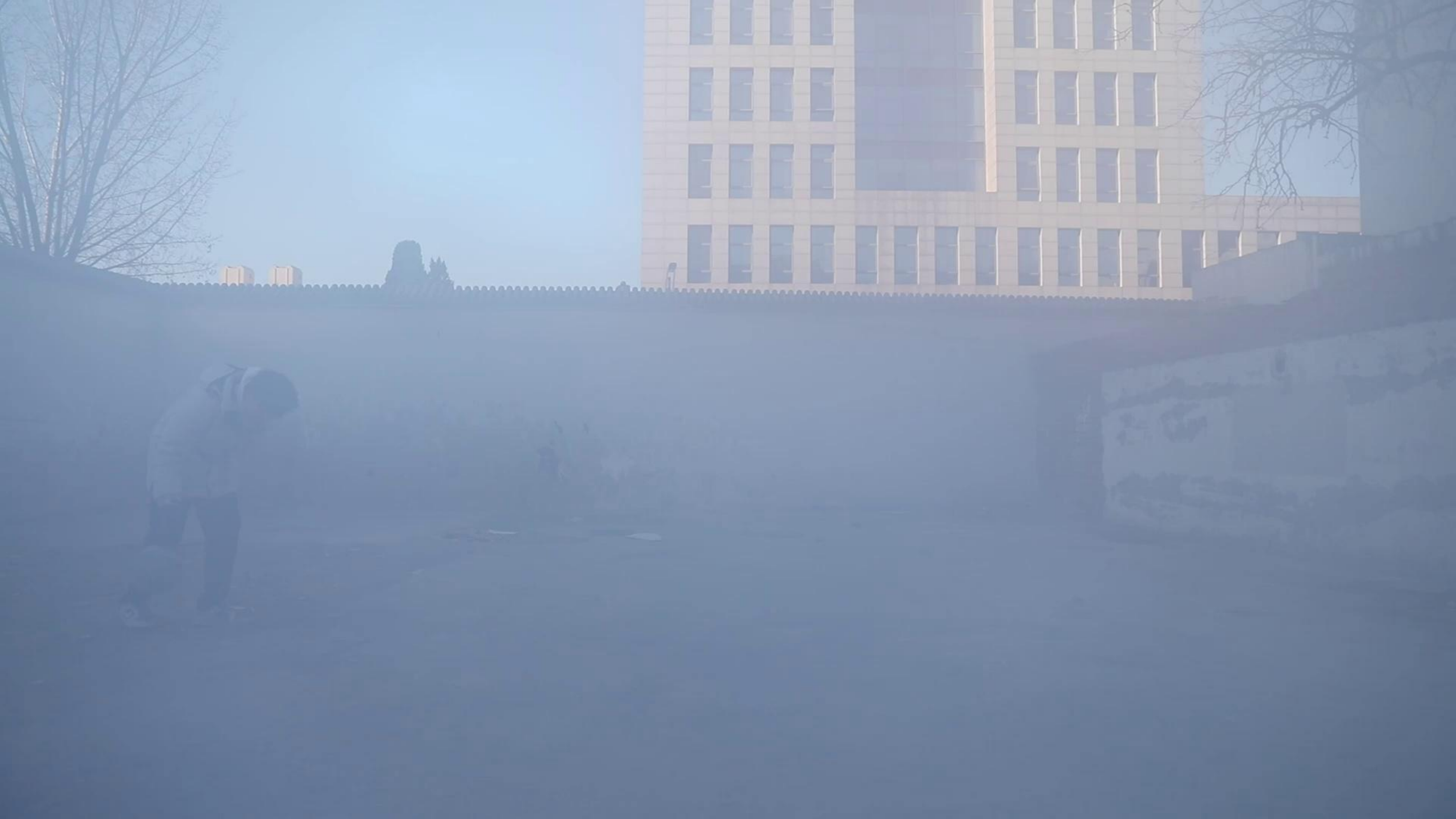}} & 
            \hspace{-0.35cm}\fcolorbox{gray}{white}{\includegraphics[width=1.4cm,height=1.03cm]{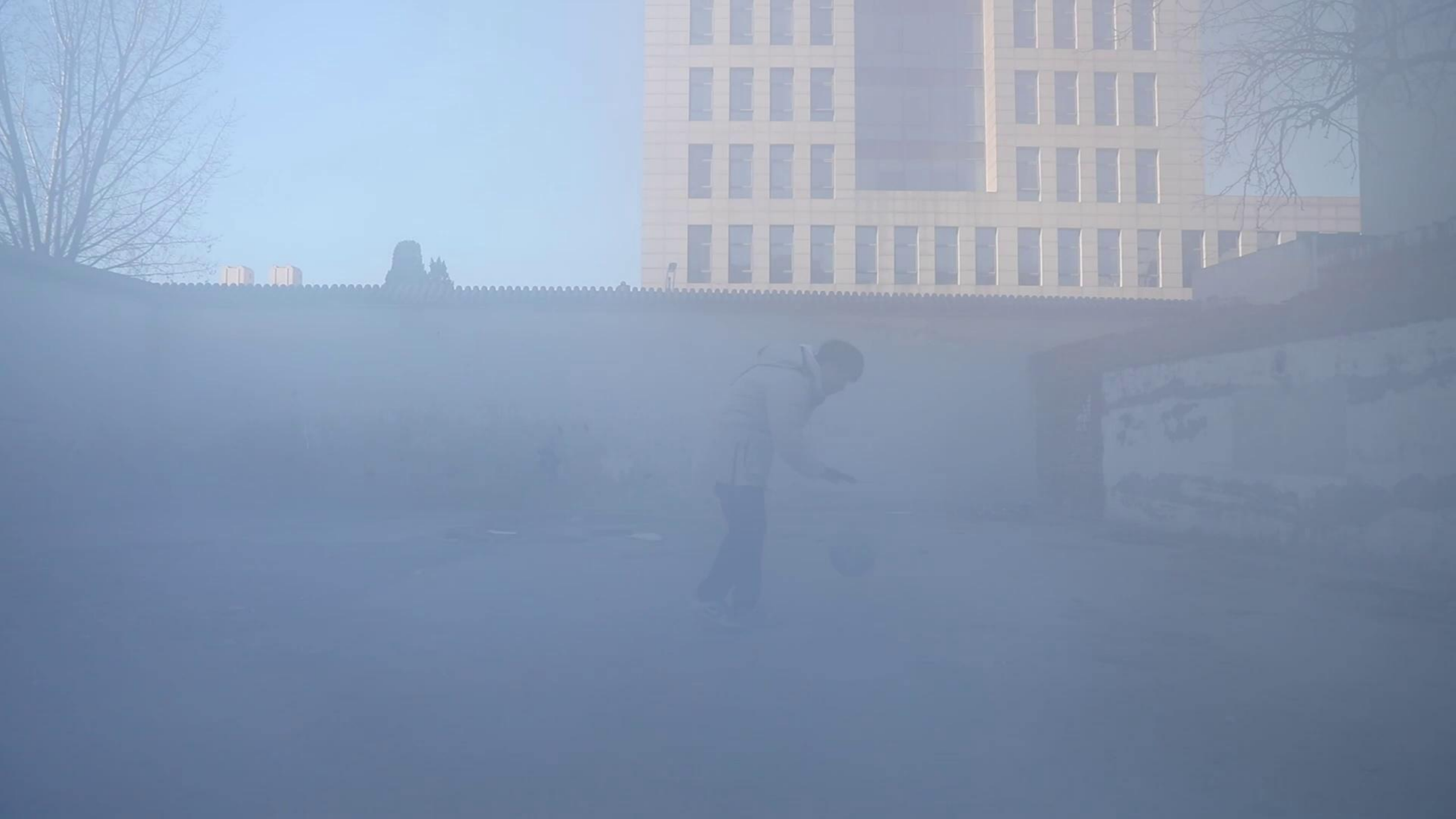}} & \hspace{-0.35cm}\fcolorbox{gray}{white}{\includegraphics[width=1.4cm,height=1.03cm]{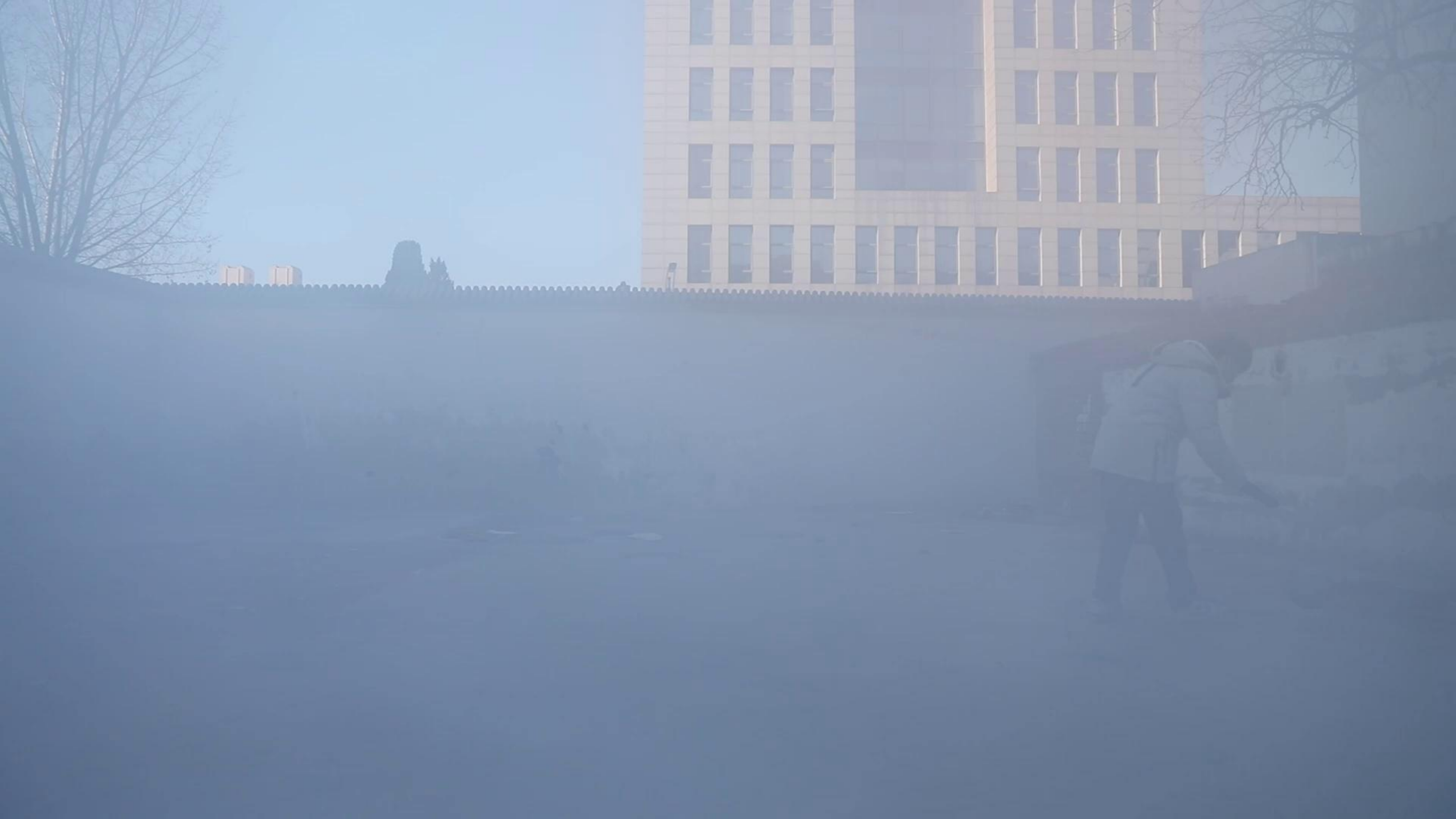}} & 
            \hspace{-0.35cm}\fcolorbox{gray}{white}{\includegraphics[width=1.4cm,height=1.03cm]{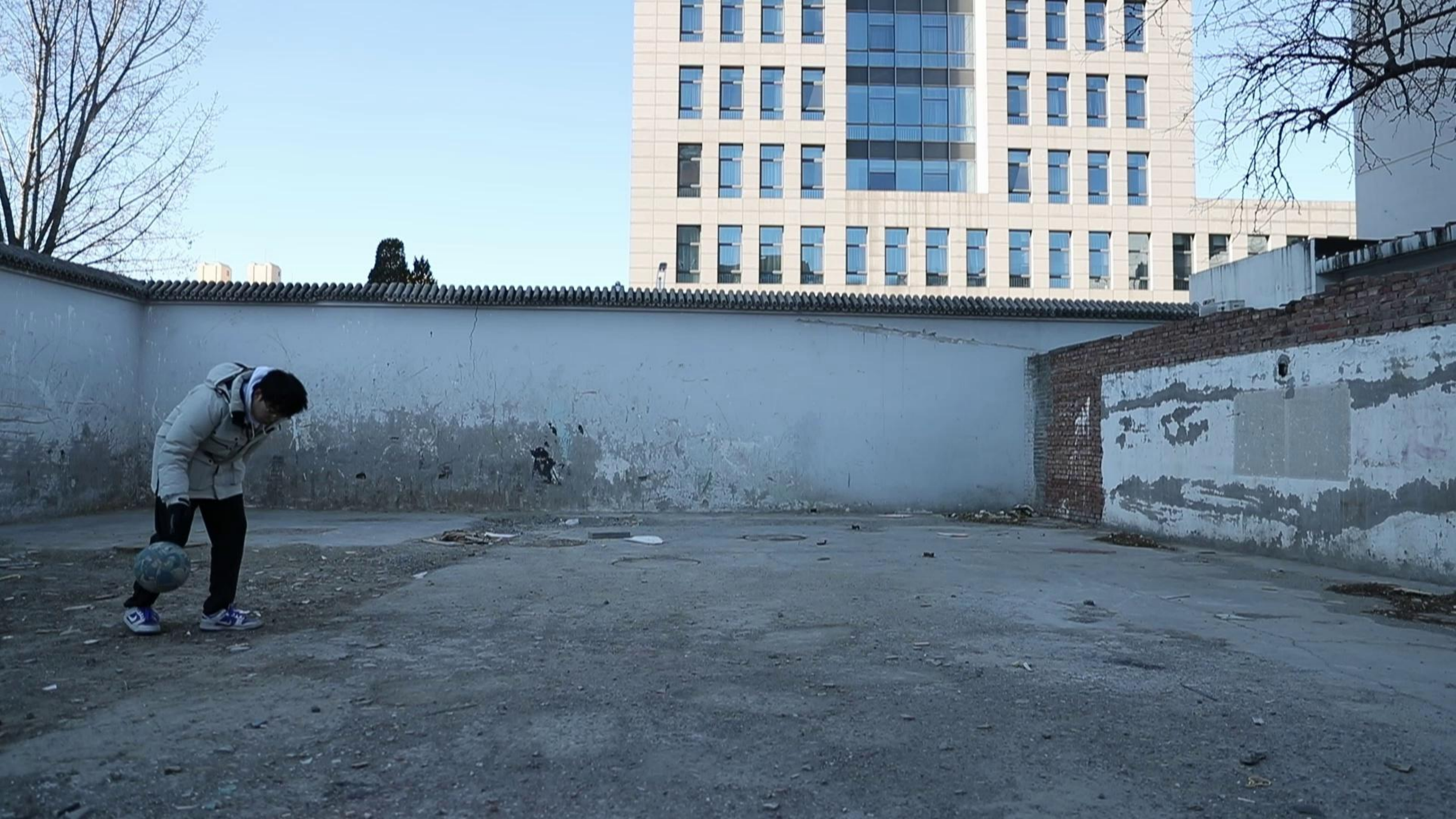}}
            & \hspace{-0.35cm}\fcolorbox{gray}{white}{\includegraphics[width=1.4cm,height=1.03cm]{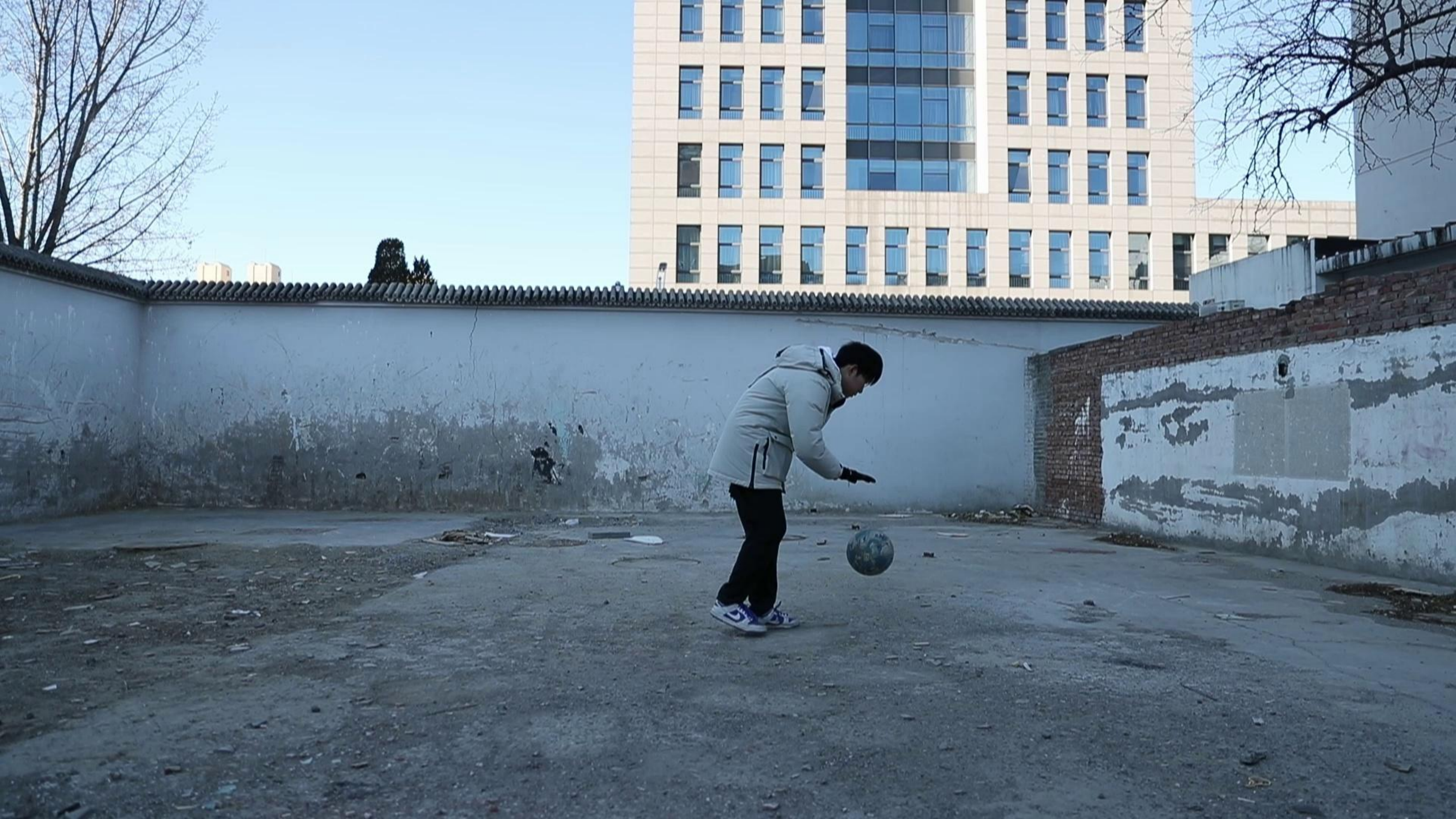}} & \hspace{-0.35cm}\fcolorbox{gray}{white}{\includegraphics[width=1.4cm,height=1.03cm]{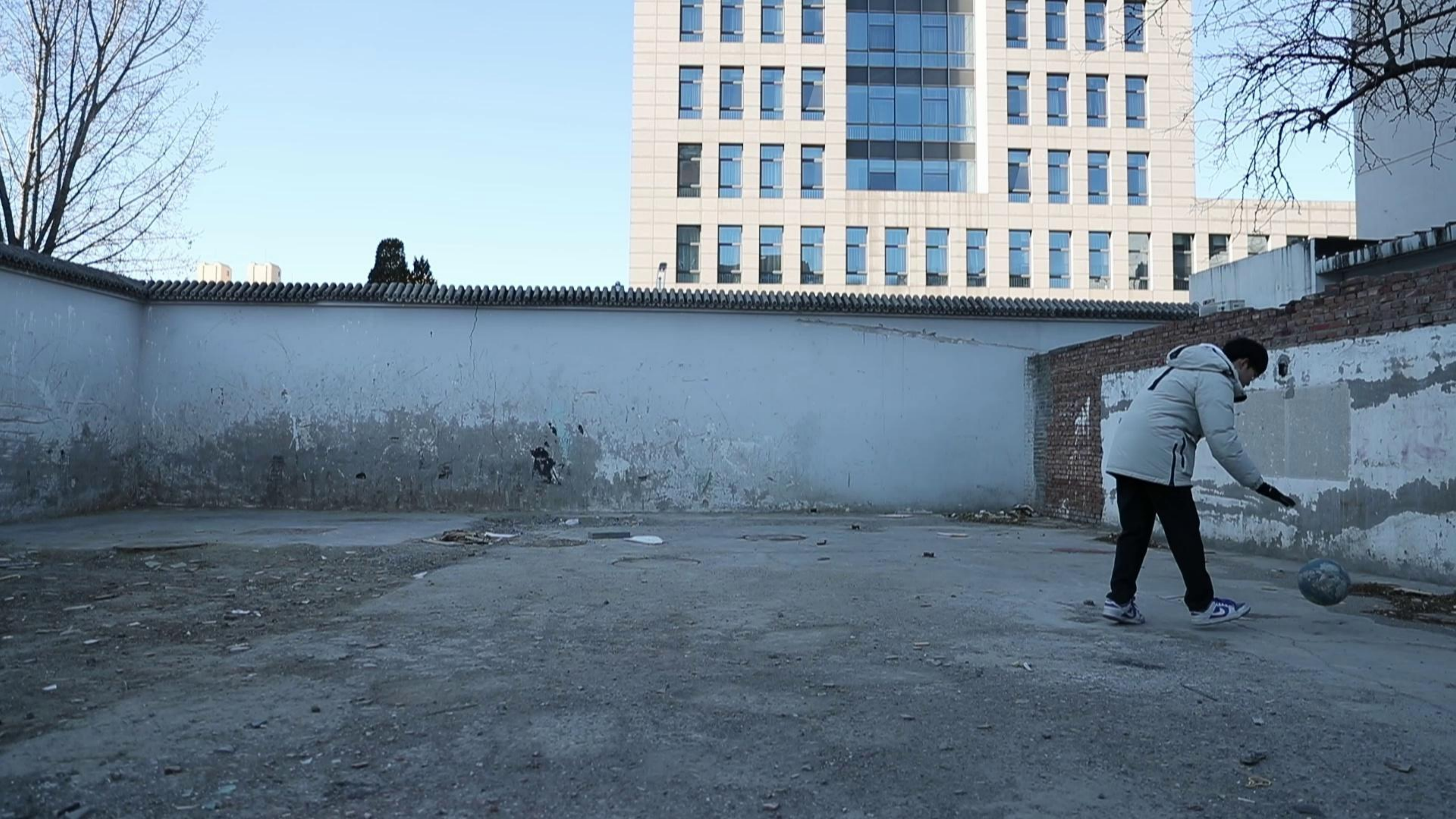}} 
			\end{tabular}
            \vspace{-0.25cm}
			\caption{
            Examples from our FogAct dataset, including four categories. Each category is captured under two fog conditions: light fog examples and dense fog examples. Additionally, each action is recorded from four different perspectives: front, back, left, and right. To illustrate this, we use \textit{`Dribble basketball'} as an example, showing four perspectives at a dense fog intensity level. Each sequence contains three frames sampled along the timeline. The three images on the left show foggy images, while the three on the right display the clean images. Additional examples are provided in the supplementary material.
            }
			\label{fig:example}
		\end{table*}	

Beyond the error accumulation caused by poorly restored images, we also observe that defogging modules fail to recover action-related semantics satisfactorily. As shown in Fig.~\ref{fig:fog} and Fig.~\ref{fig:dehaze}, despite the application of defogging techniques, the semantic feature distribution demonstrates minimal improvement compared to clean videos (Fig.~\ref{fig:clean}), both in terms of intra-class compactness and classification boundaries. These limitations motivate further exploration to address this problem effectively.

In this paper, to alleviate the aforementioned issue, we present i) FogAct, the first benchmark
dataset for foggy action recognition, and ii) FogNet, an end-to-end framework for foggy action recognition.

{\bf i) FogAct.} Unlike existing methods \cite{tanneru2021action, chaudhary2018tsnet} that employ the atmospheric scattering model (ASM)~\cite{narasimhan2003contrast} to simulate fog on datasets such as HMDB-51 \cite{kuehne2011hmdb} and UCF-101 \cite{soomro2012dataset}, our FogAct dataset is collected from videos captured in real outdoor foggy environments. We designed a stereo video acquisition system that captures clean and foggy videos simultaneously, ensuring frame-wise semantic alignment. 
Our FogAct dataset consists of 55 distinct action categories captured across 10 different scenes, including construction sites, academic buildings, and classroom playgrounds, totaling 10,000 video clips, as shown in Fig.~\ref{fig:example}. Compared to existing datasets, FogAct introduces more unpredictable visibility degradation, motion blur, and reduced contrast, which are challenging to reproduce accurately using idealized atmospheric scattering models (ASM).

\begin{tikzpicture}[overlay]
  \draw[dashed, thick, rounded corners=3pt, draw=red]
    (1.75+5.8,9+2.3+0.45+0.35) -- (11.75+5.8,9+2.3+0.45+0.35) -- (11.75+5.8,14.7+2.3+0.45+0.35) -- (1.75+5.8,14.7+2.3+0.45+0.35) -- cycle;

  \draw[dashed, thick, draw=red]
    (1.35+5.8, 13.8+2.3+0.45+0.35) -- (1.73+5.8, 14.0+2.3+0.45+0.35) -- (1.35+5.8, 14.2+2.3+0.45+0.35);
  \fill[black, opacity=1] (16.59, 17.18) rectangle (16.64, 17.20);
\fill[black, opacity=1] (15.08, 17.13) rectangle (15.13, 17.15);
\fill[black, opacity=1] (13.56, 17.15) rectangle (13.60, 17.17);
\fill[black, opacity=1] (6.34, 17.15) rectangle (6.39, 17.16);
\fill[black, opacity=1] (6.36, 15.79) rectangle (6.43, 15.81);

\fill[black, opacity=1] (3.14, 15.90) rectangle (3.19, 15.93);

\fill[black, opacity=1] (6.385, 14.59) rectangle (6.45, 14.62);

\fill[black, opacity=1] (16.34, 14.42) rectangle (16.39, 14.44);
\fill[black, opacity=1] (15.05, 14.45) rectangle (15.10, 14.47);
\fill[black, opacity=1] (14.16, 14.44) rectangle (14.20, 14.46);

\fill[black, opacity=1] (2.96, 13.205) rectangle (3.01, 13.245);

\fill[black, opacity=1] (3.14, 17.17) rectangle (3.19, 17.20);

\fill[black, opacity=1] (17.08, 13.08) rectangle (17.13, 13.10);
\fill[black, opacity=1] (15.22, 13.06) rectangle (15.27, 13.08);
\fill[black, opacity=1] (13.16, 13.01) rectangle (13.20, 13.03);

\node at (10.75, 12.34) {\scriptsize Time};
\node at (15.25, 12.34) {\scriptsize Time};

\draw[->] (8.95, 12.24) -- (12.55, 12.24);
\draw[->] (13.45, 12.24) -- (17.05, 12.24);
    
\end{tikzpicture}

{\bf ii) FogNet.} Unlike existing two-stage frameworks, our end-to-end FogNet bypasses traditional defogging by leveraging a two-stream CLIP model to find semantic similarities between foggy and clean videos. To acquire fog-invariant knowledge, FogNet integrates a Fog-Aware Selection (FAS) mechanism, generating semantically meaningful representations from both foggy and clean videos. Note that clean videos are used only during the training phase. A Mutual Enhancement (ME) module ensures complementary improvement, while a Cross-Stream Alignment (CSA) module aligns frames from clean and foggy video pairs, ensuring spatial correspondence and preserving temporal consistency, ultimately boosting recognition performance.

Our main contributions are as follows:
\begin{itemize}
\setlength\itemsep{+0.0em}
    \item We introduce FogAct, the first fog dataset with benchmarking for human action recognition, with $\approx10\mathrm{K}$ dynamic clean-foggy paired videos across 55 classes.

    \item We propose FogNet, the first end-to-end framework to leverage pre-trained vision-language models for foggy action recognition.
    
    \item Extensive experiments on real-world and synthetic datasets demonstrate that our method outperforms existing state-of-the-art approaches. Our code and dataset will be made available to facilitate reproducible research.
\end{itemize}


\begin{table*}[t]
    \centering
    \small
    \caption{Comparison of existing defogging datasets based on several key attributes. `I\&O' indicates whether the datasets include indoor or outdoor scenes. `S\&V' specifies whether the datasets contain single images or videos. `Dyn.' denotes whether the dataset includes actions that evolve over time. `Real' refers to whether foggy images are captured in real-world foggy conditions. `Mul.' indicates multiple foggy intensity levels. `Pair' highlights the inclusion of clean counterparts. `Pers.' represents the number of perspectives. Finally, `Anno.' specifies whether the datasets are annotated with action recognition labels.}
    \hspace{-0.4cm}
    \setlength{\tabcolsep}{7pt}

        \begin{tabular}{lcccccccccccc}
        \toprule
          \textbf{Dataset} & \textbf{Venue} & \textbf{Scale}   & \textbf{Resolution} &\textbf{I\&O} &\textbf{S\&V}  &\textbf{Dyn.} & \textbf{Real} & \textbf{Mul.} & \textbf{Pers.}&  \textbf{Pair} & \textbf{Anno.} \\
          \hline
           Foggy Zurich~\cite{sakaridis2018model} & ECCV'18 & 3808  & 1920 \x 1080 &O &S  & \xmark & \cmark & \cmark & 1 & \xmark & \xmark  \\
           SOTS~\cite{li2018benchmarking}& TIP'18  &  500 & 640 \x 480 &I & S &\xmark & \xmark & \cmark & 1 & \cmark & \xmark \\
           NH-HAZE~\cite{ancuti2020nh} & CVPR'20 &  55 & 1600 \x 1200 &O & S  &\xmark & \cmark & \xmark & 1 & \cmark & \xmark \\
           BeDDE~\cite{zhao2020dehazing} & TIP'20 & 208 & 1800 \x 1200 &O & S  &\xmark & \cmark & \xmark & 1 & \cmark & \xmark \\
           ACDC~\cite{sakaridis2021acdc} & ICCV'21 & 4006  & 1920 \x 1080 &O &S   & \xmark & \cmark & \xmark & 1 & \xmark & \xmark \\
            RHVD~\cite{chu2021real} & QoMEX'21 & 403  & 1600 \x 900 &O &V   & \xmark & \cmark & \xmark & 1 & \xmark & \xmark \\
           REVIDE~\cite{zhang2021learning}& CVPR'21 & 1982 &2708 \x 1800  &I & V & \xmark & \cmark & \cmark & 1 & \cmark & \xmark \\
           VIREDA~\cite{duminil2023new} & ASPAI'22 &  6 & 1670 \x 1080 &I  & S &\xmark & \cmark & \cmark & 1 & \cmark & \xmark \\
           HazeWorld~\cite{xu2023video}& CVPR'23  &  5084 & 1229 \x 700 &O & S  &\xmark & \xmark & \cmark & 1  & \cmark & \xmark \\
          \hline
          Ours &2025 &  9724 & 1920 \x 1080 &I\&O & V  & \cmark & \cmark & \cmark & 4 & \cmark & \cmark \\
       \bottomrule
      \end{tabular}
    \label{tab:1}
\end{table*}

\section{Related Work}

\noindent{\bf Action Recognition.}~Unlike detection and segmentation~\cite{bi2024learning, gupta2024robust}, foggy action recognition remains underexplored. Existing methods largely rely on handcrafted embeddings~\cite{tanneru2021action, chaudhary2019depth}, which are domain-specific and poorly adapt to complex, dynamic environments. 
They use synthetic datasets with idealized fog, missing real degradations like blur and contrast. Existing works fall into two categories. The first uses visual features with linear classifiers for direct action prediction~\cite{yang2023aim, wu2023can, rasheed2023fine, kim2024leveraging, liu2024lightweight}. The second leverages multimodal cues~\cite{wang2021actionclip, wu2023bidirectional, wang2024multimodal, chen2024ost, liu2024storyboard}, using text to enrich visual representations. However, both rely on clean-scene distributions and struggle under fog. In contrast, our method extracts fog-invariant features, bridging domain gaps and enhancing robustness in foggy conditions.

\noindent{\bf Defogging and Foggy Datasets.}~Fog degrades visual features via atmospheric scattering. Defogging methods include physical and learning-based approaches. Physical models (e.g., ASM\cite{islam2024hazespace2m, ullah2021light}) estimate transmission maps but rely on assumptions that often break in real fog. Learning-based methods (e.g., LDR~\cite{yang2024language}, DEA-Net~\cite{chen2024dea}, EDI~\cite{pan2019bringing}) restore visibility via data-driven mappings but may introduce artifacts or distort motion, impairing action recognition. Most foggy datasets are synthetic (e.g., Foggy Cityscapes~\cite{sakaridis2018semantic}, NH-HAZE~\cite{ancuti2020nh}), low-cost but unrealistic. Real datasets like Foggy Zurich~\cite{sakaridis2018model} and EDHaze~\cite{zhang2025dehazemamba} are more realistic but static. In contrast, our FogAct captures diverse real-fog actions, motivating future exploration of defogging techniques in foggy environments.

\begin{figure}[t]
    \centering
    \begin{overpic}[width=0.45\textwidth]
    {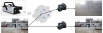}
        \put(14.5, 8.7){\color{black}\rule{1pt}{1pt}}
        \put(86.2, 8.7){\color{black}\rule{1pt}{1pt}}
        \put(3,14.3){\scriptsize \textbf{Fogging machine}}
        \put(79,-1.9){\scriptsize \textbf{Clean video}}
        \put(79,14.15){\scriptsize \textbf{Foggy video}}
        \put(52,29){\scriptsize \textbf{Right camera}}
        \put(52,12){\scriptsize \textbf{Left camera}}
        \put(10,-1.9){\scriptsize \textbf{Action}}
    \end{overpic}
    \caption{Overview of the Stereo Video Acquisition System. It includes two Canon DSLR cameras configured for stereo imaging, and a professional fogging machine. 
    The lights from a scene point are reflected and refracted by blue fog particles, resulting in a foggy video in the right camera. In contrast, light directly forms a clean video in the left camera.
    }
    \label{fig:stereo}
\end{figure}

\section{FogAct Dataset}
 
\subsection{Data Collection and Annotation}
To capture real-world foggy actions alongside their fog-free counterparts, we designed a stereo video acquisition system (Fig.~\ref{fig:stereo}). Both cameras share identical settings (focal length and lens) and are horizontally aligned with minimal separation. A professional fogging machine (DJPOWERE-1500W) generates high-quality fog before the right camera, producing realistic foggy action videos. The cameras are synchronized: the left records clean actions, and the right records foggy ones.

\begin{figure*}[t]
    \centering
    \begin{subfigure}{0.26\textwidth}
        \centering
        \vspace{-10mm}
        \rotatebox{0}{\includegraphics[width=\textwidth]{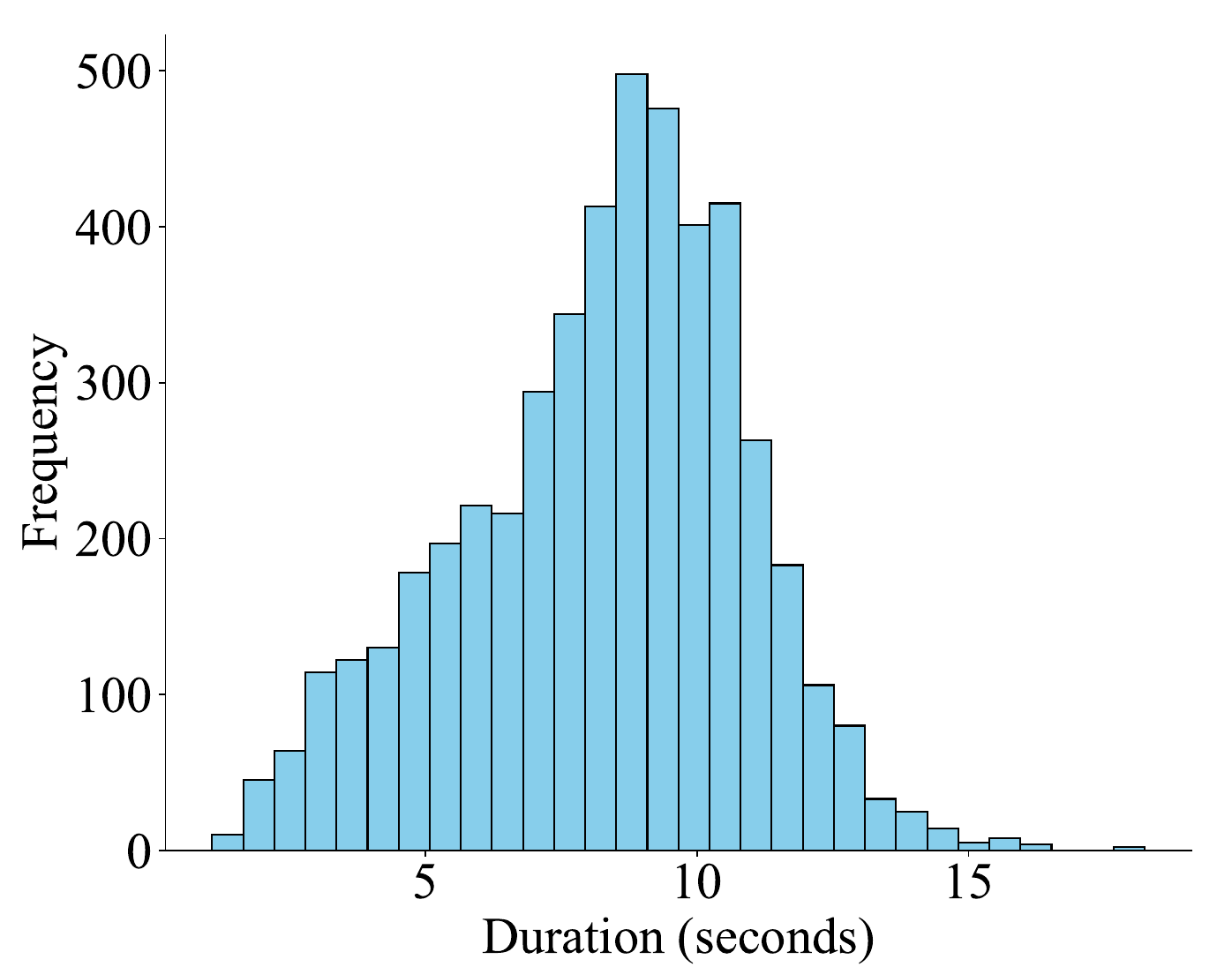}}
        \vspace{-6mm}
        \caption{Distribution of the action length.}
        \label{fig:dur}
    \end{subfigure}
    \hfill
     \begin{subfigure}{0.50\textwidth}
        \centering
        \includegraphics[width=\textwidth]{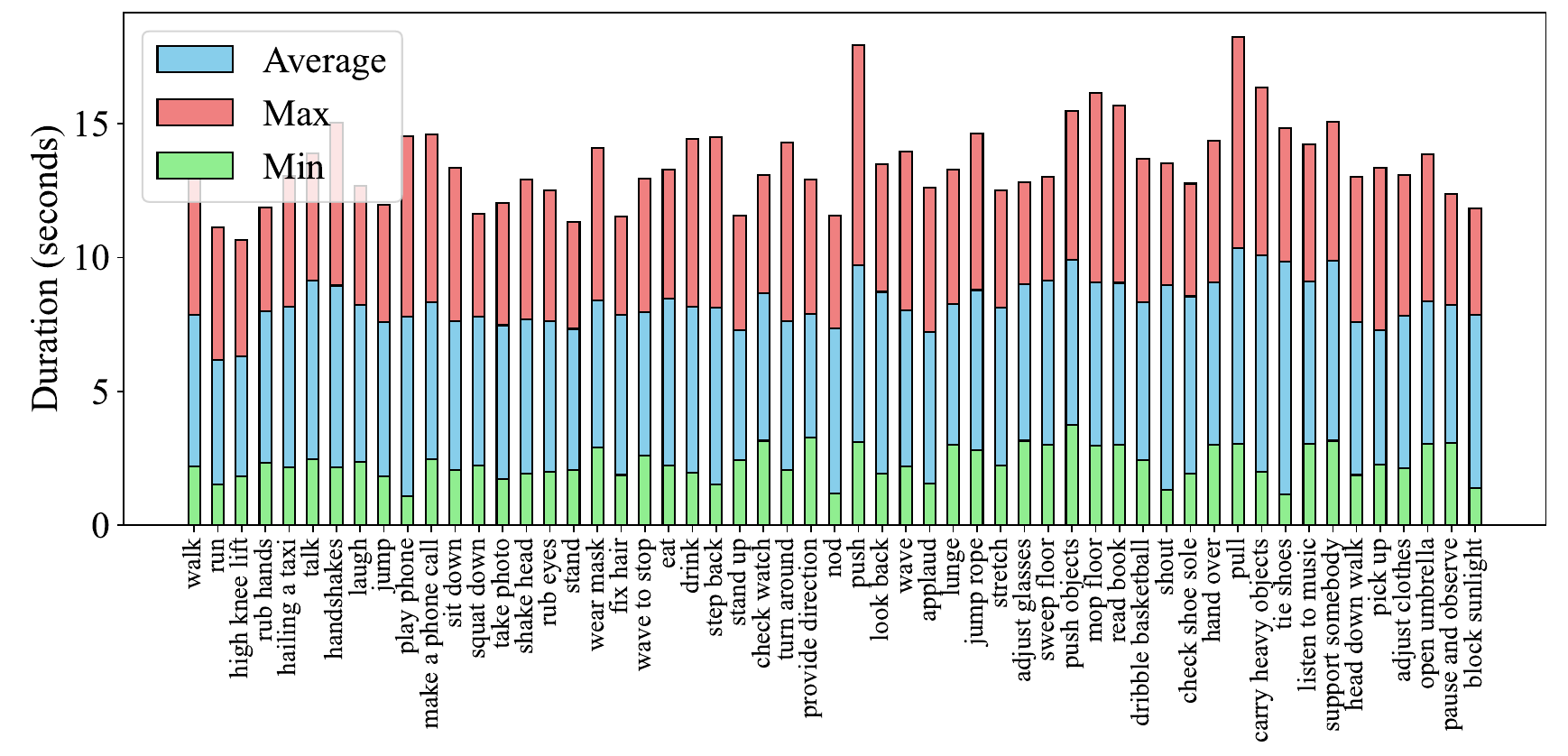}
        \vspace{-6mm}
        \caption{Duration of action per class.}
        \label{fig:dur1}
    \end{subfigure}
    \hfill
    \begin{subfigure}{0.23\textwidth}
        \centering
        \includegraphics[width=\textwidth]{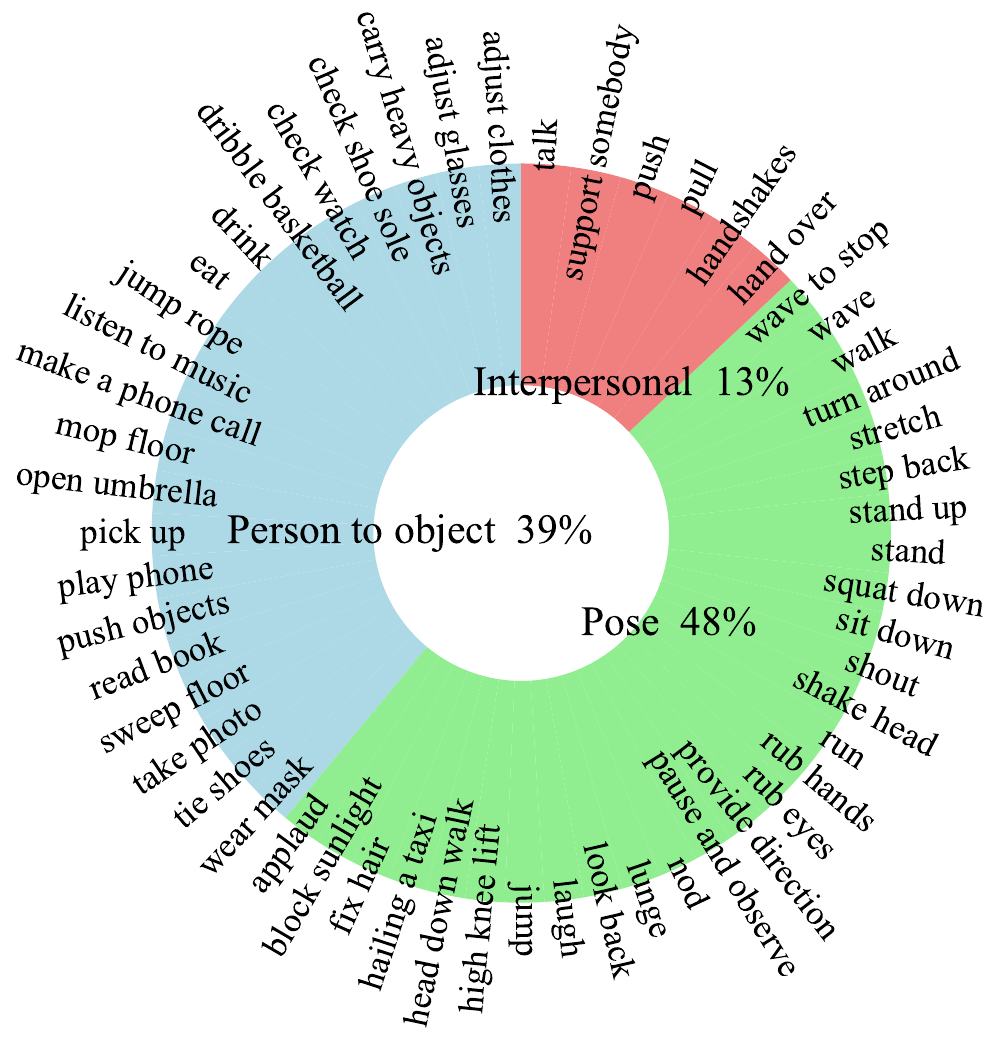}
        \vspace{-6mm}
        \caption{Actions of FogAct.}
        \label{fig:threeclass}
    \end{subfigure}  
    \caption{
    Summary of FogAct statistics. (a) 91.8\% of samples last 3–12 seconds, resembling a normal distribution. (b) Action durations across classes show minimal variance, ensuring temporal balance. (c) Actions are grouped into three categories based on interaction patterns, reflecting the dataset’s diversity. Best viewed in color.
    }   
    \label{fig:comparison_all}
\end{figure*}

The action data are collected from volunteers who are fully informed about the intended use of the dataset, ensuring compliance with ethical standards and legal requirements. Approval for data collection was obtained from the appropriate local ethics committee. 
The proposed FogAct dataset covers both indoor and outdoor scenes—such as construction sites, academic buildings, and classroom playgrounds—and comprises 55 distinct real-world foggy actions grouped into three categories, as shown in Fig.~\ref{fig:threeclass}.
Each action video is recorded from four perspectives (front, back, left, and right) at 25 fps and 1920×1080 resolution. Videos are shot under light/dense fog across diverse scenes with varying camera positions. After data collection, all videos are segmented and annotated manually by two experienced experts. To ensure label reliability and consistency, a cross-validation process between annotators is conducted, and any inconsistencies are resolved through joint review and consensus.

\begin{table}[t]
    \caption{KL divergence between FogAct, three simulated datasets, and several real-world fog datasets, including VIREDA~\cite{duminil2023new}, BeDDE~\cite{zhao2020dehazing}, RHVD~\cite{chu2021real}, SOTS~\cite{li2018benchmarking}, and Foggy Zurich~\cite{sakaridis2018model}.}
    \centering
    \small
    \setlength{\tabcolsep}{4.1pt}
    \begin{tabular}{c|ccccc}
        \toprule
          & \textbf{VIREDA} &\textbf{BeDDE} &\textbf{RHVD} &  \textbf{SOTS} & \textbf{Zurich}\\ \midrule
          HMDB-51  &0.20 &0.24 & 0.78 &0.30 &2.79 \\
          UCF-101  &0.19 &0.22 & 0.44 &0.25 &3.38 \\
          Kinetics-100  &0.14 &0.17 & 0.52 &0.21 &3.21 \\
           \hline
         FogAct  &\textbf{0.13} &\textbf{0.15}  &  \textbf{0.42} &\textbf{0.19} &\textbf{1.47}\\
        \bottomrule
    \end{tabular}
    \label{tab:distribution}
\end{table}

\subsection{Data Quality Analysis}
The dataset contains 9724 videos, represented as triplets $\{(\mathbf{S}_{\text{f}}, \mathbf{S}_{\text{c}}, \mathbf{L}) \}$, where $\mathbf{S}_{\text{f}}$ and $\mathbf{S}_{\text{c}}$ denote the foggy and corresponding clean videos, respectively. $\mathbf{L}$ denotes the action category, annotated by two experts. Among them, 552 are dual-person action pairs and 4,310 are single-person action pairs, covering diverse human behaviors under real fog.
Fig.~\ref{fig:dur} shows 91.8\% of the actions have durations between 3 and 12 seconds, with an average length of 8.27 seconds. Fig.~\ref{fig:dur1} illustrates duration distributions by category. For training and evaluation, we randomly split the dataset into 80\% training and 20\% testing, ensuring robust model performance on unseen data.

\noindent{\bf KL Distribution.} In Tab.~\ref{tab:distribution}, we compute the KL divergence between our FogAct dataset and four real-world fog datasets: VIREDA~\cite{duminil2023new}, BeDDE~\cite{zhao2020dehazing}, RHVD~\cite{chu2021real}, and SOTS~\cite{li2018benchmarking}. We also compare the three simulated fog datasets: HMDB-51, UCF-101, and Kinetics-100, which are generated through ASM simulation, with the real-world fog datasets. To simplify the computation, we randomly sample two videos per action category from all datasets. The results indicate that our dataset exhibits a lower KL divergence with all real fog datasets, suggesting that FogAct better aligns with real-world fog distributions.

\noindent{\bf Evaluation of Vision-Language Models.} 
We assess fog realism using VLM-based metrics. With Q-Align~\cite{wu2023q}, FogAct obtains a realism score of 0.32, markedly higher than the synthetic dataset’s 0.13. GPT-4o yields similar trends, scoring 3.45 vs. 2.47, respectively. These results verify the superior realism of FogAct (see supplemental for prompts).

\noindent{\bf Human Evaluation.} 
To further assess the quality of FogAct, we conduct a human study with 21 participants (all graduate-level or above). Each participant is given 30 randomly sampled image pairs, each containing one FogAct image and one ASM-generated fog image. The study is blind, and participants rate which image has higher quality on a 1–5 scale (1 = lowest, 5 = highest). An example is shown in Fig.~7 of the supplementary material.
\section{FogNet}

\begin{figure*}[t]
    \centering
    \begin{overpic}[width=\textwidth]{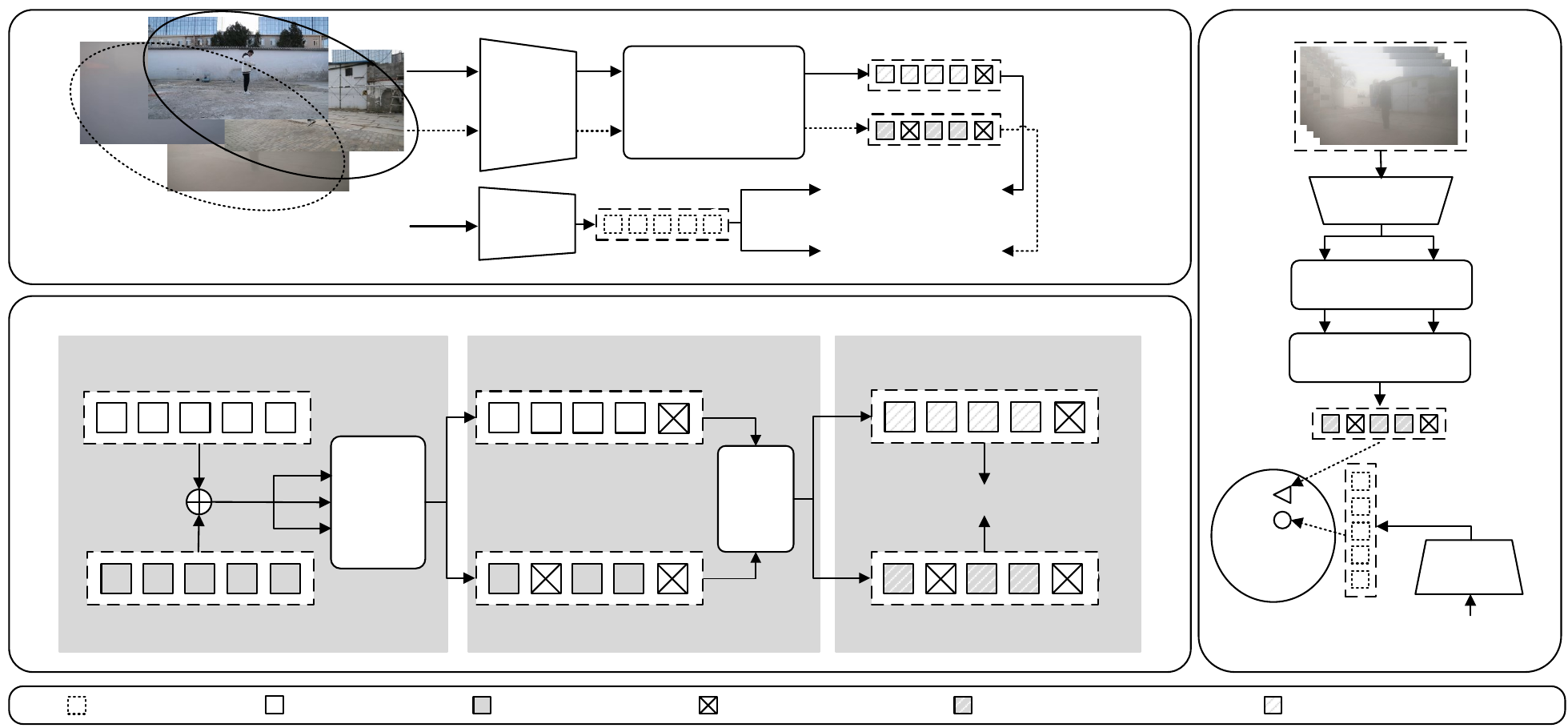}
        \put(52.2,33){\scriptsize \shortstack{\textbf{Clean Classification}\\\textbf{Loss} $\mathcal{L}_{\text{c}}$}}
        \put(52.8,29.35){\scriptsize \shortstack{\textbf{Fog Classification}\\\textbf{Loss} $\mathcal{L}_{\text{f}}$}}

        \put(28,43){\scriptsize $\mathbf{S}_{\text{c}}$}
        \put(28,39){\scriptsize $\mathbf{S}_{\text{f}}$}
        \put(6.5,35){\scriptsize \textbf{...}}
        \put(6.8,31.5){\scriptsize \textbf{This is a video of \textit{`XXX'}} $\{\mathbf{T}\}_{\text{c}=1}^{C}$}
        \put(40,29.5){\scriptsize $\{\mathbf{t}_{\text{c}}\}_{\text{c}=1}^{C}$}
        \put(78,44.0){\small \textbf{Inference}}
        \put(77.6,38.5){\scriptsize \shortstack{\textbf{Foggy}\\\textbf{Videos}}}
        \put(79,31){\scriptsize \textbf{Clone}}
        \put(40.2,37.8){\scriptsize \shortstack{\textbf{Fog-Invariant}\\\textbf{Feature Extractor}\\\textbf{via $\mathcal{L}_{\text{Temp}}$}}}
        \put(31.2,38.5){\scriptsize \textbf{\shortstack{Visual\\Encoder}}}
        \put(31.2,30.9){\scriptsize \textbf{\shortstack{Textual\\Encoder}}}
        \put(89.4,4.2){\scriptsize \textbf{\shortstack{This is a
        video \\ of \textit{`jump rope'}}}}
        \put(91,9){\scriptsize \textbf{\shortstack{Textual\\Encoder}}}
        \put(85.8,32.4){\scriptsize \textbf{\shortstack{Visual\\Encoder}}}
        \put(85,26.9){\scriptsize \textbf{\shortstack{Fog-Aware\\Selection}}}
        \put(84,22.4){\scriptsize \textbf{\shortstack{Mutual\\Enhancement}}}
        \put(79.3,4.5){\scriptsize \textbf{\shortstack{Shared\\Space}}}
        \put(81.5,19){\scriptsize \textbf{$\mathbf{v}_{\text{f}}^{\text{D}}$}}
        \put(88,11.5){\scriptsize \textbf{$\mathbf{t}_{\hat{\text{c}}}$}}
        \put(68,44){\small \textbf{Training}}
        \put(50.5,25.6){\small \textbf{Fog-Invariant Feature Extractor}}
        \put(10.4,23.8){\scriptsize \textbf{\textcolor{blue}{Fog-Aware Selection}}}
        \put(34.9,23.8){\scriptsize \textbf{\textcolor{blue}{Mutual Enhancement}}}
        \put(55.7,23.8){\scriptsize \textbf{\textcolor{blue}{Cross-Stream Alignment}}}
        \put(37.5,43){\scriptsize $\mathbf{v}_{\text{c}}$}
        \put(37.5,39){\scriptsize $\mathbf{v}_{\text{f}}$}
        \put(52.5,43){\scriptsize $\mathbf{v}_{\text{c}}^{\text{D}}$}
        \put(52.5,39){\scriptsize $\mathbf{v}_{\text{f}}^{\text{D}}$}
        \put(6.4,22){\scriptsize \textbf{Clean embedding} $\mathbf{v}_{\text{c}}$}
        \put(6.4,6.3){\scriptsize \textbf{Foggy embedding} $\mathbf{v}_{\text{f}}$}
        \put(32.9,22){\scriptsize \textbf{Selected Clean Embedding} $\mathbf{v}_{\text{c}}^{\text{A}}$}
        \put(32.9,6.3){\scriptsize \textbf{Selected Foggy Embedding} $\mathbf{v}_{\text{f}}^{\text{A}}$}
        \put(53.4,22){\scriptsize \textbf{Enhanced Clean Embedding} $\mathbf{v}_{\text{c}}^{\text{D}}$}
        \put(53.4,6.3){\scriptsize \textbf{Enhanced Foggy Embedding} $\mathbf{v}_{\text{f}}^{\text{D}}$}
        \put(47,12){\scriptsize \textbf{\rotatebox{90}{\shortstack{Cross\\Attention}}}}  
        \put(23.1,11.2){\scriptsize \textbf{\rotatebox{90}{\shortstack{Global self\\Attention}}}}  
        \put(61,14.2){\scriptsize \textbf{$\mathcal{L}_{\text{Temp}}$}}
        \put(5.6,1.1){\tiny \textbf{: textual tokens}}
        \put(18.2,1.1){\tiny \textbf{: clean tokens}}
        \put(31.45,1.1){\tiny \textbf{: foggy tokens}}
        \put(45.9,1.1){\tiny \textbf{: mask tokens}}
        \put(62.2,1.1){\tiny \textbf{: foggy-guided fusion tokens}}
        \put(82.0,1.1){\tiny \textbf{: clean-guided fusion tokens}}
    \end{overpic}
    \caption{An overview of our framework. In the joint training stage, we jointly learn label supervision and fog-invariant representations across clean and foggy videos. Fog-invariant feature extractor involves three key components: i) Fog-Aware Selection. $\mathbf{v}_{\text{c}}$ and $\mathbf{v}_{\text{f}}$ are fused via global self-attention to highlight semantically relevant regions.
    ii) Mutual Enhancement. Bidirectional cross-attention enables semantic interaction between $\mathbf{v}_{\text{c}}^{\text{A}}$ and $\mathbf{v}_{\text{f}}^{\text{A}}$ while retaining domain-specific traits.
    iii) Cross-Stream Alignment. $\mathbf{v}_{\text{c}}^{\text{D}}$ and $\mathbf{v}_{\text{f}}^{\text{D}}$ are aligned at the frame level, with the latter used for recognition. 
    During inference, only foggy videos are used as input and serve as the query, key, and value for the Fog-Aware Selection, followed by Mutual Enhancement to obtain $\mathbf{v}_{\text{f}}^{\text{D}}$.
    Prediction $\hat{\text{c}}$ is made by the nearest text embedding $\mathbf{t}_{\hat{\text{c}}}$ to $\mathbf{v}_{\text{f}}^{\text{D}}$.}
    \label{fig:framework}
\end{figure*}

\noindent{\bf Overview.}~We extract fog-invariant embeddings via the fog-invariant feature extractor, which integrates fog-aware selection, mutual enhancement, and cross-stream alignment.
Given a foggy video sequence $\mathbf{S}_{\text{f}} = \{\mathbf{I}_{\text{f}}\}_{{i}=1}^{T}$ and its clean counterpart $\mathbf{S}_{\text{c}} = \{\mathbf{I}_{\text{c}}\}_{{i}=1}^{T}$, each with $T$ uniformly sampled frames, we encode them into visual embeddings, $\mathbf{v}_{\text{f}}$ and $\mathbf{v}_{\text{c}}$, using a visual encoder.
The fog-aware selection component employs global self-attention to identify semantically meaningful embeddings $\mathbf{v}_{\text{f}}^{\text{A}}$ and $\mathbf{v}_{\text{c}}^{\text{A}}$ from both streams. 
Next, the mutual enhancement component refines two embeddings via bidirectional cross-attention to obtain $\mathbf{v}_{\text{f}}^{\text{D}}$ and $\mathbf{v}_{\text{c}}^{\text{D}}$. 
Finally, leveraging the inherent temporal consistency between foggy and clean videos, we perform frame-level alignment between
$\mathbf{v}_{\text{f}}^{\text{D}}$ and $\mathbf{v}_{\text{c}}^{\text{D}}$.
To classify foggy videos, $C$ action prompts ${\mathbf{T}}_{\text{c}=1}^{C}$ are embedded as $\{\mathbf{t}_{\text{c}}\}_{\text{c}=1}^{C}$ via a textual encoder.
The classification is performed by selecting the class $\hat{\text{c}}$ whose textual embedding exhibits the highest cosine similarity with the fog-invariant embedding, \ie, $\hat{\text{c}} = \mathop{\arg \max} \text{sim}(\mathbf{v}_{\text{f}}^{\text{D}}, \mathbf{t}_{\text{c}})$. Fig.~\ref{fig:framework} illustrates our framework. 

\subsection{Fog-Aware Selection}

Besides irrelevant background cues shared by clean and foggy embeddings, $\mathbf{v}_{\text{f}}$ also contains fog-induced degradation. Directly aligning $\mathbf{v}_{\text{f}}$ with text embeddings introduces noise and weakens classification performance.

\begin{table*}[t]
    \centering
    \small
    \caption{Comparison results on the FogAct dataset. We report the GFLOPs and parameters in the inference phase. `Views' indicates \# temporal clip $\times$ \# spatial crop. `LLaVA1.5-VI' indicates LLaVA1.5-VideoChatGPT-Instruct. The magnitude is Million ($10^{6}$) for parameters (Param). `Pre-training’ indicates training data: for the backbone in one-stage, and the defogger in two-stage methods. 
    For methods with two stages, a defogging method is first applied to remove the fog, followed by the use of the SOTA action recognition method: OST~\cite{chen2024ost} for classification.
    We achieve the highest Top-1 and Top-5 accuracy by employing an 8-frame RGB input evaluated with a single view. The best numbers are highlighted in \textbf{\textit{bold}}.
    }
    \hspace{-0.31cm} 
    \scalebox{1}{
    \setlength{\tabcolsep}{9.5pt}
    \setlength{\tabcolsep}{5pt}
  \begin{tabular}{lccccccccc}
\toprule
  \textbf{Method}  & \textbf{Venue} & \textbf{Input}   & \textbf{Pre-training} & \textbf{Top-1(\%)} & \textbf{Top-5(\%)} &  \textbf{Views} & \textbf{GFLOPs} & \textbf{Param}  \\
  \midrule
 \multicolumn{9}{l}{\emph{Methods with one stage}} \\
 \midrule
  VideoMAE ViT-B/16~\cite{tong2022videomae} & NeurIPS'22  & 8\x224$^2$  & Kinetics-400 & 11.1 & 31.1 &1\x1  & 90 & 87\\
  DINO-v2~\cite{oquab2023dinov2} & arXiv'23  & 8\x224$^2$  & LVD-142M & 45.4 & 74.2 &1\x1  & - &86.6 \\
  LLaMa-VID~\cite{li2024llama} & ECCV'24  & 8\x224$^2$  & LLaVA1.5-VI & 26.2 & - &-  &  -&7000 \\
  X-CLIP ViT-B/16~\cite{ni2022expanding} & ECCV'22  & 8\x224$^2$  & Kinetics-400 & 67.4 & 92.8 &1\x1  & 145 & 131.5\\
  ActionCLIP ViT-B/16~\cite{wang2021actionclip}  & TNNLS'23 & 8\x224$^2$  & WIT-400M & 75.0 & 95.7 &1\x1  & 141 & 141.7 \\ 
  AIM ViT-B/16~\cite{yang2023aim} & ICLR'23 & 8\x224$^2$  & WIT-400M & 80.3 & 98.0 & 1\x1 & 202 & 96.4\\
  ATM ViT-B/16~\cite{wu2023can} & ICCV'23 & 8\x224$^2$  & WIT-400M & 73.2 & 94.5 & 1\x1 & 95 & 87.8\\
  Vita-CLIP ViT-B/16~\cite{wasim2023vita} & CVPR'23 & 8\x224$^2$  & WIT-400M & 18.8 & 43.8 &1\x1  & 97 & 161.8\\
  ViFi-CLIP ViT-B/16~\cite{rasheed2023fine}& CVPR'23   & 8\x224$^2$  & WIT-400M & 78.4 & 97.7 &1\x1  &141  &124.7 \\
  BIKE ViT-B/16~\cite{wu2023bidirectional}& CVPR'23   & 8\x224$^2$  & WIT-400M & 13.4 & 19.5 &1\x1  & - & 124.1\\
  M$^2$-CLIP ViT-B/16~\cite{wang2024multimodal} & AAAI'24  & 8\x224$^2$  & WIT-400M & 56.5 & 88.8 &1\x1  & 214 &165 \\
  TC-CLIP ViT-B/16~\cite{kim2024leveraging} & ECCV'24  & 8\x224$^2$  & WIT-400M & 71.5 & 95.0 &1\x1  & - &127.5 \\
  SFAR ViT-B/16~\cite{liu2024storyboard} &NeurIPS'25 & 8\x224$^2$  & WIT-400M & 73.6 &95.4 & 1\x1 & 90.2 & 126.1 \\
  OST ViT-B/16~\cite{chen2024ost} & CVPR'24 & 8\x224$^2$  & WIT-400M & 83.2 & 98.9 & 1\x1 & 141 & 149.6 \\
  \midrule
 \multicolumn{9}{l}{\emph{Methods with two stages}} \\
 \midrule
   UCL~\cite{wang2024ucl} + OST &  TIP'24 & 8\x224$^2$  & Unsupervised & 58.0 & 86.9 &1\x1  &-  &169.1 \\  
   PTTD~\cite{chen2024prompt} + OST &  ECCV'24  & 8\x224$^2$  & DIV\&Flickr & 85.2 & 98.2 &1\x1  & - &152.2 \\
    LDR~\cite{yang2024language} + OST &  CVPR'24 & 8\x224$^2$  & All-weather & 83.3 & 98.6 &1\x1  & 776 &163.6 \\
    PTTD~\cite{chen2024prompt} + OST &  ECCV'24  & 8\x224$^2$  & FogAct & 85.4 & 98.8 &1\x1  & - &152.2 \\

  \midrule
  \multirow{1}{*}{\textbf{Ours~ViT-B/16}} & \multirow{1}{*}{2025} & 8\x224$^2$  & \multirow{1}{*}{WIT-400M} & \textbf{88.7} & \textbf{99.4} & \textbf{1\x1} & 143 & 146.9 \\ 
\bottomrule
  \end{tabular}}
    \label{tab:k400_sota}
\end{table*}

To mitigate this, we apply a global self-attention module to jointly process $\mathbf{v_{\text{c}}}$ and $\mathbf{v_{\text{f}}}$, emphasizing features that are stable in clean conditions and discriminative under fog. This suppresses fog artifacts and strengthens semantic representations. The process is given by

\allowdisplaybreaks
\begin{align}
\mathbf{v_{\text{att}}} = \sum_{i=1}^{T} \frac{\exp\Big(\mathrm{sim}\big(\mathbf{v^{\text{cat}}}, \mathbf{v^{\text{cat}}}_i\big)\Big)}{\sum_{j=1}^{T}\exp\Big(\mathrm{sim}\big(\mathbf{v^{\text{cat}}}, \mathbf{v^{\text{cat}}}_j\big)\Big)} \mathbf{v}^{\text{cat}}_{i},
\end{align}
where $\mathbf{v^{\text{cat}}} = [\mathbf{v_{\text{c}}}; \mathbf{v_{\text{f}}}]$ denotes concatenated clean and foggy embeddings, and $\mathbf{v}^{\text{cat}}_{{i}}$ represents the $i$-th element of $\mathbf{v}^{\text{cat}}$. We then split $\mathbf{v_{\text{att}}}$ back into separate clean $\mathbf{v}_{\text{c}}^{\text{A}}$ and foggy $\mathbf{v}_{\text{f}}^{\text{A}}$ as

\begin{equation}
\mathbf{v}_{\text{c}}^{\text{A}}, \mathbf{v}_{\text{f}}^{\text{A}} = \text{chunk} \left( \mathbf{v_{\text{att}}}, 2 \right).
\end{equation}

\subsection{Mutual Enhancement}
While the selected $\mathbf{v}_{\text{f}}^{\text{A}}$ and $\mathbf{v}_{\text{c}}^{\text{A}}$ capture important and discriminative features, this process may miss some key information. The clean embedding helps guide the optimization of the foggy embedding to some extent, and vice versa.

To enhance fog-invariant representations, bidirectional cross-modal attention mutually refines $\mathbf{v}_{\text{f}}^{\text{A}}$ and $\mathbf{v}_{\text{c}}^{\text{A}}$, with the clean embedding querying the foggy one for refinement.
\begin{align}
    &\mathbf{Q}_{\text{c}} = \mathbf{v}_{\text{c}}^{\text{A}} \mathbf{W}^{\text{q}}, \quad \mathbf{K}_{\text{f}} = \mathbf{v}_{\text{f}}^{\text{A}} \mathbf{W}^{\text{k}}, \quad \mathbf{V}_{\text{f}} = \mathbf{v}_{\text{f}}^{\text{A}}\mathbf{W}^{\text{v}}, \\ &\mathbf{v}_{\text{f}}^{\text{D}} = \text{Attention}(\mathbf{Q}_{\text{c}}, \mathbf{K}_{\text{f}}, \mathbf{V}_{\text{f}}) + \mathbf{v}_{\text{f}}^{\text{A}},
\end{align}
where $\mathbf{W}^{\text{q}}$, $\mathbf{W}^{\text{k}}$, and $\mathbf{W}^{\text{v}}$ project the query $\mathbf{Q}_{\text{c}}$, key $\mathbf{K}_{\text{f}}$, and value $\mathbf{V}_{\text{f}}$. We then reverse the roles, using foggy embeddings as the query and clean embeddings as key and value,
\begin{equation}
    \mathbf{v_{\text{c}}^{\text{D}}} = \text{Attention}(\mathbf{Q}_{\text{f}}, \mathbf{K}_{\text{c}}, \mathbf{V}_{\text{c}}) + \mathbf{v}_{\text{c}}^{\text{A}}.
\end{equation}

\subsection{Cross-Stream Alignment}
Considering fog does not distort action semantics, temporal dynamics should be preserved for fog-invariant embeddings.
To achieve this, we employ cross-stream alignment to align the enhanced clean embedding $\mathbf{v_{\text{c}}^{\text{D}}}$ and the foggy embedding $\mathbf{v_{\text{f}}^{\text{D}}}$, which extracts fog-invariant embeddings by maintaining temporal consistency across varying conditions.
Concretely, we construct a consistency matrix $s^{\text{c}}$ to align foggy and clean video embeddings at the frame-level
\begin{equation}
    s^{\text{c}} =  \mathrm{sim}\left(\mathbf{v_{\text{f}}^{\text{D}}}, \mathbf{v_{\text{c}}^{\text{D}}}\right).
\end{equation}

Each element $s_{i,j}^{\text{c}}$ represents the correlation between the $i$-th frame of the foggy sequence and the $j$-th frame of the clean sequence. 
To learn temporal alignments between clean and foggy videos, we optimize with a contrastive loss,
\begin{equation}
\mathcal{L}_{\text{Temp}} = \frac{1}{T} \sum_{i \in T} \log \frac{\exp(s_{i, i}^{\text{c}})}{\sum_{{j}=1}^{T} \exp(s_{i, j}^{\text{c}})} \ .
\end{equation}

\subsection{Loss}
We optimize the network to maximize the similarity $s_{\text{b}, \text{c}^{*}}$ between each foggy video embedding and its ground-truth text embedding, while suppressing similarities to all other classes $\{s_{\text{b}, \text{c}}\}_{\text{c}=1, \text{c} \neq \text{c}^{*}}^{C}$. Following \cite{infonceloss}, we adopt the InfoNCE loss.

\begin{equation}
\mathcal{L}_{\text{f}}^{\text{T2V}} =  \frac{1}{B} \sum_{b=1}^{B} \frac{1}{|\mathbf{k}_{\text{b}}|} \sum_{k \in \mathbf{k}_{\text{b}}} \log \frac{\exp(s_{{k}, \text{c}^{*}})}{\sum_{j=1}^{B} \exp(s_{{j}, \text{c}^{*}})}
\end{equation}
\begin{equation}
\mathcal{L}_{\text{f}}^{\text{V2T}} =  \frac{1}{B} \sum_{{b}=1}^{B} \frac{1}{|\mathbf{k}_{\text{b}}|} \sum_{{k} \in \mathbf{k}_{\text{b}}} \log \frac{\exp(s_{{k}, \text{c}^{*}})}{\sum_{\text{c}=1}^{C} \exp(s_{{k}, \text{c}})}
\end{equation}
\begin{equation}
\mathcal{L}_{\text{f}} = \mathcal{L}_{\text{f}}^{\text{T2V}} +  \mathcal{L}_{\text{f}}^{\text{V2T}}
\end{equation}
{where $B$ is the batch size, $\mathbf{k}{\text{b}}$ denotes indices of videos sharing the same class as the ${b}\hbox{-}\text{th}$ video, and $|\mathbf{k}{\text{b}}|$ is their count.}

Similarly, we define the loss for aligning text embeddings with clean video features as $\mathcal{L}_{\text{c}} = \mathcal{L}_{\text{c}}^{\text{T2V}} + \mathcal{L}_{\text{c}}^{\text{V2T}}$, computed analogously to $\mathcal{L}_{\text{f}}^{\text{T2V}}$ and $\mathcal{L}_{\text{f}}^{\text{V2T}}$ but using clean instead of foggy embeddings.
The total loss is formulated as:
\begin{align}
    \mathcal{L}_{\text{all}} = \mathcal{L}_{\text{f}} + \lambda \mathcal{L}_{\text{c}} + \beta \mathcal{L}_{\text{Temp}}, 
\end{align}
where $\lambda=0.4$, $\beta=0.1$ are hyperparameters.

\section{Experiment}
\subsection{Experiment Setup}
\noindent{\bf Datasets.} We conduct experiments on four benchmarks: the collected FogAct dataset and three simulated datasets (apply ASM~\cite{narasimhan2003contrast} to UCF-101~\cite{soomro2012dataset}, HMDB-51~\cite{kuehne2011hmdb}, and Kinetics-100~\cite{kay2017kinetics} following \cite{tanneru2021action}).
In addition, two experts manually annotate the real-world foggy action videos.

\begin{table}[t]
    \centering
    \caption{Comparison on the synthesized datasets, UCF-101~\cite{soomro2012dataset}, HMDB-51~\cite{kuehne2011hmdb}, and Kinetics-100~\cite{kay2017kinetics}.}
    \small
    \setlength{\tabcolsep}{5pt}
        \begin{tabular}{lcccc}
                \toprule
                \textbf{Method} &  \textbf{UCF} &  \textbf{HMDB} & \textbf{Kinetics} & \textbf{Average} \\ \midrule
                ActionCLIP~\cite{wang2021actionclip} & 84.3 & 57.1 & 70.6 &70.7  \\
                AIM~\cite{yang2023aim} & 90.0 & 64.7 & 79.2 & 78.0\\
                ATM~\cite{wu2023can} & 90.2 & 63.4 & 77.2 &76.9 \\
                M$^2$-CLIP~\cite{wang2024multimodal}& 89.0 & 62.0 & 78.0 & 76.3 \\
                OST~\cite{chen2024ost}&92.4  & 69.1 & 75.4 & 79.0 \\
                \midrule
                \textbf{Ours} & \textbf{93.2} &\textbf{71.1} &\textbf{85.2}  & \textbf{83.2}\\
                \bottomrule
            \end{tabular}
    \label{tab:super}
\end{table}

\noindent{\bf Implementation Details.} 
We initialize our network with CLIP~\cite{radford2021learning} pre-trained on WIT-400M. Then, we fine-tune the model for 30 epochs using the AdamW optimizer with a batch size of 128. Learning rate is $5 \times 10^{-5}$ with cosine annealing and 5-epoch warm-up.

\begin{table*}[t]
\centering
\caption{Comparative experiments are conducted on UCF-101 , HMDB-51 , Kinetics-100 (denoted as `UCF', `HMDB', and `K100', respectively), and the FogAct dataset. We report accuracy (\%) for a single 8-frame clip with a spatial resolution of 224×224, unless otherwise specified. `Pers.' indicates perspectives. `Sim.' indicates simulated.
}
\hspace{-5mm}
\raisebox{3mm}{
\begin{subtable}[t]{0.34\textwidth}
    \centering
    \small
    \setlength{\tabcolsep}{3pt}
    \begin{tabular}{c c c|c c}
            \toprule
            \textbf{FAS} &  \textbf{ME} &  \textbf{CSA} & \textbf{Top-1(\%)} &\textbf{Top-5(\%)} \\
            \midrule
            \multicolumn{3}{c|}{Baseline} &75.0  &95.7  \\
            \midrule
            \cmark & \xmark & \xmark &86.9  & 98.9 \\
            \xmark & \cmark & \xmark & 79.4 & 97.2 \\
            \xmark & \xmark & \cmark & 84.1 & 98.5 \\
            \cmark & \cmark & \xmark & 88.1 & 99.4 \\
            \cmark & \cmark & \cmark & \textbf{88.7} & \textbf{99.4} \\
            \bottomrule
        \end{tabular}
        \caption{The effectiveness of our model components.
    }
    \label{tab:comp}  
\end{subtable}
}
\hspace{-3mm}
\begin{subtable}[t]{0.46\textwidth}
    \centering
    \small
    \raisebox{3mm}{ 
    \setlength{\tabcolsep}{4pt}
    \begin{tabular}{c |c c c c}
                \toprule
                \textbf{Model} & \textbf{Train}  &\textbf{Test} & \textbf{Top-1(\%)} & \textbf{Top-5(\%)}\\ 
                \midrule
                \multirow{5}{*}{ActionCLIP} & clean &clean  & 86.7  & 98.9 \\
                & foggy &foggy  & 75.0   & 95.7 \\
                & clean &foggy  & 59.8  & 85.9 \\
                & clean+foggy &foggy  & 75.7 & 96.3 \\
                \midrule
                \multirow{2}{*}{FogNet} 
                & clean+sim. & foggy  & 71.5 & 93.6\\
                & clean+foggy & foggy  & \textbf{88.7} & \textbf{99.4}\\
                \bottomrule
            \end{tabular}
            }
            \caption{Comparison for different input settings on FogAct.}
    \label{tab:motivation}  
\end{subtable}
\hspace{1mm}
\raisebox{3.2mm}{
\begin{subtable}[t]{0.18\textwidth}
    \centering
    \small
    \raisebox{7.6mm}{
    \setlength{\tabcolsep}{4pt}
    \begin{tabular}{c c c}
        \toprule
        \textbf{Pers.}  & \textbf{Top-1} & \textbf{Top-5} \\ \midrule
        Single  & 52.7 & 79.4 \\
        \midrule
        Four    & \textbf{88.7} & \textbf{99.4} \\
        \bottomrule
    \end{tabular}
    }
    \caption{
     Results of models trained on single or four perspectives and evaluated on four.
    }
    \label{tab:single}
\end{subtable}
}
\\
\vspace{-0.122cm}
\hspace{-5mm}
\begin{subtable}[t]{0.35\textwidth}
    \centering
    \small
    \raisebox{3mm}{ 
    \setlength{\tabcolsep}{3pt}
    \begin{tabular}{c c c c c}
        \toprule
         \textbf{Backbone}  &  \textbf{Top-1} & \textbf{+ FAS} &\textbf{+ ME} &\textbf{+ CSA} \\ \midrule
          RN50 & 56.1 &70.2  &71.0 &  71.5\\
         ViT-B/32  & 63.0 & 78.8 &79.0 & 79.2 \\
        ViT-B/16  & 75.0 &86.9  &88.1  & 88.7\\
        \bottomrule
    \end{tabular}
    }
    \caption{Evaluation of using backbones with different components. The frame rate is set to 8.}
    \label{tab:backbone}
\end{subtable}
\hspace{-2mm}
\begin{subtable}[t]{0.38\textwidth}
    \centering
    \small
    \raisebox{1mm}{ 
    \setlength{\tabcolsep}{2pt}
    \begin{tabular}{c c c c c c}
        \toprule
           \textbf{Frames} &  \textbf{Views} & \textbf{FogAct} &$\textbf{HMDB}$  &$\textbf{UCF}$  &$\textbf{K100}$ \\ \midrule
        4 & 1\x1 & 85.2 &69.6  &92.5 &82.9 \\
        4 & 4\x1 & 86.6 & 70.4 &93.3 & 85.1\\
        8 & 1\x1 & 88.7 & \textbf{71.1} &93.2 &85.1 \\
        8 & 4\x1 & \textbf{89.1} & 70.9 &\textbf{93.4} &\textbf{85.4} \\
        \bottomrule
    \end{tabular}
    }
    \caption{Effects of different inference schemes. 
    }
    \label{tab:views}
\end{subtable}
\hspace{-1mm}
\begin{subtable}[t]{0.28\textwidth}
    \centering
    \small
     \raisebox{2mm}{
    \setlength{\tabcolsep}{2pt}
    \begin{tabular}{c c c c}
        \toprule
        \textbf{Intensity} & \textbf{Views} & \textbf{Top-1(\%)} & \textbf{Top-5(\%)} \\ \midrule
        Light & 1×1 & 66.0 & 87.2 \\
        Dense   & 1×1 & 67.5 & 92.2 \\
        \midrule
        Multi   & 1×1 & \textbf{88.7} & \textbf{99.4} \\
        \bottomrule
    \end{tabular}
    }
    \caption{
    Results of different fog levels.
    }
    \label{tab:multi}
\end{subtable}

		\label{tab:all}
\end{table*}

\noindent{\bf Compared Methods.}~We compare two paradigms: \textbf{one-stage} and \textbf{two-stage}. One-stage methods directly recognize actions in foggy videos, while two-stage methods first defog the videos and then apply OST~\cite{chen2024ost}. Unless otherwise specified, ablations are conducted on FogAct with ActionCLIP~\cite{wang2021actionclip} as the default baseline.

\begin{table}[t]
		\centering
        \captionsetup{type=figure}
\begin{minipage}[t]{\textwidth}
        \begin{minipage}[t]{0.8\textwidth}
            \begin{tabular}{cccccc}
            \vspace{-0.03cm}
            \vspace{-0.1cm}
            \hspace{-0.25cm}\raisebox{0.25cm}{\rotatebox{90}{\parbox[c]{1.cm}{\centering \scalebox{0.8}{\small\text{Clean}}}}} &
            \hspace{-0.43cm}
            \begin{tikzpicture}
              \node[inner sep=0pt] (img) at (0,0)
                {\includegraphics[width=1.6cm, height=1.6cm]{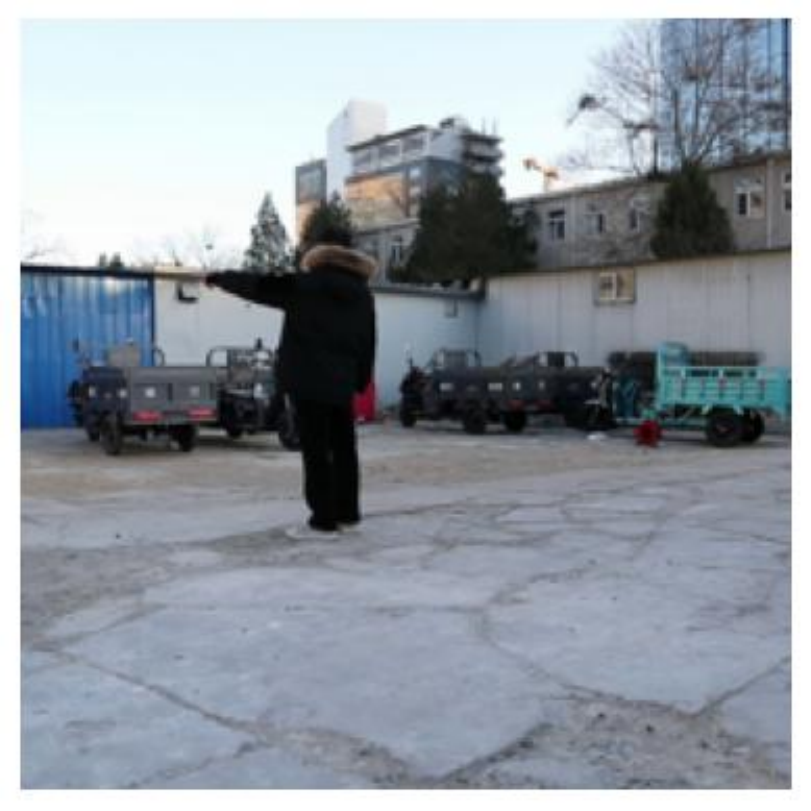}};
            \end{tikzpicture}
            & 
            \hspace{-0.48cm}
            \begin{tikzpicture}
              \node[inner sep=0pt] (img) at (0,0)
                {\includegraphics[width=1.6cm, height=1.6cm]{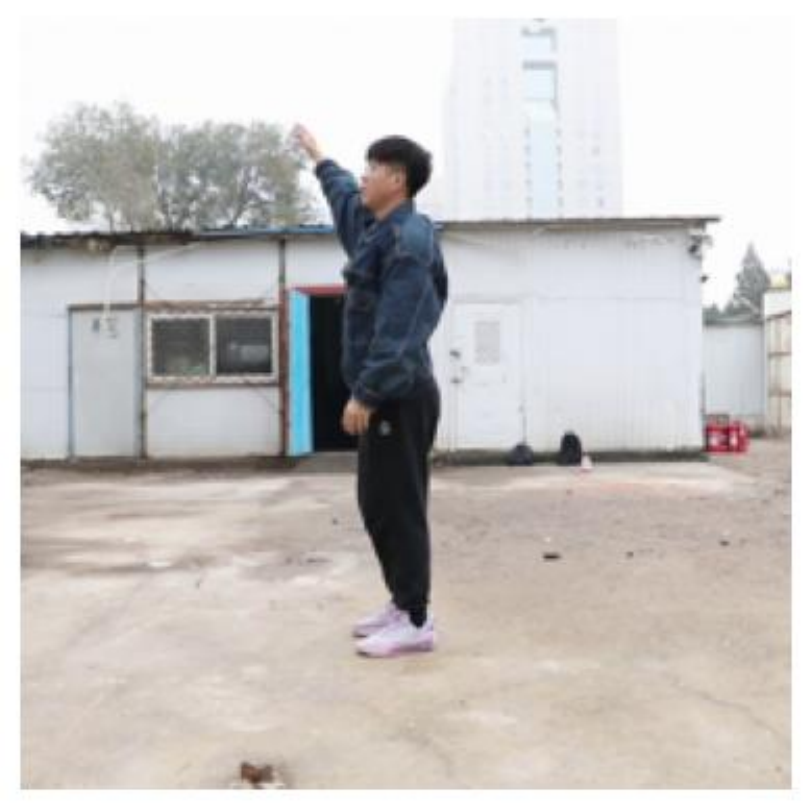}};
              \fill[black, opacity=1] (-0.08,0.39) rectangle (0.0,0.48);
            \end{tikzpicture}
            &
            \hspace{-0.48cm}
            \begin{tikzpicture}
              \node[inner sep=0pt] (img) at (0,0)
                {\includegraphics[width=1.6cm, height=1.6cm]{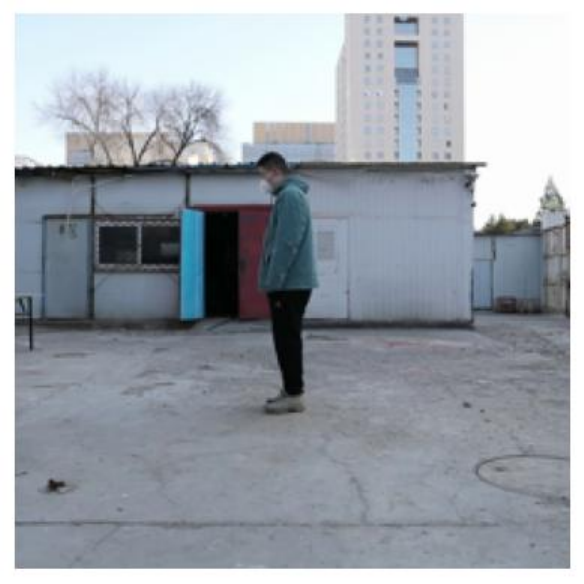}};
              \fill[black, opacity=1] (-0.1,0.28) rectangle (-0.02,0.35);
            \end{tikzpicture}
            &
            \hspace{-0.48cm}
            \begin{tikzpicture}
              \node[inner sep=0pt] (img) at (0,0)
                {\includegraphics[width=1.6cm, height=1.6cm]{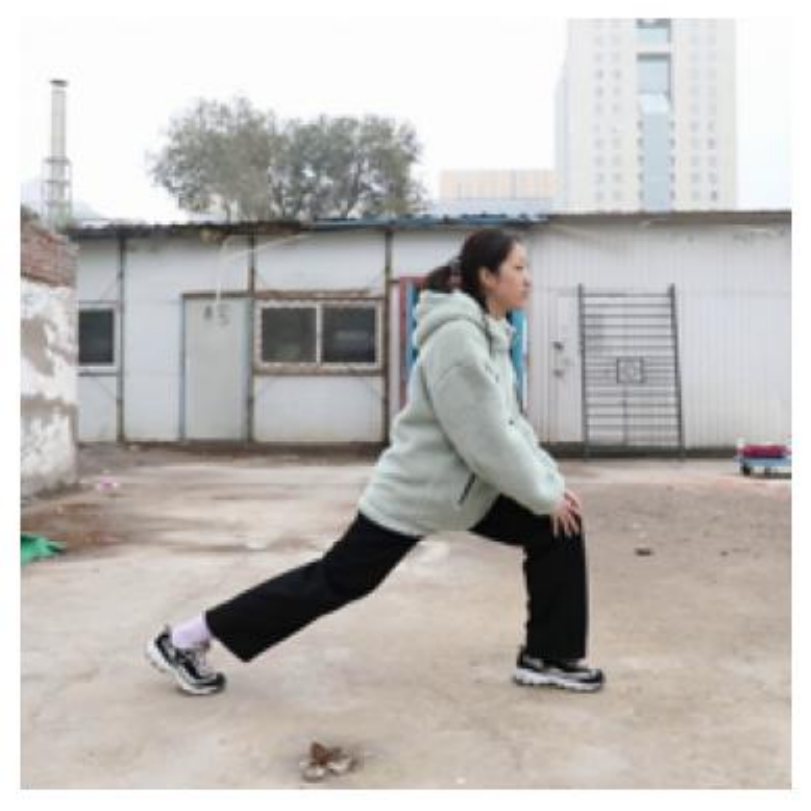}};
              \fill[black, opacity=1] (0.18,0.2) rectangle (0.25,0.3);
            \end{tikzpicture}
            &
            \hspace{-0.48cm}
            \begin{tikzpicture}
              \node[inner sep=0pt] (img) at (0,0)
                {\includegraphics[width=1.6cm, height=1.6cm]{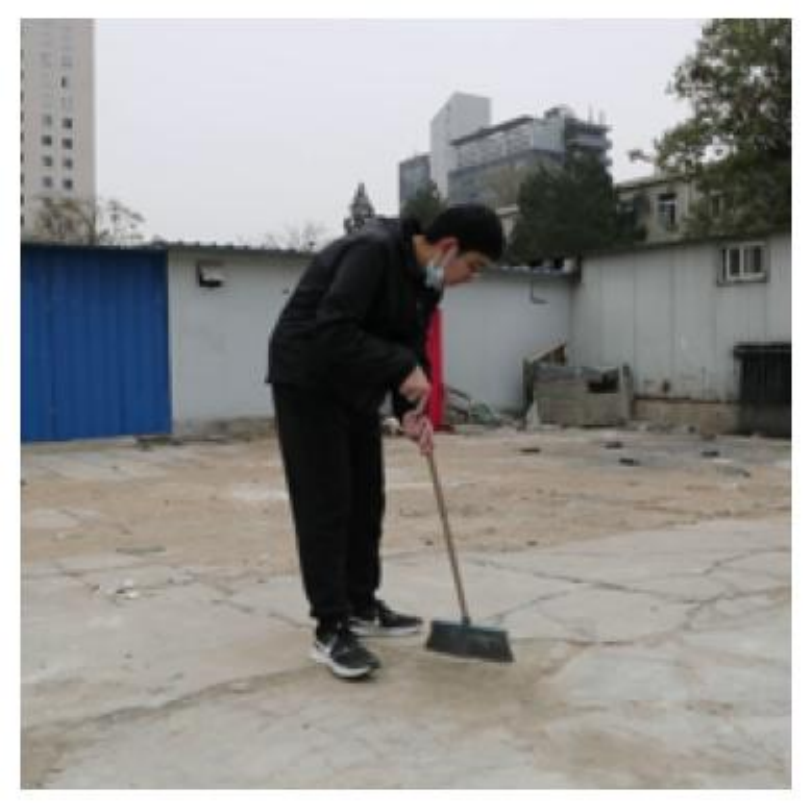}};
              \fill[black, opacity=1] (0.09,0.25) rectangle (0.16,0.3);
            \end{tikzpicture}
            \\
            \vspace{-0.03cm}
            \vspace{-0.1cm}
            \hspace{-0.25cm}\raisebox{0.25cm}{\rotatebox{90}{\parbox[c]{1.cm}{\centering \scalebox{0.8}{\small\text{Foggy}}}}} &
            \hspace{-0.35cm}\includegraphics[width =1.6cm, height=1.6cm]{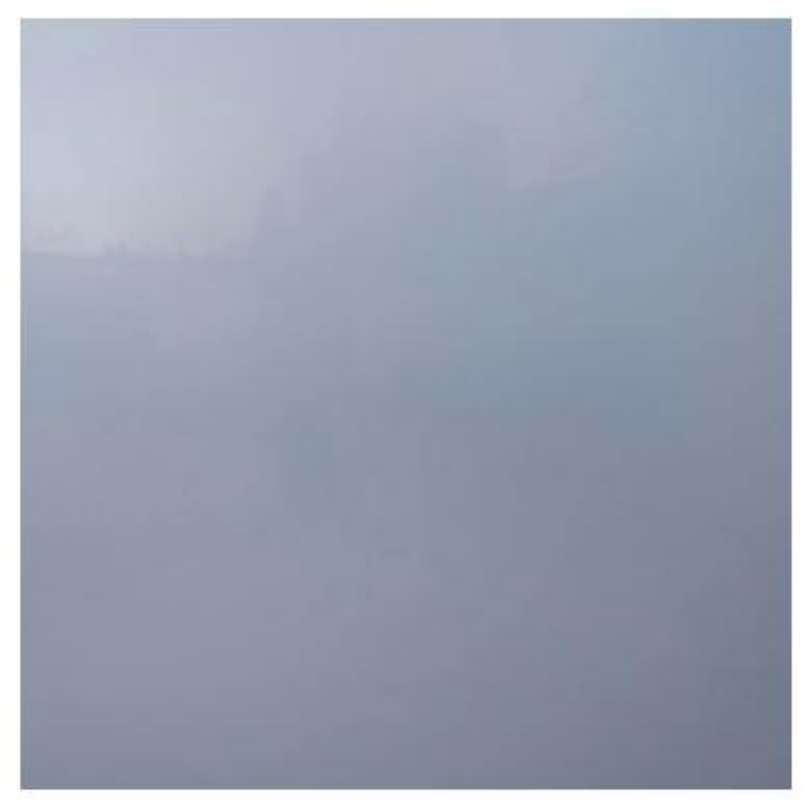} & 
            \hspace{-0.4cm}\includegraphics[width =1.6cm,height=1.6cm]{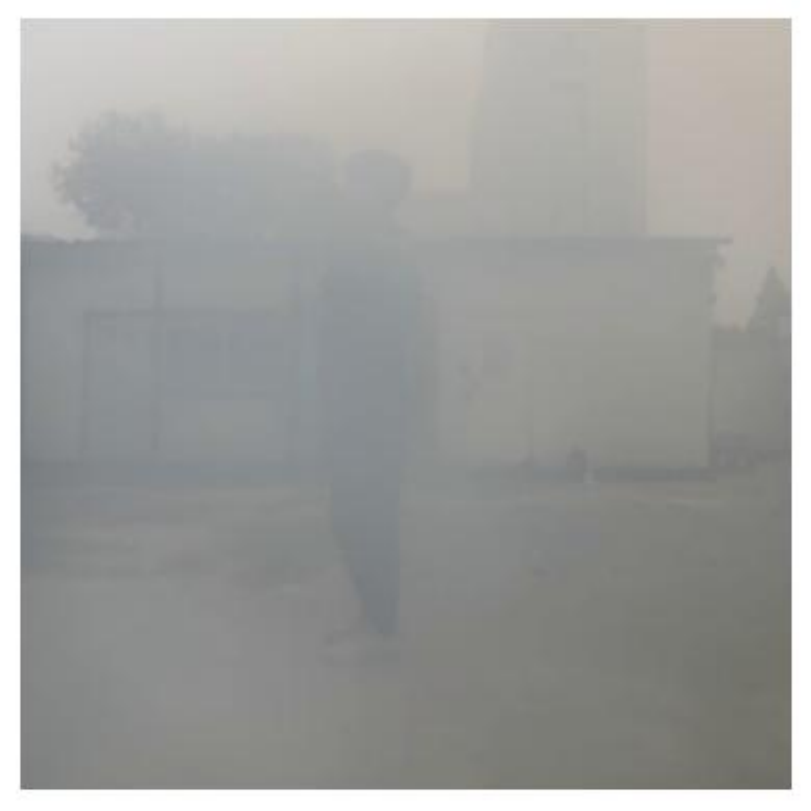} & \hspace{-0.4cm}\includegraphics[width =1.6cm,height=1.6cm]{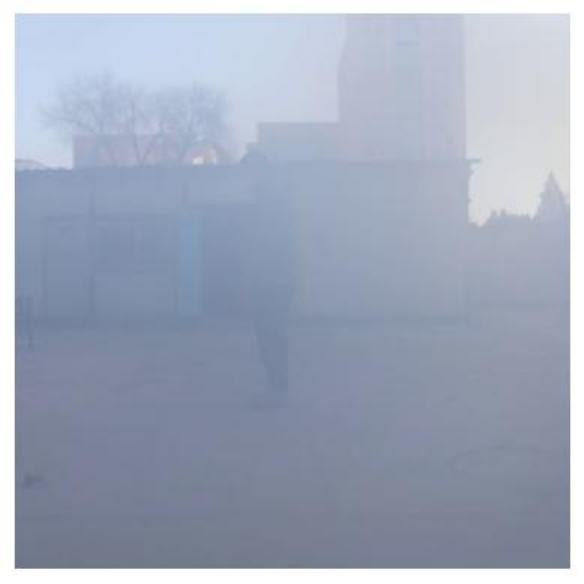} & 
            \hspace{-0.4cm}\includegraphics[width =1.6cm,height=1.6cm]{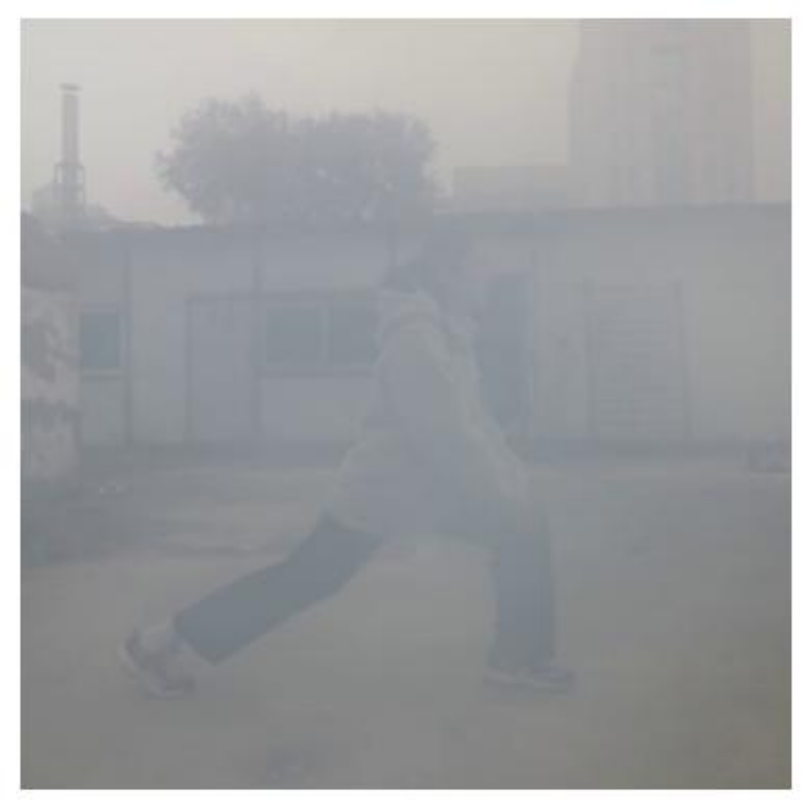} &
            \hspace{-0.4cm}\includegraphics[width =1.6cm,height=1.6cm]{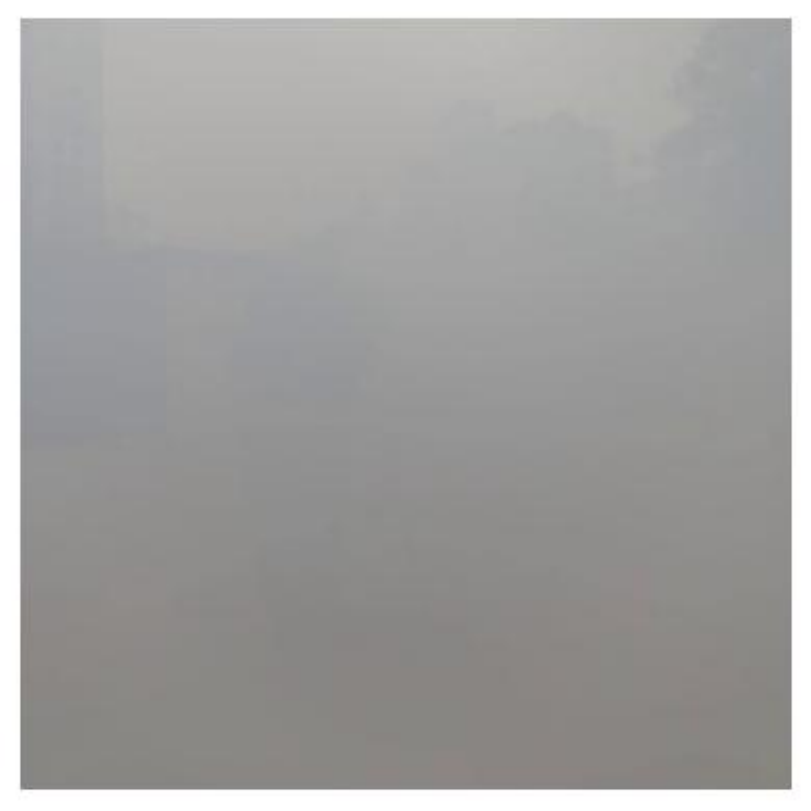}
            \\
            \vspace{-0.03cm}
            \vspace{-0.1cm}
            \hspace{-0.25cm}\raisebox{0.25cm}{\rotatebox{90}{\parbox[c]{1.cm}{\centering \scalebox{0.8}{\small\text{Baseline}}}}} &
            \hspace{-0.35cm}\includegraphics[width =1.6cm, height=1.6cm]{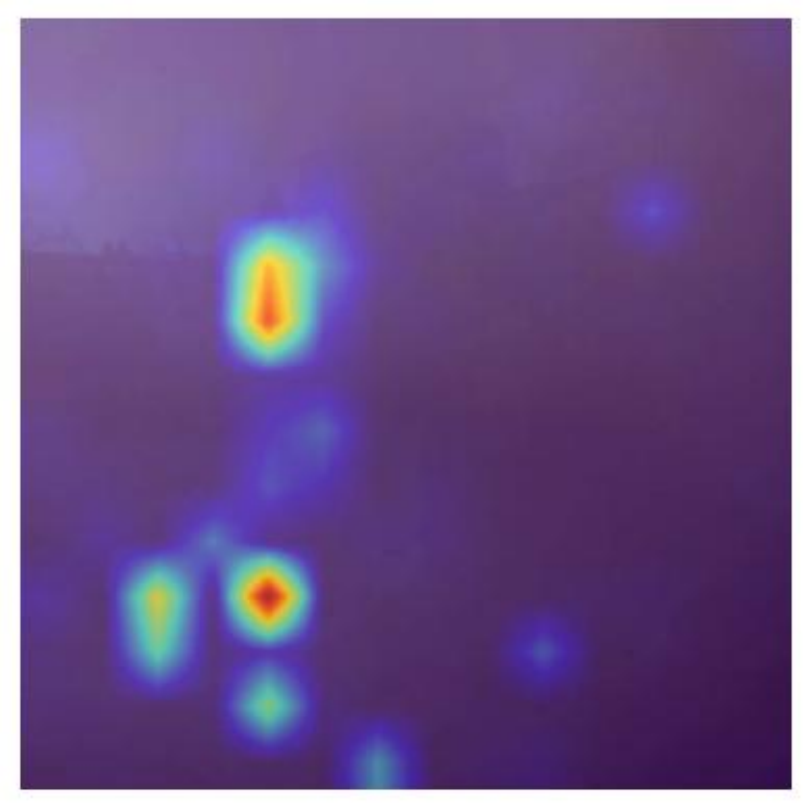} & 
            \hspace{-0.4cm}\includegraphics[width =1.6cm,height=1.6cm]{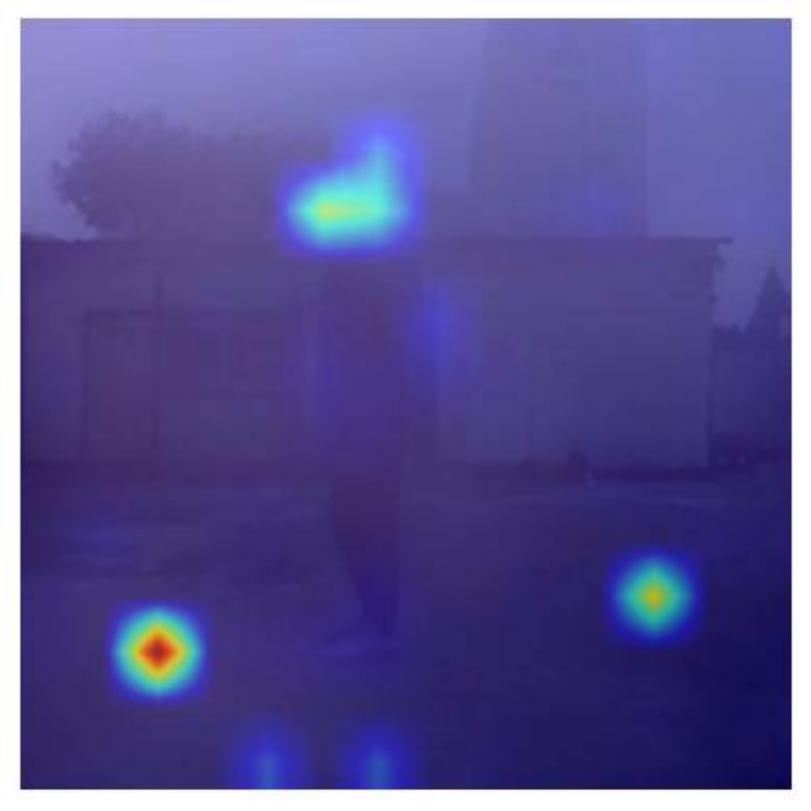} & \hspace{-0.4cm}\includegraphics[width =1.6cm,height=1.6cm]{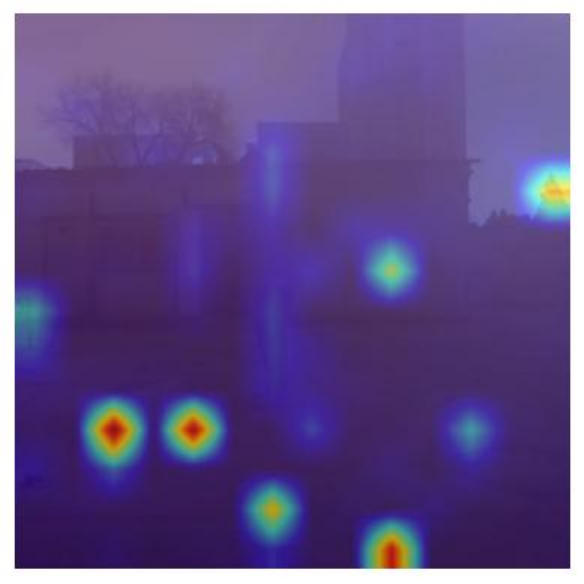} & 
            \hspace{-0.4cm}\includegraphics[width =1.6cm,height=1.6cm]{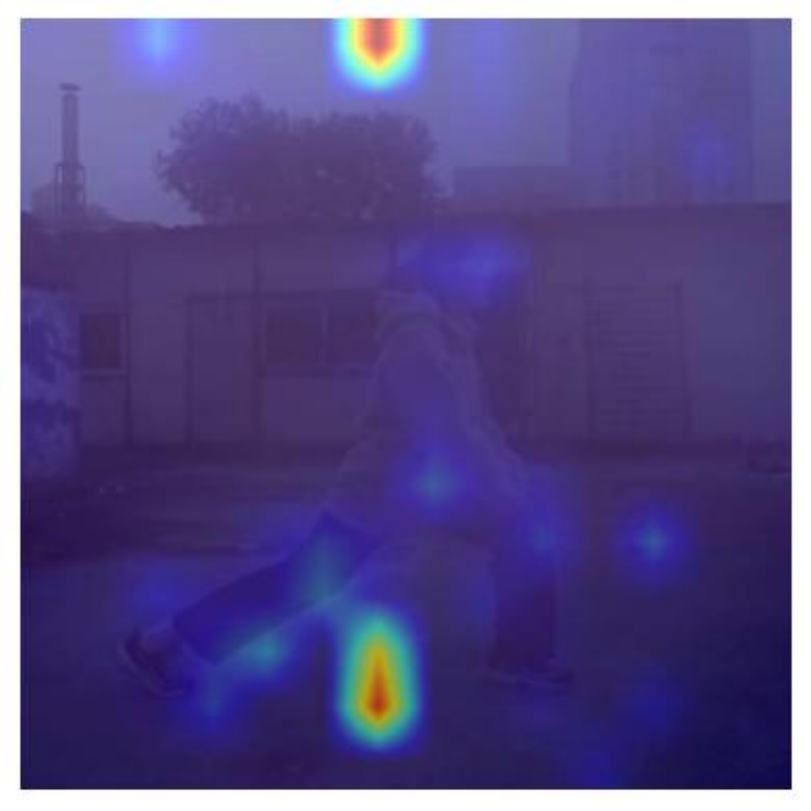} &
            \hspace{-0.4cm}\includegraphics[width =1.6cm,height=1.6cm]{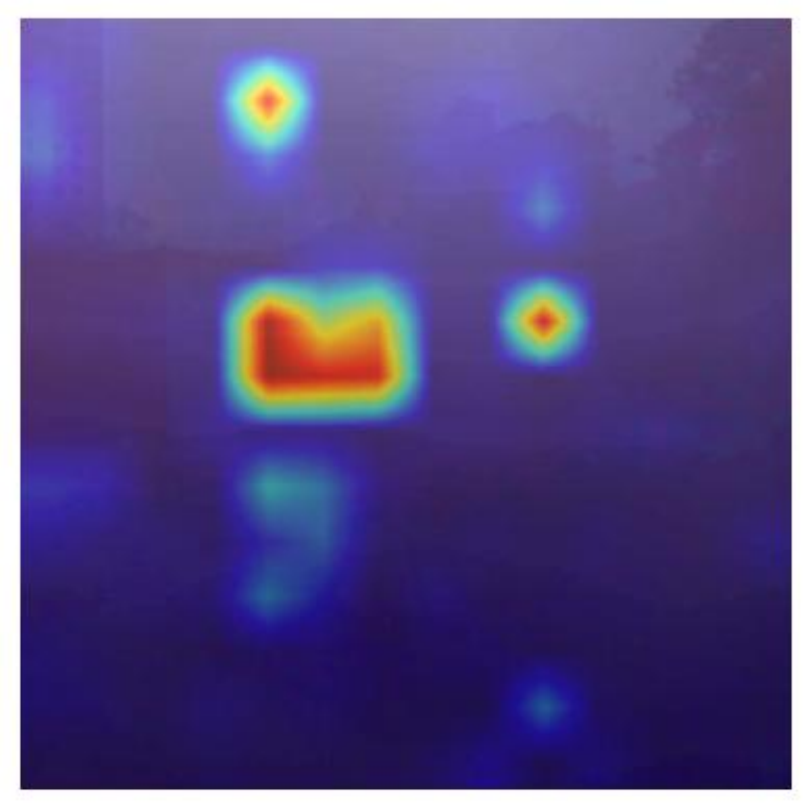}
            \\
            \vspace{-0.03cm}
            \vspace{-0.1cm}
            \hspace{-0.25cm}\raisebox{0.31cm}{\rotatebox{90}{\parbox[c]{1.0cm}{\centering \scalebox{0.8}{\small\text{+FAS}}}}} &
            \hspace{-0.35cm}\includegraphics[width =1.6cm, height=1.6cm]{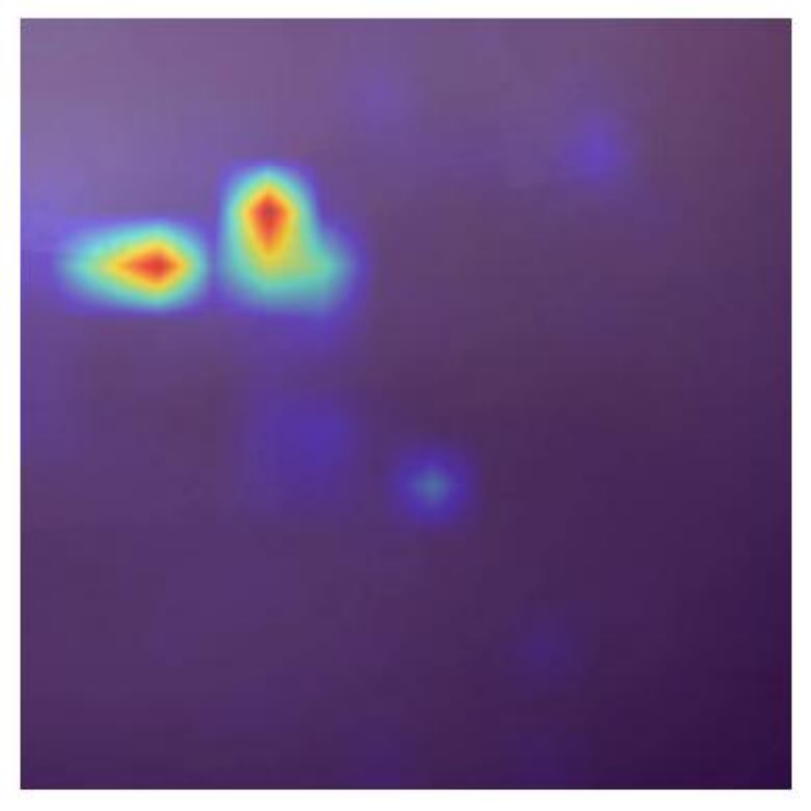} & 
            \hspace{-0.4cm}\includegraphics[width =1.6cm,height=1.6cm]{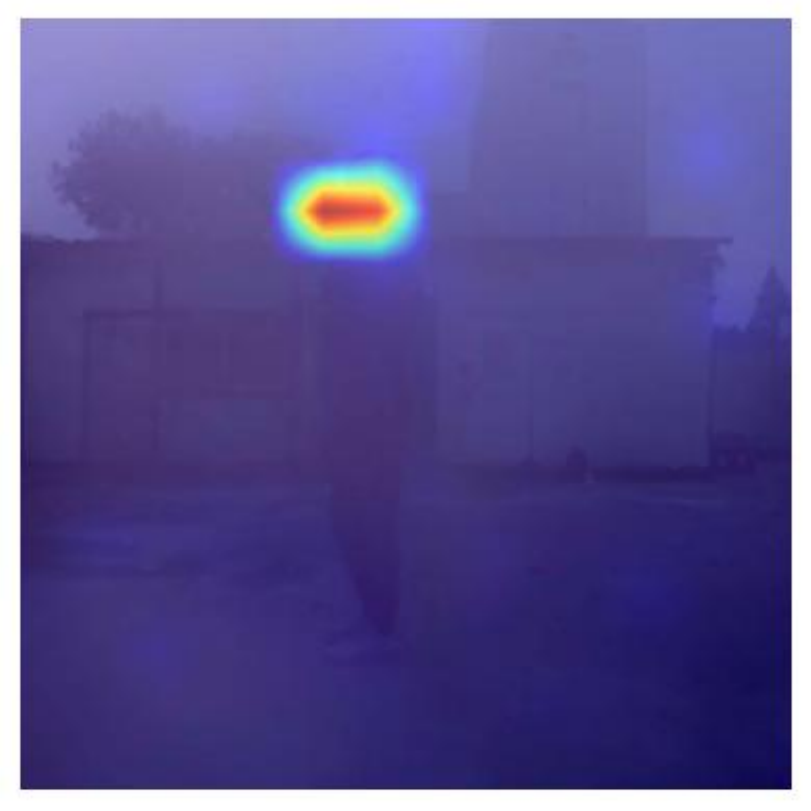} & \hspace{-0.4cm}\includegraphics[width =1.6cm,height=1.6cm]{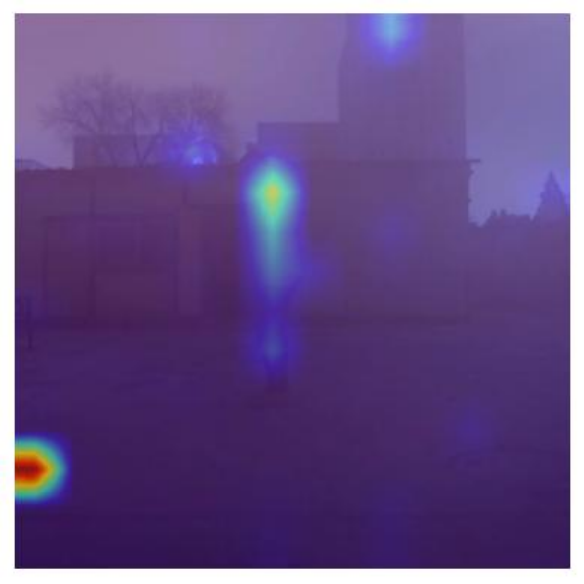} & 
            \hspace{-0.4cm}\includegraphics[width =1.6cm,height=1.6cm]{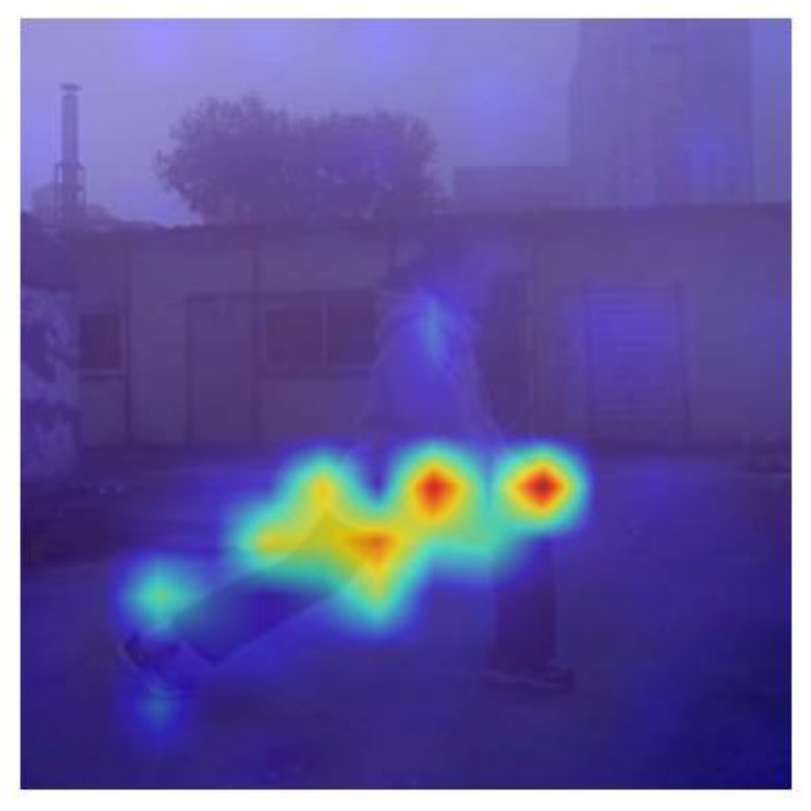} &
            \hspace{-0.4cm}\includegraphics[width =1.6cm,height=1.6cm]{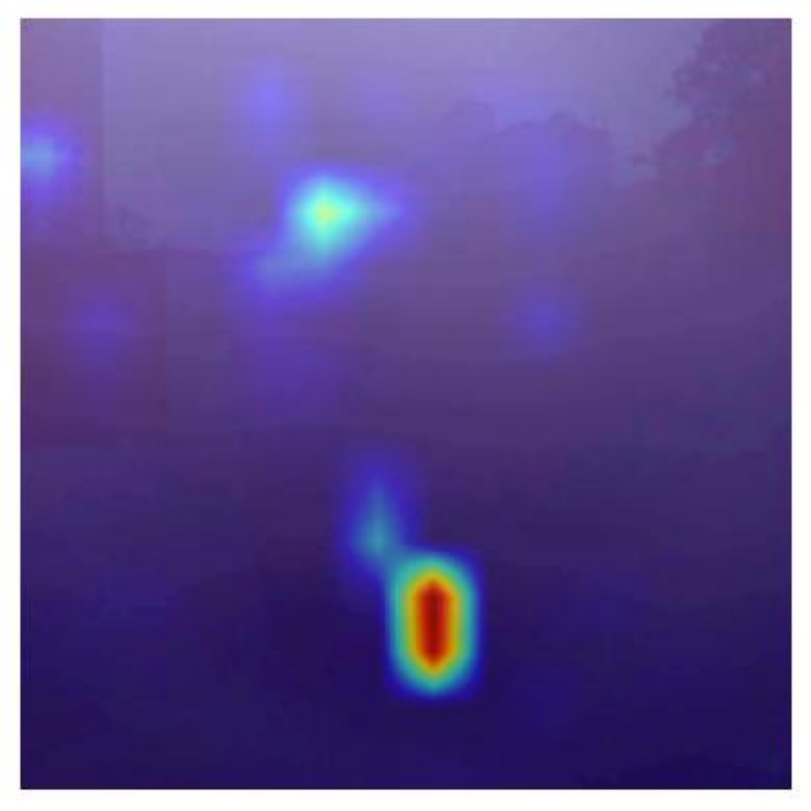}
            \\
            \vspace{-0.03cm}
            \vspace{-0.1cm}
            \hspace{-0.25cm}\raisebox{0.31cm}{\rotatebox{90}{\parbox[c]{1.0cm}{\centering \scalebox{0.8}{\small\text{+ME}}}}} &
            \hspace{-0.35cm}\includegraphics[width =1.6cm, height=1.6cm]{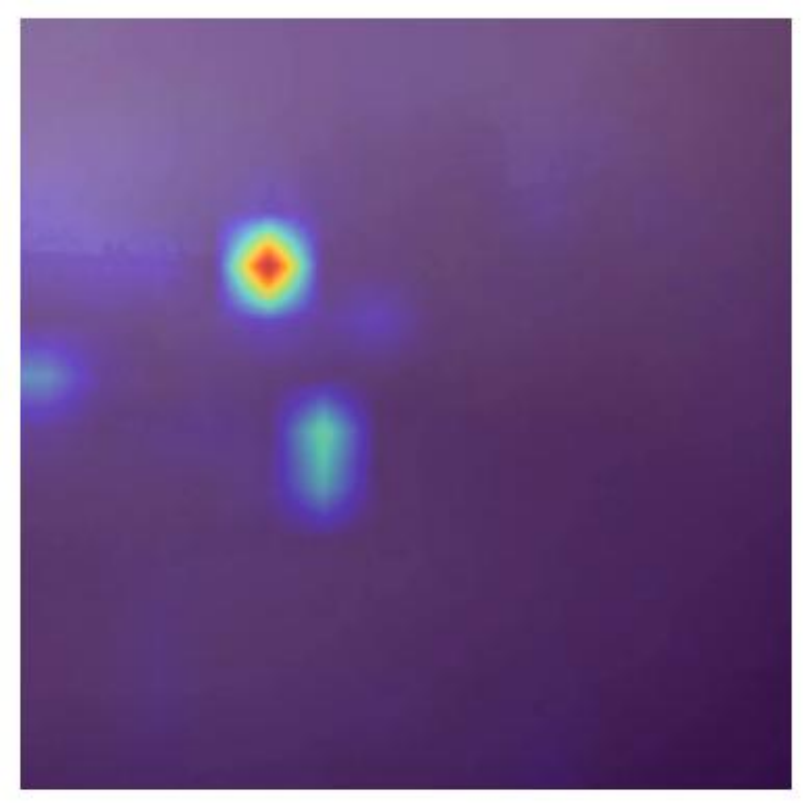} & 
            \hspace{-0.4cm}\includegraphics[width =1.6cm,height=1.6cm]{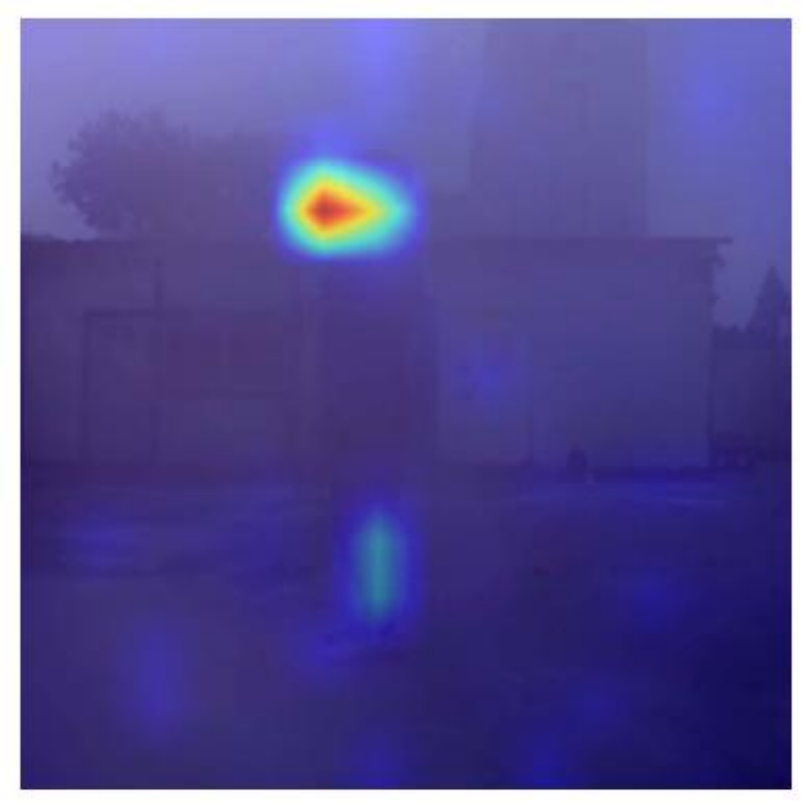} & \hspace{-0.4cm}\includegraphics[width =1.6cm,height=1.6cm]{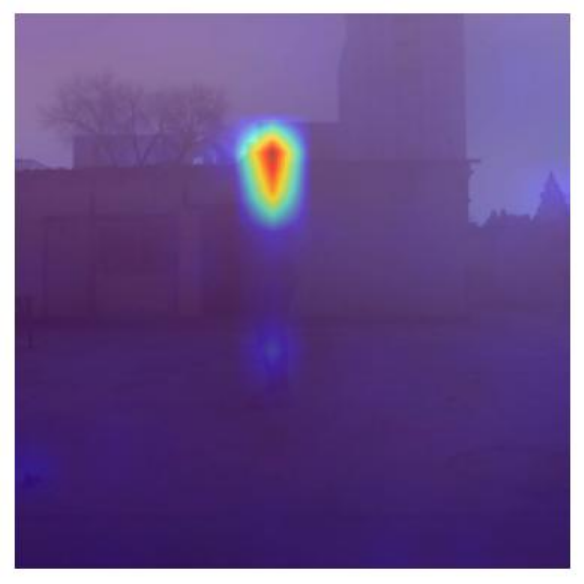} & 
            \hspace{-0.4cm}\includegraphics[width =1.6cm,height=1.6cm]{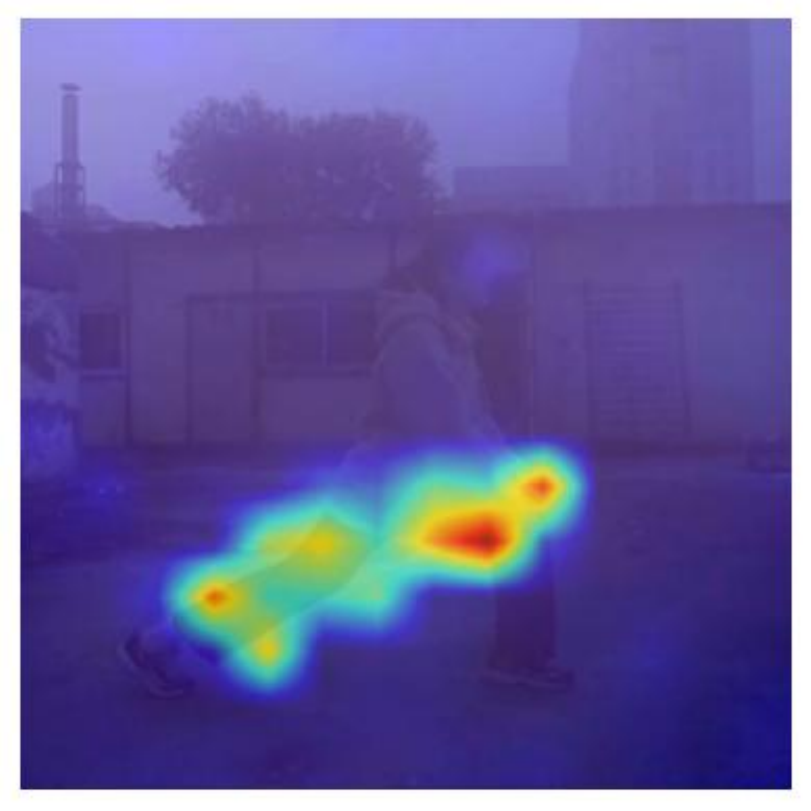} &
            \hspace{-0.4cm}\includegraphics[width =1.6cm,height=1.6cm]{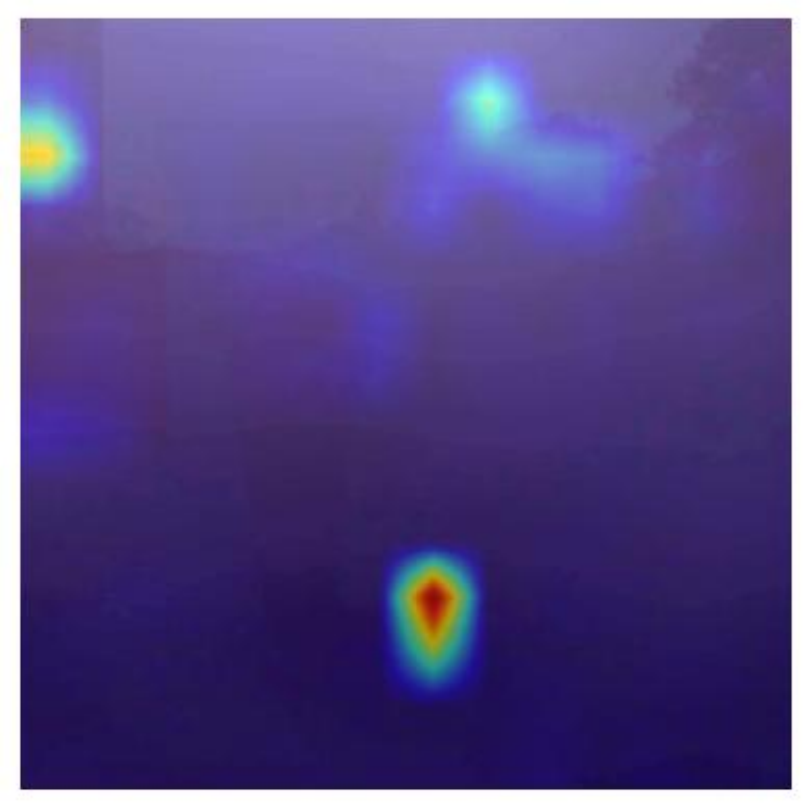}
            \\
            \vspace{-0.03cm}
            \hspace{-0.25cm}\raisebox{0.28cm}{\rotatebox{90}{\parbox[c]{1.0cm}{\centering \scalebox{0.8}{\small\text{Ours}}}}} &
            \hspace{-0.35cm}\includegraphics[width =1.6cm, height=1.6cm]{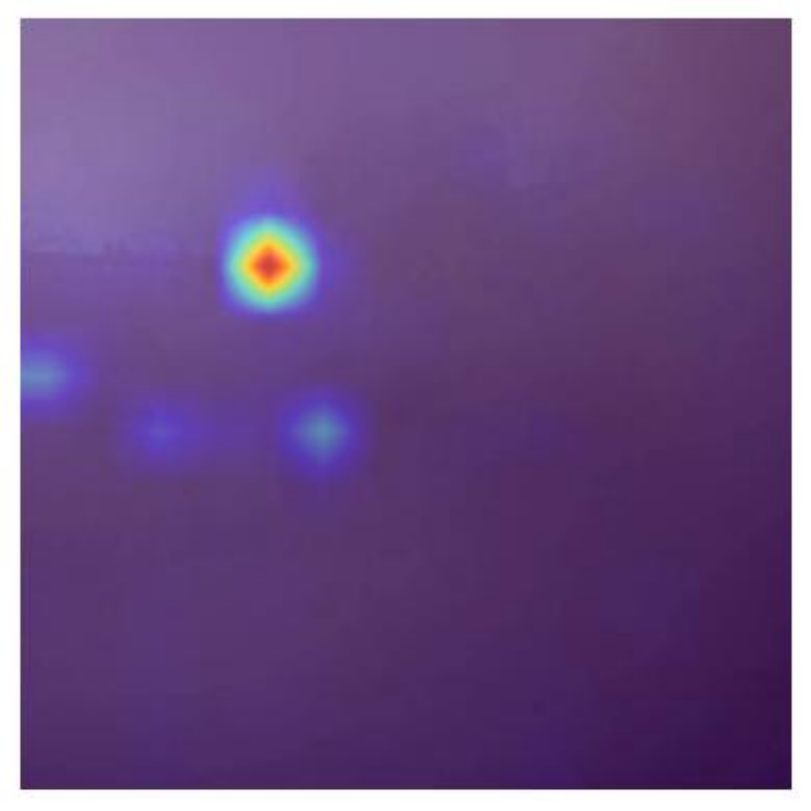} & 
            \hspace{-0.4cm}\includegraphics[width =1.6cm,height=1.6cm]{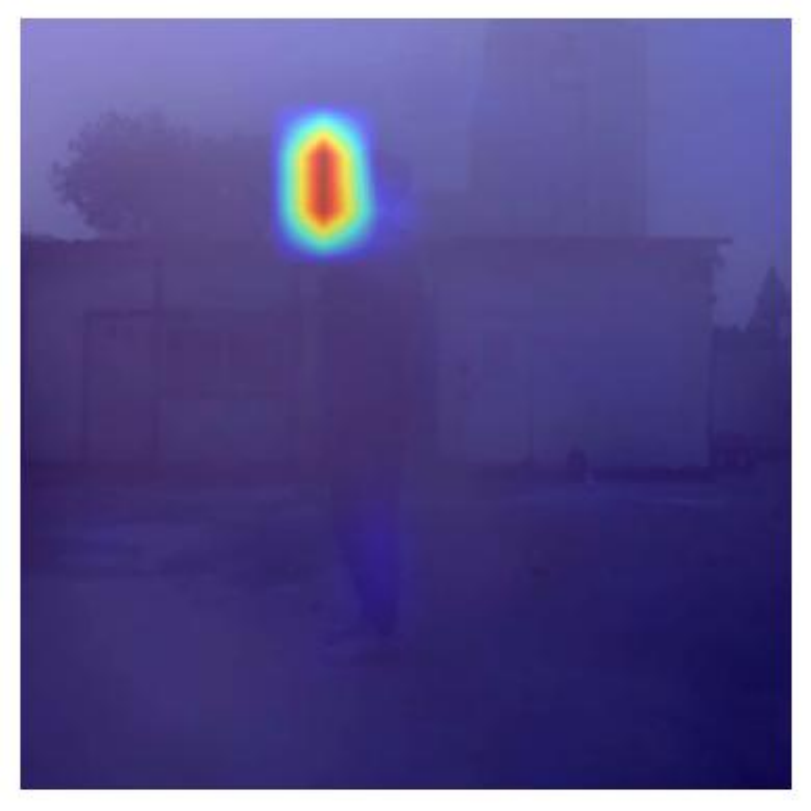} & \hspace{-0.4cm}\includegraphics[width =1.6cm,height=1.6cm]{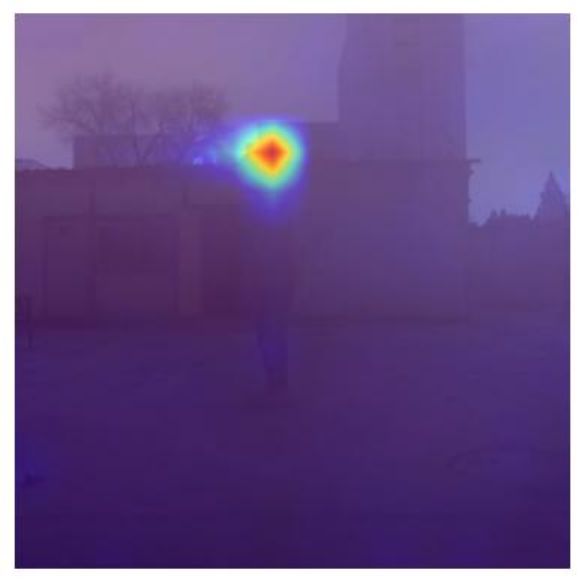} & 
            \hspace{-0.4cm}\includegraphics[width =1.6cm,height=1.6cm]{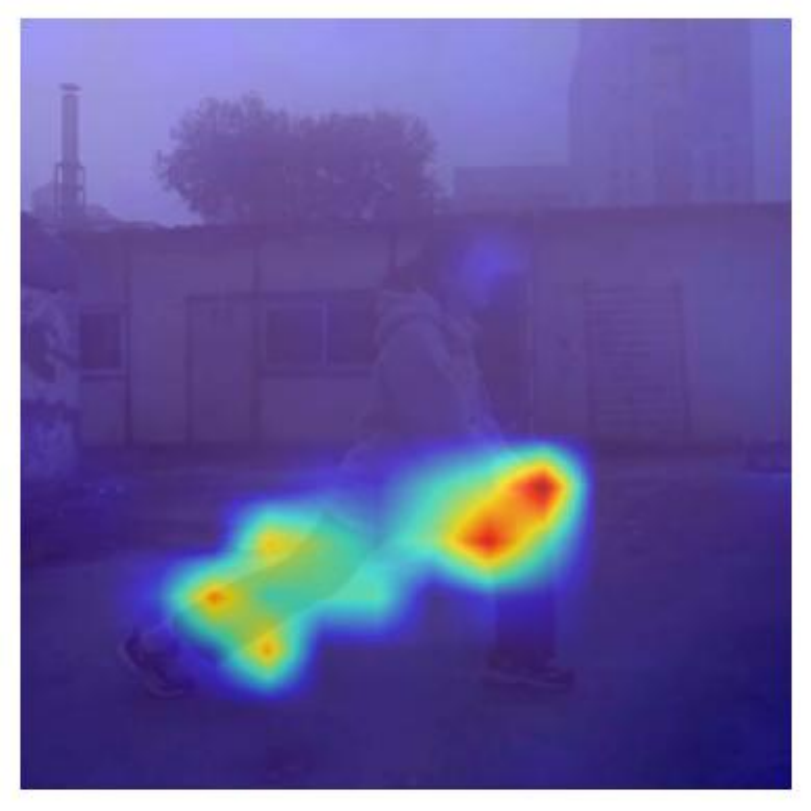} &
            \hspace{-0.4cm}\includegraphics[width =1.6cm,height=1.6cm]{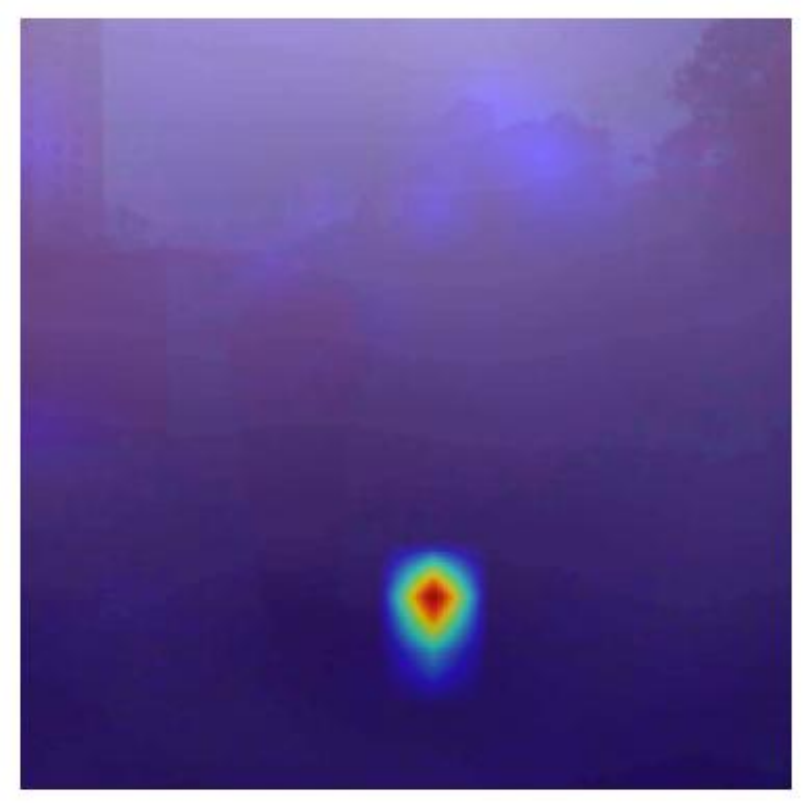}
            \end{tabular}
        \end{minipage}
    \end{minipage}
    \begin{minipage}[t]{\textwidth}
        \vspace{-0.2cm}
        \hspace{0.4cm}
        \begin{minipage}[t]{0.65\textwidth}
            \raisebox{0cm}{
                \begin{tabular}{ccccc}
                    \hspace{-0.58cm}\scalebox{0.8}{\small Provide direction} & 
                    \hspace{0.13cm}\scalebox{0.8}{\small Wave} &
                    \hspace{0.73cm}\scalebox{0.8}{\small Nod} & 
                    \hspace{0.68cm}\scalebox{0.8}{\small Lunge} & 
                    \hspace{0.3cm}\scalebox{0.8}{\small Sweep floor} 
                \end{tabular}
            }
        \end{minipage}
    \end{minipage}
\caption{
Heatmap comparisons with the baseline on FogAct.}
    \label{fig:heatmap}
\end{table}	

\subsection{Experimental Results} 
The recognition results comparing our method with others are shown in Tab.~\ref{tab:k400_sota}.
\textbf{One-stage methods}. 
(i) Image reconstruction-based methods perform suboptimally, with even the best-performing approach in this category, DINO-based frameworks~\cite{oquab2023dinov2}, achieving only a 45\% Top-1 accuracy. This is because they primarily focus on restoring visual details but struggle with action-relevant semantic features.
(ii) Video understanding method LLaMa-VID~\cite{li2024llama} achieves only 26.2\% accuracy. 
Despite leveraging large language models, it lacks effective temporal modeling under fog, failing to extract fog-invariant features from low-quality, blurred videos.
(iii) Action recognition methods relying solely on visual signals~\cite{rasheed2023fine, kim2024leveraging, yang2023aim} or vision-language-based strategies~\cite{wang2024multimodal, chen2024ost, wang2021actionclip} perform well in standard settings. However, they struggle under fog. This highlights the difficulty of transferring pre-trained models to degraded environments, where low visibility impairs feature extraction and action recognition.

The \textbf{two-stage approaches} combine the SOTA defogging works with SOTA action recognition approach.  
Regardless of using paired \cite{chen2024prompt} or unpaired \cite{wang2024ucl} defogging strategies,
they focus on restoring visual details yet struggle to recover action-relevant semantics. We even introduce the SOTA all-in-one approach \cite{yang2024language}, which claims better generalization as it is trained on all-weather datasets. However, the two-stage framework still struggles to improve classification performance of the defogged video embeddings. The best two-stage setup, combining the SOTA defogging method PTTD (fine-tuned on FogAct) and the SOTA action recognition model OST, achieves only 85.4\% Top-1 accuracy. In comparison, our one-stage approach bypasses the image-defogging step and achieves the best results, with 88.7\% Top-1 accuracy and 99.4\% Top-5 accuracy.

Additionally, to further demonstrate the effectiveness of our framework, we conduct experiments on the simulated datasets HMDB-51, UCF-101, and Kinetics-100, synthesized using common practices. As shown in Tab.~\ref{tab:super}, our method outperforms OST \cite{chen2024ost} by 9.8\% on the Kinetics-100 dataset, with an average accuracy improvement of 4.2\% across all three simulated datasets.

\subsection{Ablation Study and Analysis}

\noindent{\bf Model Architecture.} We study the effectiveness of our three components in Tab.~\ref{tab:comp}, including Fog-Aware Selection (FAS), Mutual Enhancement (ME), and Cross-Stream Alignment (CSA).
We observe that the three components all enhance action recognition accuracy in foggy conditions. The FAS structure outperforms the baseline by 13\%, while ME and CSA contribute increases of 6\% and 11\%, respectively. When combined, using all components achieves a Top-1 accuracy of 88.71\% and a Top-5 accuracy of 99.40\%. 

\noindent{\bf Attention Visualization.} As shown in Fig.~\ref{fig:heatmap}, our model effectively highlights action-relevant regions under fog by progressively refining attention. This improved focus over the baseline supports the effectiveness of our components, consistent with `Model Architecture'.

\noindent{\bf Necessity of FogNet and FogAct.}
Tab.~\ref{tab:motivation} shows that models trained on clean data degrade under fog due to low visibility, and defogging methods help only slightly while ignoring action semantics. Clean-trained models generalize poorly under fog due to domain shifts. In contrast, FogNet learns fog-invariant features, improving Top-1 accuracy by ~13\% under fog. Evaluations on ASM-simulated versus real-world foggy data further show that FogAct is more challenging and better reflects practical scenarios.

\noindent{\bf Effectiveness of Multiple Perspectives.} 
We compare single- and multi-perspective training on FogAct, evaluating both on all views as shown in Tab.~\ref{tab:single}. Using all four perspectives increases Top-1 by 36.0\% and Top-5 by 20.0\%. To control for data volume, we replicate the single-view data to match the size of the multi-view set.

\noindent{\bf Different Backbones.} Tab.~\ref{tab:backbone} presents an evaluation of the applicability of our method using different backbones. We observe that the effectiveness of the proposed modules remains consistent across different backbone architectures.

\noindent{\bf Analysis of Inference.}
Inference performance is strongly influenced by frame rate and views (\# temporal clip $\times$ \# spatial crop), as shown in Tab.~\ref{tab:views}. Increasing frames from 4 to 8 yields consistent accuracy gains across all datasets, highlighting the importance of temporal richness. Furthermore, using 4 temporal clips significantly boosts Top-1 accuracy over a single clip on FogAct, UCF-101, and Kinetics-100.

\noindent{\bf Fog Intensity Analysis.} 
Tab.~\ref{tab:multi} shows training with multiple fog intensities outperforms single-intensity settings. Models trained on dense fog generalize better than those on light fog. Even when controlling data volume via replication, the multi-intensity model achieves the best performance, surpassing dense fog by 21.2\% in Top-1 accuracy.

\begin{figure}[t]
\begin{center}
\includegraphics[width=0.48\textwidth]{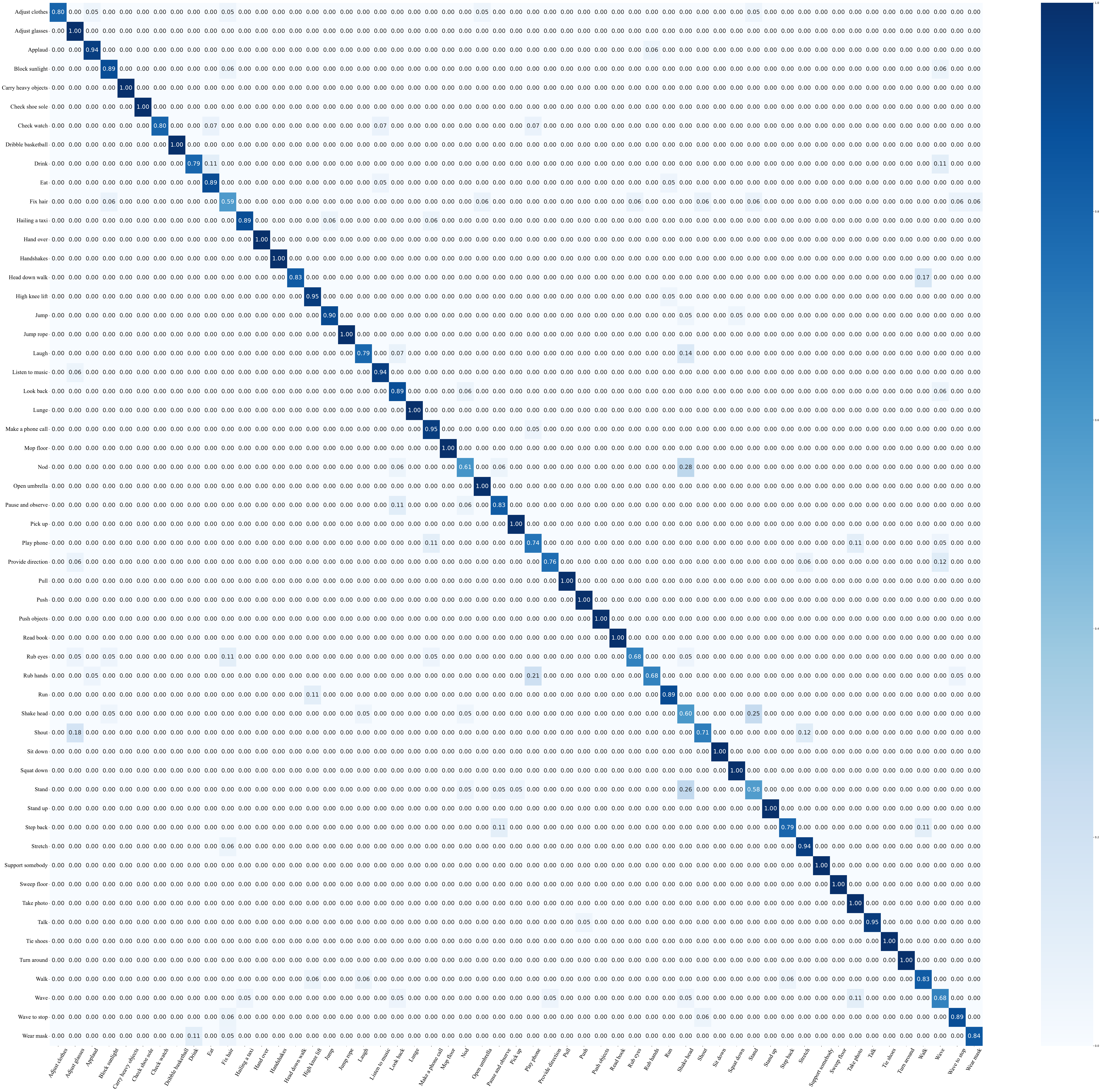}
\end{center}
\caption{Confusion matrix of our FogNet on the FogAct dataset.
}
\label{fig:confuse}
\end{figure}

\noindent{\bf Confusion Matrix.} Fig.~\ref{fig:confuse} presents the confusion matrix on FogAct. FogNet delivers high accuracy on actions with distinct motion patterns, such as \textit{Talk}, \textit{Adjust Glasses}, and \textit{Hand Over}. Performance drops on subtle or short-duration actions like \textit{Stand}, \textit{Fix Hair}, and \textit{Nod}, which exhibit high visual similarity and are easily obscured under dense fog.

\section{Conclusion}
In this paper, we present FogAct, the first real-world dataset for foggy action recognition with foggy–clean video pairs, covering 10 scenes and 55 action categories, providing sufficient training data and benchmarking. We further propose FogNet, an end-to-end framework that extracts fog-invariant features through three components: fog-aware selection, mutual enhancement, and cross-stream alignment. Leveraging large-scale pre-trained models, FogNet demonstrates strong generalization on real-world data. Extensive experiments on four challenging benchmarks (FogAct, UCF-101, HMDB-51, and Kinetics-100) validate the effectiveness of both FogAct and FogNet.

\section{Acknowledgement}
This work is supported by the National Natural Science Foundation of China (62302045), the Fundamental Research Funds for the Central Universities, and the BIT Special-Zone. This work is supported (in part) by the Opening Project of the State Key Laboratory of General Artificial Intelligence, BIGAI/Peking University, Beijing, China. (Project NO. SKLAGI20250P05).

{
    \small
    \bibliographystyle{ieeenat_fullname}
    \bibliography{main}
}

\end{document}